\documentclass[10pt,twocolumn,letterpaper]{article}

\usepackage{cvpr}      

\usepackage[accsupp]{axessibility} 

\usepackage{dashrule}
\usepackage{multirow}
\usepackage{makecell}
\usepackage{tabularx}
\usepackage[export]{adjustbox}
\usepackage{setspace}

\usepackage{algorithmic}
\usepackage{algorithm}
\usepackage[figuresright]{rotating}
\usepackage{bbm}
\usepackage{wrapfig}
\usepackage{stfloats}

\DeclareMathOperator{\Ccal}{\mathcal{C}}

\DeclareMathOperator{\Ecal}{\mathcal{E}}

\DeclareMathOperator{\Gcal}{\mathcal{G}}

\DeclareMathOperator{\Ical}{\mathcal{I}}

\DeclareMathOperator{\Vcal}{\mathcal{V}}

\DeclareMathOperator{\fb}{\mathbf{f}}

\DeclareMathOperator{\vb}{\mathbf{v}}

\DeclareMathOperator{\xb}{\mathbf{x}}

\newcommand{\argmax}{\mathop{\mathrm{argmax}}}

\newcommand{\softmax}{\mathop{\mathrm{softmax}}}
\newcommand{\concat}{\mathop{\mathrm{Concat}}}
\newcommand{\mlp}{\mathop{\mathrm{MLP}}}

\usepackage[dvipsnames]{xcolor}

\definecolor{cvprblue}{rgb}{0.21,0.49,0.74}
\usepackage[pagebackref,breaklinks,colorlinks,citecolor=cvprblue]{hyperref}

\def\paperID{14775} 
\def\confName{CVPR}
\def\confYear{2024}

\title{Constrained Layout Generation with Factor Graphs}

\author{
Mohammed Haroon Dupty{$^{1,}$}\thanks{Corresponding Author}
\hfil Yanfei Dong{$^{1,2}$}
\hfil Sicong  Leng{$^{3}$}
\hfil Guoji Fu{$^{1}$} \\
\hspace{6em} Yong Liang Goh{$^{1}$}
\hfil Wei Lu{$^{3}$} 
\hfil  Wee Sun Lee {$^{1}$}  \hspace{6em} \\
\normalsize{$^1$ National University of Singapore} \quad
\normalsize{$^2$ PayPal} \quad
\normalsize{$^3$ Singapore University of Technology and Design}\\
{\tt\normalsize \{haroon, dyanfei, guoji.fu, gyl\}@u.nus.edu, sicong\_leng@mymail.sutd.edu.sg,}\\
{\tt\normalsize luwei@sutd.edu.sg,  leews@comp.nus.edu.sg}
}
\begin{document}
\maketitle
\begin{abstract}
This paper addresses the challenge of object-centric layout generation under spatial constraints, seen in multiple domains including floorplan design process.
The design process typically involves specifying a set of spatial constraints that include object attributes like size and inter-object relations such as relative positioning. 
Existing works, which typically represent objects as single nodes, lack the granularity to accurately model complex interactions between objects. For instance, often only certain parts of an object, like a room's right wall, interact with adjacent objects. To address this gap, we introduce a factor graph based approach with
four latent variable nodes for each room, and a factor node for each constraint. The factor nodes represent dependencies among the variables to which they are connected, effectively capturing constraints that are potentially of a higher order.  We then develop message-passing on the bipartite graph, forming a factor graph neural network that is trained to produce a floorplan that aligns with the desired requirements. 
Our approach is simple and generates layouts faithful to the user requirements,  demonstrated by a large improvement in IOU scores over existing methods. Additionally, our approach, being inferential and accurate, is well-suited to the practical human-in-the-loop design process where specifications evolve iteratively, offering a practical and powerful tool for AI-guided design. 
\end{abstract}    
\section{Introduction}
\label{sec:intro}

The incorporation of AI into the layout design process represents a significant advancement in design methodologies. Researchers have predominantly adopted a generative modeling approach, enabling the generation of diverse designs with limited user inputs~\cite{arroyo2021variational, li2019layoutgan,lee2020neural,nauata2020house,nauata2021house}. Although very useful, this approach faces practical challenges when we have a fixed uneven boundary and users have several specific requirements and seek active involvement in the design process.  In such instances, inferential models specifically trained to produce layouts that align precisely with user requirements are likely more beneficial. 

The challenge of layout design generation with constraints is more prominently seen in the design of \emph{floorplan layouts}~\cite{arroyo2021variational, li2019layoutgan,lee2020neural,nauata2020house,nauata2021house, wu2019data,hu2020graph2plan,laignel2021floor,para2021generative,sun2022wallplan}, which can include both user-defined and structural constraints. User-defined constraints stem from individual preferences regarding room sizes, adjacencies, and other design considerations. Structural constraints may require the layout to remain within the boundary of a building or adhere to other geometric restrictions.
For example, given a boundary, a user may want two rooms adjacent to each other in a particular corner of the house. A good inferential model would be able to produce a layout satisfying these constraints. With partial output available, users can append or update their requirements and thereby, refine the design iteratively. 

\begin{figure}[t]
\begin{center}
   \includegraphics[width=0.95\linewidth,height=3.85cm]{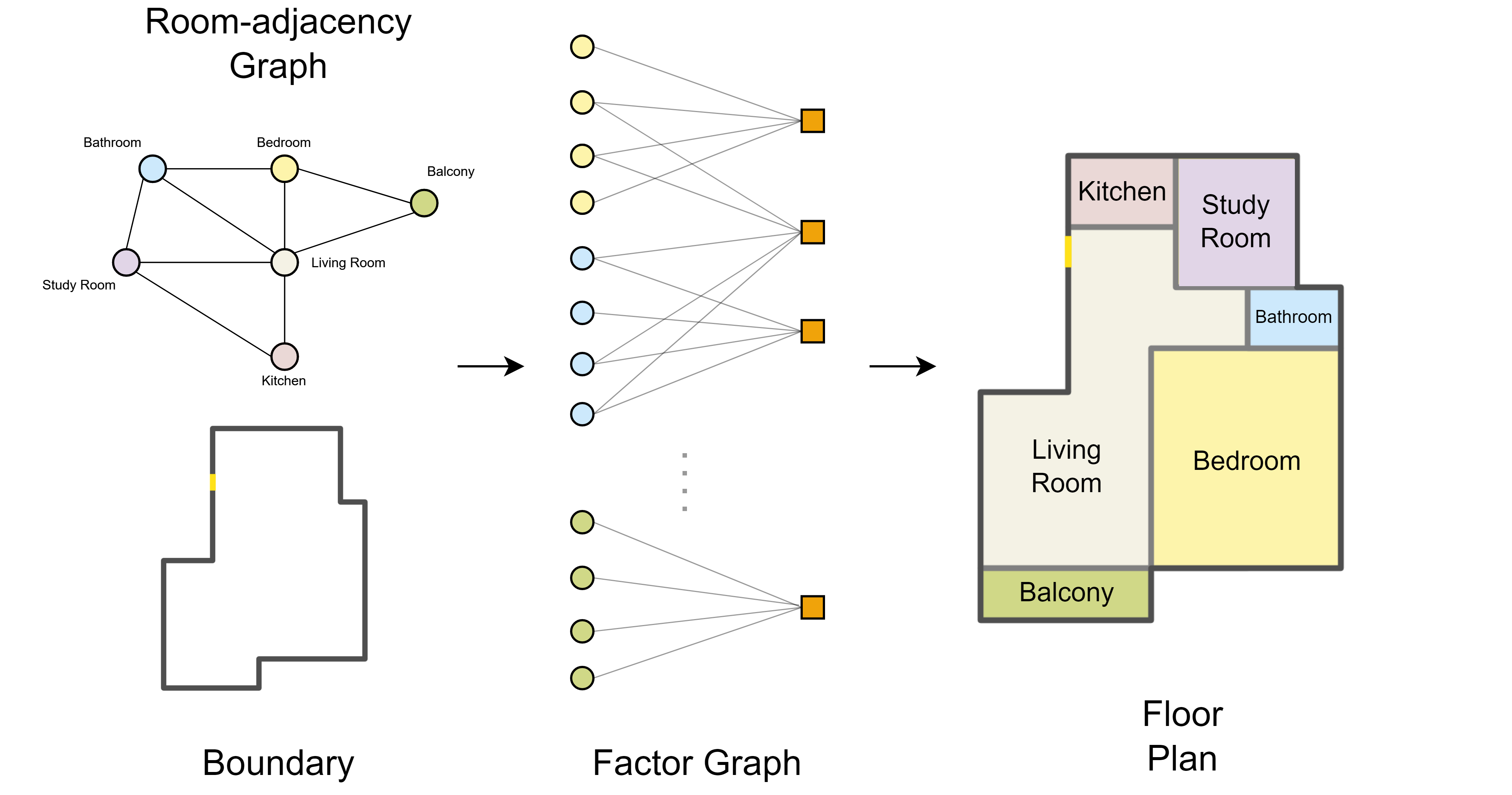}
\end{center}
\setlength{\abovecaptionskip} {-2pt}
   \caption{\small Given a building boundary and a graph that encodes user requirements and room constraints, we first transform the inputs into a factor graph that can model higher-order constraints, and then produce the floorplan based on the factor graph model.}
\label{fig:intro}
\vskip -0.175in
\end{figure}

One common approach for producing floorplan layouts is to use a two-stage method, where the first stage involves predicting bounding boxes for individual objects/rooms, followed by a refinement network that converts the bounding boxes into an image~\cite{hu2020graph2plan,chen2017photographic}. For the first stage, the room-adjacency graph is taken as input, and a Graph Neural Network(GNN)-based model is often used to predict the bounding boxes of rooms~\cite{hu2020graph2plan,chen2020intelligent}. 
However, the representation of each object as a single node, while straightforward, falls short in terms of granularity and effectiveness. It notably struggles to accurately reflect the intricate interactions that occur between objects, as these interactions are frequently localized to specific parts of an object. For example, it is often the case that only one side of a room, such as the right wall, is actively engaged with an adjacent object.
Therefore, we represent each room using four separate latent variables to represent the four boundaries of the bounding box: $x_{min}$, $x_{max}$, $y_{min}$, and $y_{max}$. This approach effectively captures fine-grained interactions, yet it also introduces the necessity to model higher-order dependencies. For instance, determining the size of a room requires all four variables. Modeling the potentially higher-order dependencies, however, presents a non-trivial challenge.

To handle the challenge, we propose a factor graph based model to encode the dependencies. A factor graph is a bipartite graph consisting of factor nodes and variable nodes. In our model, each room is represented by four variable nodes, and each factor node is designed to represent the dependencies among its connected variables. 
This enables modeling arbitrary constraints including those of a higher-order, a capability lacking in traditional GNNs which can model only pairwise dependencies.
This allows us to leverage domain knowledge more effectively than the adjacency-graph approach typically used in existing works. For instance, we impose a boundary constraint ensuring that all rooms are contained within a predefined boundary 
by employing a factor that connects to all the relevant variables. 
We associate an embedding with each factor graph node to represent the latent information associated with the node and design message passing operations on the factor graph, leading to a Factor Graph Neural Network (FGNN) \cite{zhang2020factor,dupty2020neuralizing}.

As our objective is to produce floorplan layouts that satisfy all input constraints, we evaluate our model's performance with a focus on fidelity to the ground-truth image that is used to derive the constraints. Our model surpasses other strong baselines, achieving a significant improvement in both box-level IOUs and pixel-level metrics, indicating the effectiveness of our approach. 
Furthermore, we demonstrate our model's usefulness in two distinct practically useful scenarios. One, we show that our model is well-suited to the iterative refinement of human requirements, enabling the development of varied designs with user-interactions. Two, we show that our model can form part of a generative pipeline and generate diverse realistic layout designs for the same given input boundary, validated both qualitatively and quantitatively.
\section{Related Work}
 
The predominant approach taken by the researchers towards addressing the automated design problem is through conditional generative modeling~\cite{arroyo2021variational, li2019layoutgan,lee2020neural,nauata2020house,nauata2021house, wu2019data,hu2020graph2plan,laignel2021floor,para2021generative,sun2022wallplan,kikuchi2021constrained}. Layout generative models with a focus on residential floorplan layouts, aim to create room layouts that satisfy limited user constraints of room types and adjacencies, with or without building boundaries~\cite{zhong2019publaynet,deka2017rico,xiao2013sun3d,cao2022geometry,wu2019data}. Previous works have used various strategies to achieve this, such as training probabilistic methods~\cite{merrell2010computer,rosser2017data} or using evolutionary strategies based on user specifications and constraints~\cite{rodrigues2013evolutionarya, rodrigues2013evolutionaryb}, and designing constraint graphs~\cite{hu2020graph2plan, para2021generative}.

In\cite{wu2019data}, a two-stage approach learning network was proposed for generating floorplans for residential buildings, and the RPLAN dataset was released. Since then, follow-up works have focused mainly on building GAN-based generative models. These models typically take in a room-adjacency graph called bubble diagrams. Some of the popular models in this line of research include Housegan, Housegan++, and other variants~\cite{nauata2020house,nauata2021house,chang2021building,para2021generative,ozerol2021generation,upadhyay2023floorgan,sun2022wallplan,shabani2022housediffusion}. 
While these approaches generate realistic floorplans, they have limitations when users are seeking to actively vet most aspects of the layout. Recently,~\cite{he2022iplan} have proposed a generative modeling approach which allows human-in-the-loop mechanism. However, they do not exploit the relational constraints that users are likely to provide. Furthermore, in this work, we target the problem from an iterative design perspective where users refine their requirements iteratively upon seeing the results of their partial requirements. In such cases, an inferential model which focuses on maximizing fidelity to the user requirements is likely more useful.

Research on floorplan layout generation, focused on creating layouts from input room-adjacency graphs that match ground truth is found in Graph2Plan~\cite{hu2020graph2plan} and HPGM~\cite{chen2020intelligent}. Our work builds upon Graph2Plan's two-stage methodology, initially predicting room bounding boxes and then utilizing these predictions for layout creation. 
We propose transforming input constraints into a factor graph and employing neuralized message passing for effective constraint utilization and domain knowledge integration. Neural message passing on factor graphs has recently shown impressive results on some tasks like molecular prediction and navigation
~\cite{dupty2020neuralizing,zhang2020factor,satorras2021neural,yoon2019inference,du2020compositional,sodhi2022leo}. 
Our work mainly builds on~\cite{dupty2020neuralizing, zhang2020factor,JMLR:v24:21-0434} with learnable bipartite message passing between variable and factor nodes. 

\section{Preliminaries}
\subsection{Factor Graphs}
A factor graph is a bipartite graph $\Gcal = (\Vcal,\Ccal,\Ecal)$ with two disjoint sets of nodes, variable nodes $\Vcal$, and factor nodes $\Ccal$. A variable node $i\in\Vcal$ represents a variable $x_i \in \xb$ and each factor node $c\in \Ccal$ specifies that there are dependencies among a set of variables $\xb_c$; an edge $(c,i)\in\Ecal$ exists between variable node $i$ and factor node $c$ if $x_i \in \xb_c$. 

Factor graphs are widely used to capture dependencies in probabilistic graphical models (PGM). In PGMs, approximation algorithms of inferring optimal values of variables $\xb^*$ often operate by running message passing on factor graphs, for example, the \emph{belief propagation} algorithms. 

\subsection{Factor Graph Neural Network}
Message-passing operations on a graph can be represented as a graph neural network. In a similar way, we can represent message passing on a factor graph as a factor graph neural network (FGNN)~\cite{dupty2020neuralizing,zhang2020factor,satorras2021neural}. 
By assuming tensor decomposed factor potentials, both sum-product and max-product belief propagation algorithms for PGMs can be represented with compact FGNNs~\cite{dupty2020neuralizing,zhang2020factor,JMLR:v24:21-0434}. With the use of more powerful approximators within the FGNN, it may be possible to learn even more powerful inference algorithms specialized for the problem domain.

In FGNN, a bipartite factor graph is established with variable and factor nodes, where each factor node connects to its dependent variable subset. These nodes are represented by variable ($\vb_i$) and factor ($\fb_c$) embeddings, initialized using respective features. The embeddings are then updated iteratively through message-passing equations.
\begin{align}
\tilde{\vb}_i = \text{AGG}_{c \in N(i)}\;\Psi_{\text{FV}}(\fb_c, \vb_i)\\
\tilde{\fb}_c = \text{AGG}_{i \in N(c)}\;\Psi_{\text{VF}}(\vb_i, \fb_c)
\end{align}
where $\Psi_{\text{FV}}$ and $\Psi_{\text{VF}}$ are functions with learnable parameters. Note that the factor graph is a bipartite graph. Therefore, $N(i)$ will contain only factor nodes $c$ and $N(c)$ will contain only variable nodes $i$. This process can be iterated multiple times, followed by a readout function for tasks like regression or classification.

\section{Proposed Method}
\subsection{Problem Setup}
In our problem setup, we are given a room-adjacency graph $G$ and a boundary of the building $B$. The graph $G$ is as described in Graph2Plan~\cite{hu2020graph2plan} and contains a node for each room with attributes including the size and position, and edges are typed adjacencies describing the spatial relationship between rooms. The boundary $B$ is a 2D binary mask describing the inside area of the building. The task is to produce the floorplan layout image $\Ical$ which satisfies all the constraints in the input.
Our problem setup differs from many GAN-based methods, as they often prioritize generation diversity. In contrast, our objective is to produce a layout that satisfies the specified constraints. Therefore, in addition to generating realistic layouts, we also aim to produce layouts that closely resemble the ground truth.

To address the task, we follow conventional methods in adopting a two-stage approach~\cite{hu2020graph2plan,chen2020intelligent}.
In the first stage, we output the coordinates of bounding boxes for each room, while in the second stage, we produce the floor plan layout image using the predicted bounding boxes. Our primary focus is on enhancing the accuracy of the first stage by leveraging domain knowledge, such as higher-order relations. 
To this end, we seek to effectively incorporate domain knowledge, including higher-order relations, into the model.

\begin{figure*}[t]
\begin{center}
   \includegraphics[width=0.8\textwidth]{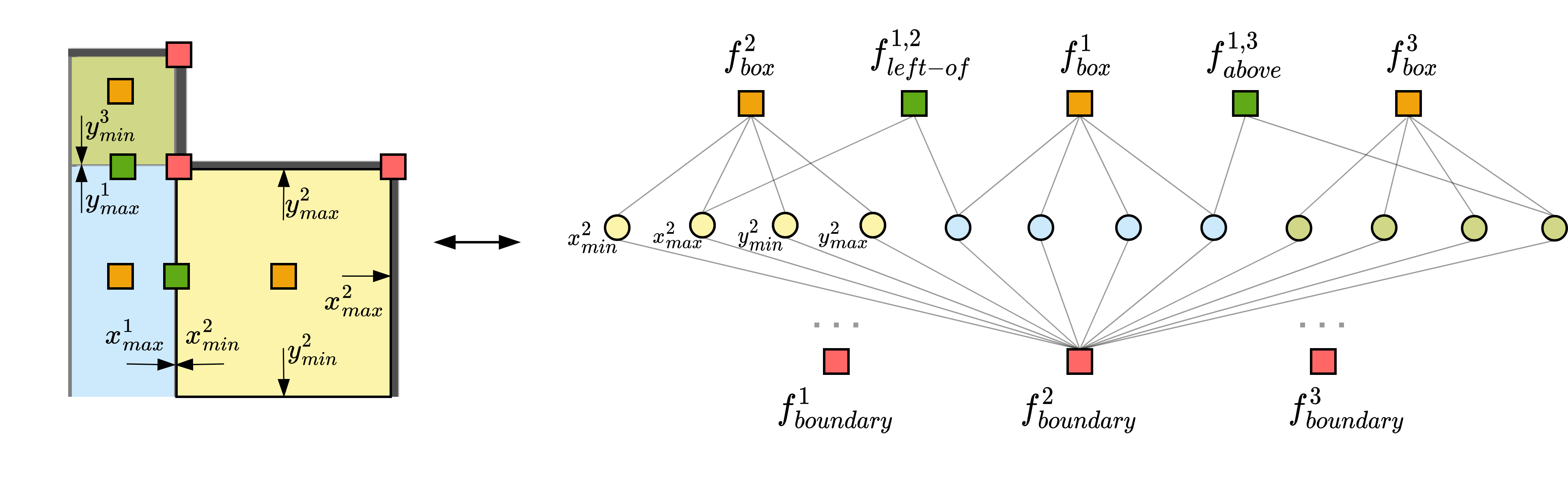}
\end{center}
\setlength{\abovecaptionskip}{-12pt}
\setlength{\belowcaptionskip}{-5pt}
   \caption{\small Illustration of the proposed factor graph model for floorplan design. Each room is represented with four bounding-box variables. Factors connect the variables based on input constraints and domain knowledge.$f_{box}^i$ connects four variables of room $i$. $f_{rel}^{s,o}$ connects only relevant subset of variables between rooms $s$ and $o$. $f_{boundary}^k$ represents $k^{th}$ corner-point and connects to all the variables. Message passing on this factor graph helps learn better room coordinates which are then used to produce the layout image. }
\label{fig:factor_graph_main}
\end{figure*}
\subsection{Representation}

For the first stage of predicting bounding boxes, conventional methods leverage GNNs to process the input graph and output bounding box coordinates, which are then used to produce the layout image. However, representing each object as a single node is a straightforward approach but lacks in granularity and effectiveness. This method often fails to capture the complex interactions between objects, which are typically concentrated on certain parts of an object. A common example is that only a specific side of a room, like the right wall, might be actively interacting with a neighboring object.

To capture the fine-grained interactions, we propose to represent each room with four different variables. 
Consider a floorplan image as a single-channel image of size $H \times W$ with each room represented by an enclosing bounding box. Let the room $i$'s bounding box be described by four coordinates in the form $(x_{min}^i, x_{max}^i, y_{min}^i, y_{max}^i)$ with $x_{min/max}\in \{1\dots W\}$ and $y_{min/max}\in \{1\dots H\}$. Then the set of variables $\xb$, describing the bounding boxes of all the rooms, can be stated as 
\[\xb = \big\{x_{min}^i, x_{max}^i, y_{min}^i, y_{max}^i |\; i\in\{1\dots N\}\big\}\]
where $N$ is the number of rooms in a given floorplan. 
Given the boundary $B$ and the input graph $G$, we aim to predict the values of the configuration of variables $\xb$ which optimally satisfies all the constraints in $B$ and $G$. 

\subsection{Floor-Plan Factor Graph Model}
In layout design, we frequently encounter high-order constraints that traditional GNNs cannot capture due to their limitation in modeling only pairwise dependencies. To effectively encode these constraints, we propose the adoption of factor graphs. Factor graphs provide the needed modeling capacity to capture the information available from domain knowledge. With factor graphs, the mutual dependencies between the variables can be easily encoded in the form of factor nodes. Importantly, input constraints provide important clues regarding the variable dependencies. For example, with our defined variables $x_{min}, x_{max}, y_{min}, y_{max}$ per each box, 
when two rooms are adjacent with a \emph{left-right} relationship, it is likely that the $x_{max}$ and $x_{min}$ values of the two rooms are correlated. By recognizing these correlations, we can leverage domain knowledge to enhance the model's accuracy in predicting bounding box coordinates.

Overall, we model four kinds of factors: box factors, relational factors, boundary factors, and a complete factor. Except relational factors, the rest are higher-order factors that capture dependencies between multiple variables. An illustration of the model is shown in Fig.~\ref{fig:factor_graph_main}.

\subsubsection{Bounding Box Factors}
We have four variables per bounding box, and these variables are likely to be highly correlated. For example, if the room size is 100 sqft, the four variables of the room must satisfy the constraint that $(x_{max}-x_{min}) * (y_{max}-y_{min}) = 100$. This constraint is a higher-order factor that can provide a useful inductive bias for learning the values of the $x$ and $y$ variables. To encode this information, we introduce a box factor for each room $i$, denoted as $f_{Box}^i = \{x_{max}^i, x_{min}^i, y_{max}^i, y_{min}^i\}$, into the factor graph model. 

\subsubsection{Relation factors}
\begin{table*}
\aboverulesep=0ex
 \belowrulesep=0ex
\centering
\begin{minipage}[l]{0.65\textwidth}\centering
    \begin{minipage}[t]{0.95\textwidth}\centering
    \resizebox{\textwidth}{!}{
        \begin{tabular}{ c c c c c }
            \toprule
            \rule{0pt}{1.2EM}
            \textbf{Relation} & \emph{left-of} & \emph{right-of} & \emph{above} & \emph{below}  \\  
            \midrule
            \rule{0pt}{1.2EM}
            \textbf{Factors} & 
            \{$x_{max}^s$, $x_{min}^o$\} &
            \{$x_{min}^s$, $x_{max}^o$\} &
            \{$y_{max}^s$, $y_{min}^o$\} &
            \{$y_{min}^s$, $y_{max}^o$\}\\
            \addlinespace
            \midrule 
           \midrule 
           \rule{0pt}{1.2EM}
            \textbf{Relation} & \emph{left-above} & \emph{right-above} & \emph{left-below} & \emph{right-below} \\
            \midrule
            \rule{0pt}{1.2EM}
            \multirow{2}{*}{\textbf{Factors}} &
            \{$x_{max}^s$, $x_{min}^o$\} & \{$x_{min}^s$, $x_{max}^o$\} & \{$x_{max}^s$, $x_{min}^o$\} & \{$x_{min}^s$, $x_{max}^o$\}\\ 
            & \{$y_{max}^s$, $y_{min}^o$\} &  \{$y_{max}^s$, $y_{min}^o$\} & \{$y_{min}^s$, $y_{max}^o$\} & \{$y_{min}^s$, $y_{max}^o$\}\\
            \addlinespace
            \bottomrule
        \end{tabular}
        }
    \end{minipage}%
    \end{minipage}\hfill
    \begin{minipage}[r]{0.35\textwidth}\centering
        \resizebox{0.95\textwidth}{!}{
        \begin{tabular}{ l c  c  }
           \toprule
           \rule{0pt}{1.4EM}
            \textbf{Relation} & \emph{inside} & \emph{surrounding}  \\  
            \addlinespace
            \midrule
            \rule{0pt}{1.4EM}
            \multirow{4}{*}{\textbf{Factors}} & \{$x_{min}^s$, $x_{min}^o$\} & \{$x_{min}^s$, $x_{min}^o$\} \\  
            \addlinespace
            & \{$y_{min}^s$, $y_{min}^o$\} & \{$y_{min}^s$,  $y_{min}^o$\} \\ 
            \addlinespace
            & \{$x_{max}^s$, $x_{max}^o$\} & \{$x_{max}^s$, $x_{max}^o$\} \\ 
            \addlinespace
            & \{ $y_{min}^s$, $y_{max}^o$\} & \{$y_{min}^s$, $y_{max}^o$\} \\  
            \addlinespace
            \bottomrule
            
        \end{tabular}
        }
    \end{minipage}
    \caption{\small Full set of relation types and their corresponding factors $f_{rel}$ defined in our factor graph model.}
    \label{tab:relation-factors}
    \vskip -0.15in
\end{table*}

The spatial room adjacencies available in the input graph provide rich information about the relative locations of the rooms and need to be effectively exploited. We work on the typed adjacencies defined in Graph2Plan~\cite{hu2020graph2plan}, which are based on the spatial relation type of the pair of adjacent rooms. The spatial relation types are: \emph{left-of, right-of, above, below, left-above, left-below, right-above, right-below, inside, and surrounding}. 

Our factor-graph-based modeling with $4$ variables per room, allows us to only link the variables which are likely to have direct dependencies. 
We introduce factors connecting specific variables in rooms $s$ and $o$ depending on the relation type $p$ of the typed adjacency $<s,p,o>$. For instance, if the relation is $<room_s, \emph{left-of}, room_o>$, it is probable that the $x_{max}$ coordinate of room $s$ is almost equal to the $x_{min}$ coordinate of room $o$. Therefore, we add a factor $f_{rel}^{s,o}=\{x_{max}^s,x_{min}^o\}$ to capture this relation.

Similarly, we introduce factors based on the relation types which imply specific coordinate relationships between the subject and the object rooms.
Table~\ref{tab:relation-factors} describes the full set of relational factors in our model for each of the relation types. Overall, we have $20$ distinct relational factor types for the $10$ types of defined relations.

\subsubsection{Boundary factors}\label{subsec:boundaryfactors}
In practice, the outer boundary of a floorplan is one of the most important constraints as it directly affects the placement of each room and the way in which different rooms should be aligned within the floorplan boundary. However, incorporating such boundary information is not trivial. Previous methods like RPLAN and Graph2Plan~\cite{hu2020graph2plan,wu2019data} use a CNN encoder with a 2-dimensional binary boundary mask to output a boundary embedding, which is then used to produce the floorplan. However, this process is less effective because only the outer wall structure contains important information, not the whole binary mask.

To address this, we introduce a highly effective method to incorporate boundary information. We observe that the most influential information contained in the boundary is the corner points of the boundary. The corner points can be described by the $(x,y)$ coordinates and hence provide better inductive bias than a plain binary mask. Therefore, we first extract the set of corner points from the boundary mask. 
Then, for each corner point, we introduce a higher-order boundary factor that connects to all the variable nodes within the factor graph model. Specifically, let the boundary $b = \{(x, y)_b^k\;|\;k \in \{1, 2, \dots n_b\}\}$ where $n_b$ is the number of corner points extracted from the given boundary mask.
Then for $k^{th}$ corner point, we define a higher-order factor $f_{b}^k = \{x_{min}^i, x_{max}^i, y_{min}^i, y_{max}^i | i\in\{1\dots N\}\}$ which connects to all variables in the factor graph. 

We encode the information describing the location and surroundings of the corner point as an input feature $\fb_b^k$ of the corner-point factor $f_b^k$. 
Specifically, for each corner point $b^k$, a feature vector is formed and initialized with three types of information: the $(x,y)$ coordinates of the corner point, the distance $d$ of the corner point from the bounding box enclosing the boundary, and the binary boundary $mask$ values surrounding the corner point at a small offset of $\epsilon$. The surrounding values indicate the direction of the corner point enclosing the inside of the house. This vector is used as an input feature for each higher-order corner-point factor.
\begin{gather*}
\fb_b^k=\big[x,\; y,\; d_{left},\; d_{right},\; d_{top},\; d_{bottom},\; mask_{(x+\epsilon,y+\epsilon)},\\
\;mask_{(x+\epsilon,y-\epsilon)},\; mask_{(x-\epsilon,y+\epsilon)},\; mask_{(x-\epsilon,y-\epsilon)}\big]_b^k
\end{gather*}

\subsubsection{Complete Factor}
In addition to the variable dependencies described above, there can be other dependencies present in the data that are not apparent. In order to enable variables to learn from such hidden dependencies, we introduce an additional factor called the \emph{complete factor}. This factor connects all the variables without any restriction with a factor feature indicating the type of the factor.

\subsection{FloorPlan Factor Graph Neural Network}
We turn our factor graph model into a neural network with learnable potential functions. Specifically, we initialize our factor graph model with variable node features $[\vb_i]_{i\in\Vcal}$ and factor node features $[\fb_c]_{c\in\Ccal}$. The variable node feature contains room-type, location, size and a 4-dim one-hot vector indicating the variable type \ie $x_{min}, x_{max}, y_{min}$ or $y_{max}$. Note that the location here is not the $(x,y)$ coordinate but the approximate relative placement indicating \emph{north, south, north-west} \etc, as discussed in Graph2Plan~\cite{hu2020graph2plan}. The factor features are initialized with the factor type (\ie box factor, relation-type, boundary or complete) and additional specific attributes for some factors (\ie The bounding box factors $f_{box}^i$ contain room-type, location, and size, and the boundary factors $f_b^k$ additionally contain the boundary feature vector as defined in Section~\ref{subsec:boundaryfactors}.

Algorithm~\ref{algo:FGNN} details our Factor Graph Neural Network (FP-FGNN) for floorplan design. It takes a Factor graph $\Gcal = (\Vcal, \Ccal, \Ecal)$ as input, with variables $\Vcal$ and factors $\Ccal$ initialized using variable features $[\vb_i]{i \in\Vcal}$ and factor features $[\fb_c]{c\in\Ccal}$. The edge set $\Ecal$ contains edges between variable and factor nodes. The edge feature $e_{ci}$ contains the features of factor $c$. 
 
 In each iteration, the variable nodes receive and aggregate messages from the factor nodes and vice versa. The message function is an MLP acting on the concatenated variable and factor embeddings from the previous iteration along with the edge feature. Furthermore, our message aggregation allows weighted aggregation of messages with the help of a learnable softmax-based aggregator defined as
\begin{equation}
 \mathrm{softmax}(\vb|\theta) = \sum_{\vb_i\in\vb}
\frac{\exp(\theta\cdot\vb_i)}{\sum_{\vb_j\in\vb}
\exp(\theta\cdot\vb_j)}\cdot\vb_{i}
\end{equation}
 This enables variables to learn from factors most relevant to them and vice versa. After a fixed number of iterations, we use an MLP to output the coordinate value of each variable. 
\begin{algorithm}[t]
	\small
	\setstretch{1.25} 
	\renewcommand{\algorithmicrequire}{\textbf{Input:}}
	\renewcommand{\algorithmicensure}{\textbf{Output:}}
	\caption{\small FloorPlan-FGNN (FP-FGNN) Algorithm }\label{algo:FGNN}
	\begin{algorithmic}[1]
		\REQUIRE $\Gcal=(\Vcal,\Ccal,\Ecal),[\vb_i]_{i \in\Vcal},[\fb_c]_{c\in\Ccal},[e_{ci}]_{(c,i) \in\Ecal}$
		\ENSURE $[v_i]_{i \in\Vcal},\; \Ical$ \qquad // \text{Box coordinates and Layout image} 
        \FOR{ $l = 1, 2, \dots, L$}
    		\STATE \textbf{Variable-to-Factor Message Passing:}
    		\STATE \quad$\tilde \fb_{ci}^l =\; \mlp_{\text{VF}}\big(\concat[\fb_c^l, \vb_i^l, e_{ci}] \big)$
    		\STATE \quad$\fb_{c}^{l+1} \;=\; \softmax\limits_{i\in N(c)} \big(\tilde \fb_{ci}^l | \theta_{\text{VF}}^l\big)$
    		
      \STATE \textbf{Factor-to-Variable Message Passing:}
            \STATE \quad$\tilde \vb_{ic}^l =\; \mlp_{\text{FV}} \big(\concat[\fb_c^l, \vb_i^l, e_{c,i}] \big)$
            \STATE \quad$\vb_{i}^{l+1} \;\;=\; \softmax\limits_{c \in N(i)} \big(\tilde \vb_{ic}^l | \theta_{\text{FV}}\big)$
        \ENDFOR
            \STATE {$v_i = \mlp(\vb_i^\text{L})$} \qquad\qquad\hspace{21pt} // Train with L1-Loss
            \STATE {$\Ical = \text{CRN-Network}\big(\;[v_i]_{i\in\Vcal}\big)$} \quad // Pixel-wise cross-entropy 
	\end{algorithmic}
\end{algorithm}

\begin{figure*}[t]
    \centering
    \begin{tabular}{l}
    \resizebox{0.8\textwidth}{!}{
    \subfloat{
        \raisebox{0.8in}{\rotatebox[origin=r]{90}{Ground Truth}}
    }
    \subfloat{
        \includegraphics[width=0.15\textwidth]{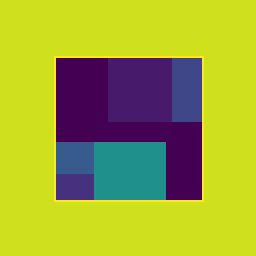}
    }
    \subfloat{
        \includegraphics[width=0.15\textwidth]{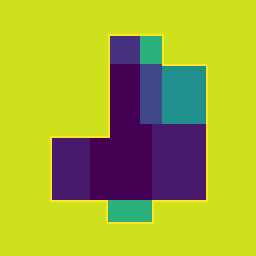}
    }
    \subfloat{
        \includegraphics[width=0.15\textwidth]{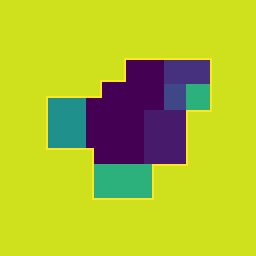}
    }
    \subfloat{
        \includegraphics[width=0.15\textwidth]{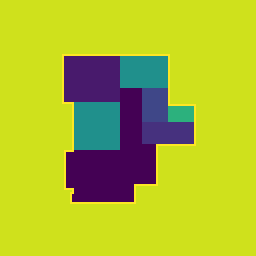}
    }
    \subfloat{
        \includegraphics[width=0.15\textwidth]{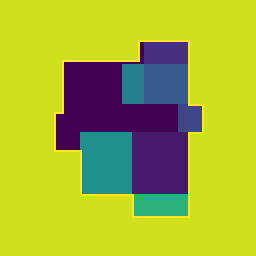}
    }
    \subfloat{
        \includegraphics[width=0.15\textwidth]{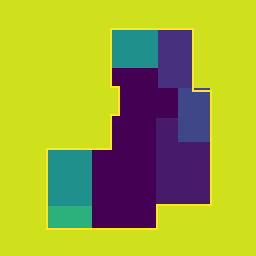}
    }}
    \\
    \resizebox{0.8\textwidth}{!}{
    \subfloat{
        \raisebox{0.8in}{\rotatebox[origin=r]{90}{Graph2Plan}}
    }\hskip -2pt
    \subfloat{
        \includegraphics[width=0.15\textwidth]{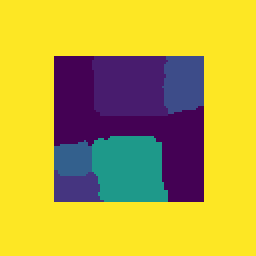}
    }
    \subfloat{
        \includegraphics[width=0.15\textwidth]{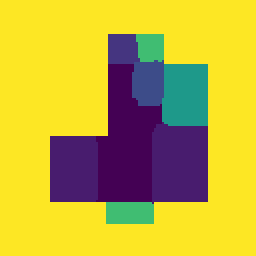}
    }
    \subfloat{
        \includegraphics[width=0.15\textwidth]{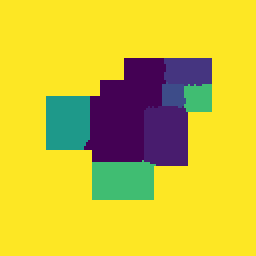}
    }
    \subfloat{
        \includegraphics[width=0.15\textwidth]{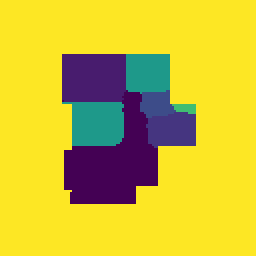}
    }
    \subfloat{
        \includegraphics[width=0.15\textwidth]{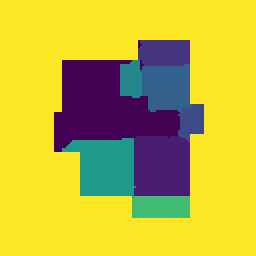}
    }
    \subfloat{
        \includegraphics[width=0.15\textwidth]{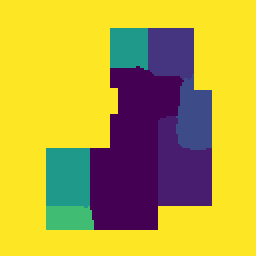}
    }}
    \\
    
    \resizebox{0.8\textwidth}{!}{
    \subfloat{
        \raisebox{0.75in}{\rotatebox[origin=r]{90}{FP-FGNN}}
    }
    \subfloat{
        \includegraphics[width=0.15\textwidth]{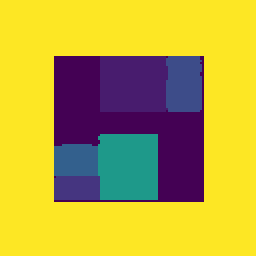}
    }
    \subfloat{
        \includegraphics[width=0.15\textwidth]{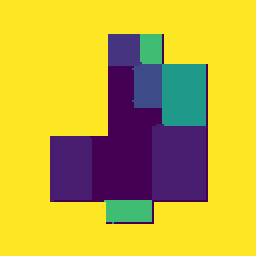}
    }
    \subfloat{
        \includegraphics[width=0.15\textwidth]{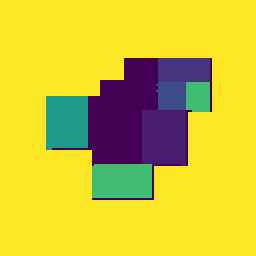}
    }
    \subfloat{
        \includegraphics[width=0.15\textwidth]{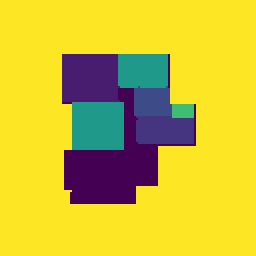}
    }
    \subfloat{
        \includegraphics[width=0.15\textwidth]{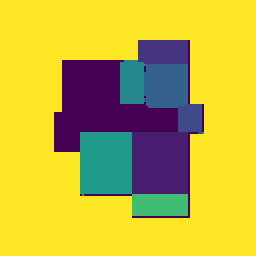}
    }
    \subfloat{
        \includegraphics[width=0.15\textwidth]{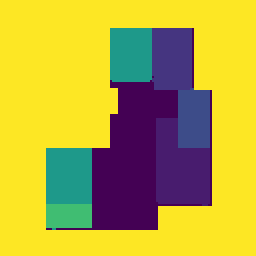}
    }}
    \end{tabular}
    \caption{\small Visualization of layout predictions of Graph2Plan and FP-FGNN without postprocessing. 
    FP-FGNN is able to produce more accurate predictions of room dimensions and the shape of walls, resulting in layouts that are more visually coherent and well-defined.}
    \label{fig:qualitative-predictions-pre}
    \vskip -0.1in
\end{figure*}

\begin{table*}
\centering
\resizebox{0.9\linewidth}{!}{
    \begin{tabular}{ c c c c c c c c}
        \toprule
        \multirow{2}{*}{\textbf{Method}} & \multicolumn{2}{c}{Box-level} & \multicolumn{3}{c}{Pixel-level} & \multicolumn{2}{c}{Constraints-level} \\
        \cmidrule(lr){2-3}  \cmidrule(lr){4-6} \cmidrule(lr){7-8}
         & IOU-Macro & IOU-Micro & Accuracy & IOU-Macro & IOU-Micro & Relation Acc & Location Acc \\
        \midrule
        \rule{0pt}{1.2EM}
        HPGM~\cite{chen2020intelligent} & 0.4701 & 0.5424 &  0.6464 &  0.4812 & 0.5179 & - & - \\
        Graph2Plan~\cite{hu2020graph2plan} & 0.6796 &  0.7402 & 0.8683 & 0.4476 & 0.7575 & 0.9358 & 0.9805 \\
        \midrule
        FP-FGNN & \textbf{0.8685} &  \textbf{0.9165} & \textbf{0.8901} & \textbf{0.5954} & \textbf{0.8019} & \textbf{0.9680} & \textbf{0.9913} \\
        \bottomrule
    \end{tabular}
    }
\caption{\small Box-level,Pixel-level and Constraints-level comparisons. We train and evaluate all three models on the same dataset.}
\label{tab:main-results}
\vskip -0.18in
\end{table*}

\subsection{Bounding boxes to floorplan image}
To translate the box coordinates to the floorplan image, we feed the predicted bounding box coordinates to a cascaded refinement network (CRN)\cite{chen2017photographic} based network in the form of multi-channel input. 
We borrow this part of the network from Graph2Plan~\cite{hu2020graph2plan}, which can produce the floorplan image given the predicted bounding boxes.

\subsection{Training}
Our network is trained with only two loss functions. The bounding box coordinate variables are trained with mean absolute loss (L1-Loss) against the ground-truth box-coordinate values and the final layout image is trained with the pixel-wise cross-entropy loss with room categories as class variables. Note that out model does not rely on additional losses, such as interior-loss, coverage-loss or mutex-loss \etc, as used in Graph2Plan~\cite{hu2020graph2plan}. 

\section{Experiments}\label{sec:experiments}

\begin{figure*}
  \centering
  \includegraphics[width=0.85\textwidth,height=7cm]{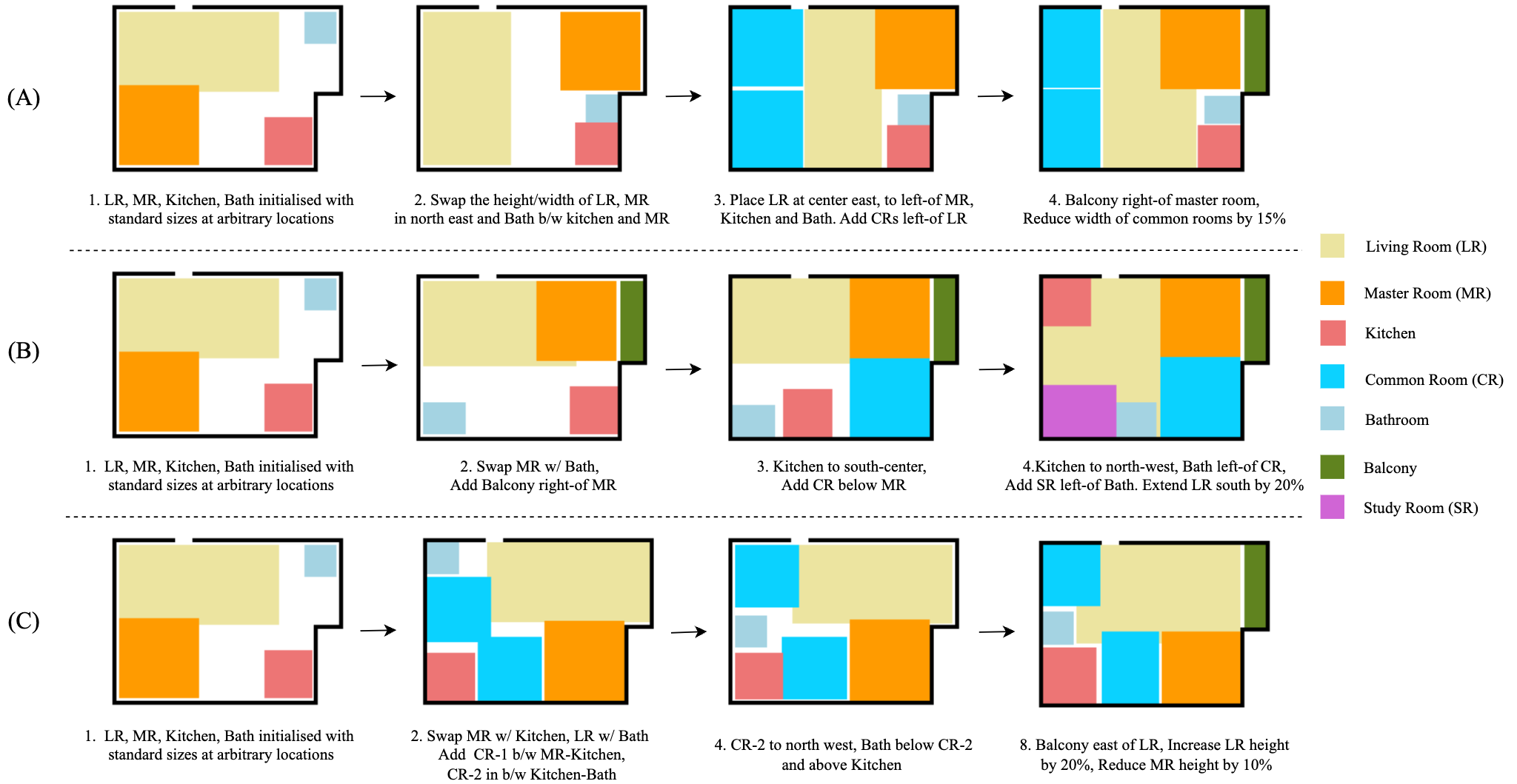}
  \caption{Illustration of iterative design with user iteration. Starting from the same arbitrary partial layout, the interaction process results in three different final layout. The room sizes are automatically adjusted for alignment with boundaries and adjacent rooms.}
  \label{fig:iterative_visualization}
 \vskip -0.15in
\end{figure*}

\subsection{Experiments Setup}
\paragraph{Dataset.}
We conduct experiments on the large-scale RPLAN dataset~\cite{wu2019data}, which consists of more than $80729$ vectorized floorplan layouts. Since our model is based on graph-constrained layout design, we use its updated version released by Graph2Plan~\cite{hu2020graph2plan}. This version contains a corresponding room-adjacency graph for each floorplan. Overall, there are $15$ room-types including \emph{`LivingRoom, MasterRoom, Kitchen, Bathroom, DiningRoom, ChildRoom, StudyRoom, SecondRoom, GuestRoom, Balcony, Entrance, Storage, Wall-in, External, ExteriorWall'} and the room-adjacency types are shown in Tab~\ref{tab:relation-factors}. Each room node comes with attributes of position and size. The method of generating room positions and size is as described in~\cite{hu2020graph2plan} and~\cite{chen2020intelligent}. Specifically, the floorplan is divided into even-sized grids and the position of the room is the grid where its centroid belongs to. The size of rooms is represented by the normalized area of the room's bounding box. For validation, we use the splits released by~\cite{hu2020graph2plan}, which contains $56511$ samples for training and $12108$ for validation/testing.

\paragraph{Evaluation.}
Unlike other floorplan design methods that build generative models and evaluate based on Diversity and Compatibility~\cite{nauata2020house,nauata2021house,sun2022wallplan,luo2022floorplangan,luo2022floorplangan}, our method focuses on accurately predicting room bounding box coordinates. We primarily use \emph{Intersection Over Union (IOU)}, both Macro/Micro, for box-level predictions, and pixel IOUs with Accuracy for pixel-level assessment. While our main emphasis is on box-level metrics for the initial stage of bounding box prediction, pixel-level metrics also offer insights into layout quality. Furthermore, we assess the preservation of constraints against ground truth, particularly focusing on the accuracy of Relations and Locations, which is central to our model's motivation.
\paragraph{Comparisons.}
Our main baseline methods for comparison are Graph2Plan~\cite{hu2020graph2plan} and HPGM~\cite{chen2020intelligent}. These models run GNN on the adjacency graph to predict bounding boxes before generating the layout. To best of our knowledge, these models are the only published methods that are predictive models and take in a graph as input and output the room bounding boxes along with the floorplan layout. Note that Graph2Plan is made generative with a separate retrieval process. 
Unlike Graph2Plan, HPGM does not take in input boundary. Therefore, for a fair comparison, we include boundary mask into HPGM model the same way as in Graph2Plan. Overall, we make sure to train and test all models with the same input data.


\subsection{Results}

\subsubsection{Comparisons with Baselines}
The results, as presented in Tab.~\ref{tab:main-results}, demonstrate that our model outperforms the baselines in terms of both box-level and pixel-level metrics, with an impressive 20\% improvement observed in box-level IOU metrics. However, we notice that the pixel-level IOU-Macro score is lower for all models, as some classes (\emph{wall-in, Entrance}) have small areas that are hard to be represented with bounding boxes which pull down the macro average since it treats all classes equally. This is supported by class-specific IOU scores provided in Appendix~\ref{sec:analysis_iou}. Box-level improvements are more pronounced than at pixel-level which further supports our methods effectiveness. The constraints-level scores further validate our model where the location of rooms in predictions matches to  ground truth for almost all rooms and preserves relations with high accuracy as well.

\subsubsection{Qualitative Results}
The benefit of FP-FGNN is equally clear in the visualized qualitative results. We first compare the direct output images of the Graph2Plan and FP-FGNN models with the ground truth images. As shown in Fig.~\ref{fig:qualitative-predictions-pre}, Graph2Plan images visibly contain uneven boundaries for most of the rooms. In contrast, FP-FGNN's output closely resembles the ground truth images with sharper internal boundaries between rooms. We believe the uneven room-boundaries in Graph2Plan is because of the higher overlap in the predicted bounding boxes compared to FP-FGNN, which translates to poor room-boundaries. We show supporting evidence for this by measuring the overlap areas in Appendix~\ref{app:overlap_analysis} and additional qualitative results in Appendix~\ref{subsec:sup_qualitative}.

\subsubsection{Iterative Design with Partial Constraints}
\label{sec:partial_input}

In the real world, the layout design process is typically iterative. Users often begin with a vague idea of their desired layout, which then evolves and becomes more refined through an iterative feedback loop. Success in this iterative design process hinges on consistently adhering to the requirements at each step. Our inferential model, characterized by its high fidelity and low inference time, is particularly helpful in such scenarios. 

\paragraph{Qualitative analysis}
Figure~\ref{fig:iterative_visualization} illustrates the iterative design process enabled by our model, demonstrating how varying user specifications lead to distinct final layouts. In practical floorplan design, the process often starts with a predetermined boundary. Our model aids users by initially setting up the layout within this boundary, starting with basic elements like the Living Room, Master Room, Kitchen, and Bathroom. These are placed at arbitrary locations in standard sizes (mean size of the room type with an aspect ratio of 1x1) and an aspect ratio of 2x1 for the Living Room. A notable feature of our model is its ability to automatically align walls with the boundary seamlessly, without explicit user input. As users modify or add constraints, our model consistently meets these evolving requirements. This consistent alignment with user inputs enables a productive feedback loop, providing users with the necessary clarity for further refinement of their requirements. Moreover, with a rapid average inference time of 0.023 seconds, our model is highly efficient for real-time user interaction.

\paragraph{Quantitative analysis}
We carry out a quantitative evaluation of our model in realistic scenarios, where users have a clear idea of the number and types of rooms but lack a detailed plan for room placement and sizes. This evaluation involves scenarios with different degrees of incomplete information. To emulate this, we randomly select certain number of rooms, and then drop their properties and relational constraints, retaining only the room type information. Note that the dropped rooms are isolated nodes in the graph. In this process, all rooms, except the Living Room, Master Room, and Kitchen, are subject to potential information omission. We train the model with different levels of partial information, and evaluate its effectiveness by measuring its ability to satisfy the constraints: those explicitly defined by the user as well as the complete set of original constraints, including those that were omitted.

\begin{table}[htbp]
\centering
\resizebox{\linewidth}{!}{
    \begin{tabular}{c l c c c c}
        \toprule 
        & \multirow{2}{*}{Constraints} & \multicolumn{4}{c}{Number of Rooms with Omitted Information (K)} \\
        \cmidrule{3-6}
        &&  $K=0$ & $K=1$ & $K=2$ & $K=3$\\
        \midrule
        \multirow{2}{*}{\rotatebox{90}{Ours}} & Relation(ours) & 96.80 & 96.57/94.00 & 95.46/87.75 & 94.69/80.40\\
        & Location(ours) & 99.13 & 99.23/98.26 & 99.07/95.47 & 99.05/91.79\\ 
        \midrule
        \multirow{2}{*}{\rotatebox{90}{G2P}} & Relation(G2P) & 93.58 & 77.35/74.72 & 78.79/75.86 & 78.10/70.44 \\
        & Location(G2P) & 98.05 & 94.58/81.04 & 91.26/76.34 & 91.80/75.79\\
        \bottomrule
    \end{tabular}
}
\caption{\small Accuracy of constraints preserved under partial inputs. We measure Relations and Locations, correctly captured in the output layout, both for specified constraints (first term) and complete set of constraints including omitted ones (second term)}
\label{tab:ierative_design}
\end{table}

\begin{wrapfigure}{r}{0.55\linewidth}
\begin{center}
\setlength{\belowcaptionskip} {12pt}
\setlength{\abovecaptionskip} {-16pt}
\vspace{-1.5em}
\includegraphics[width=\linewidth]{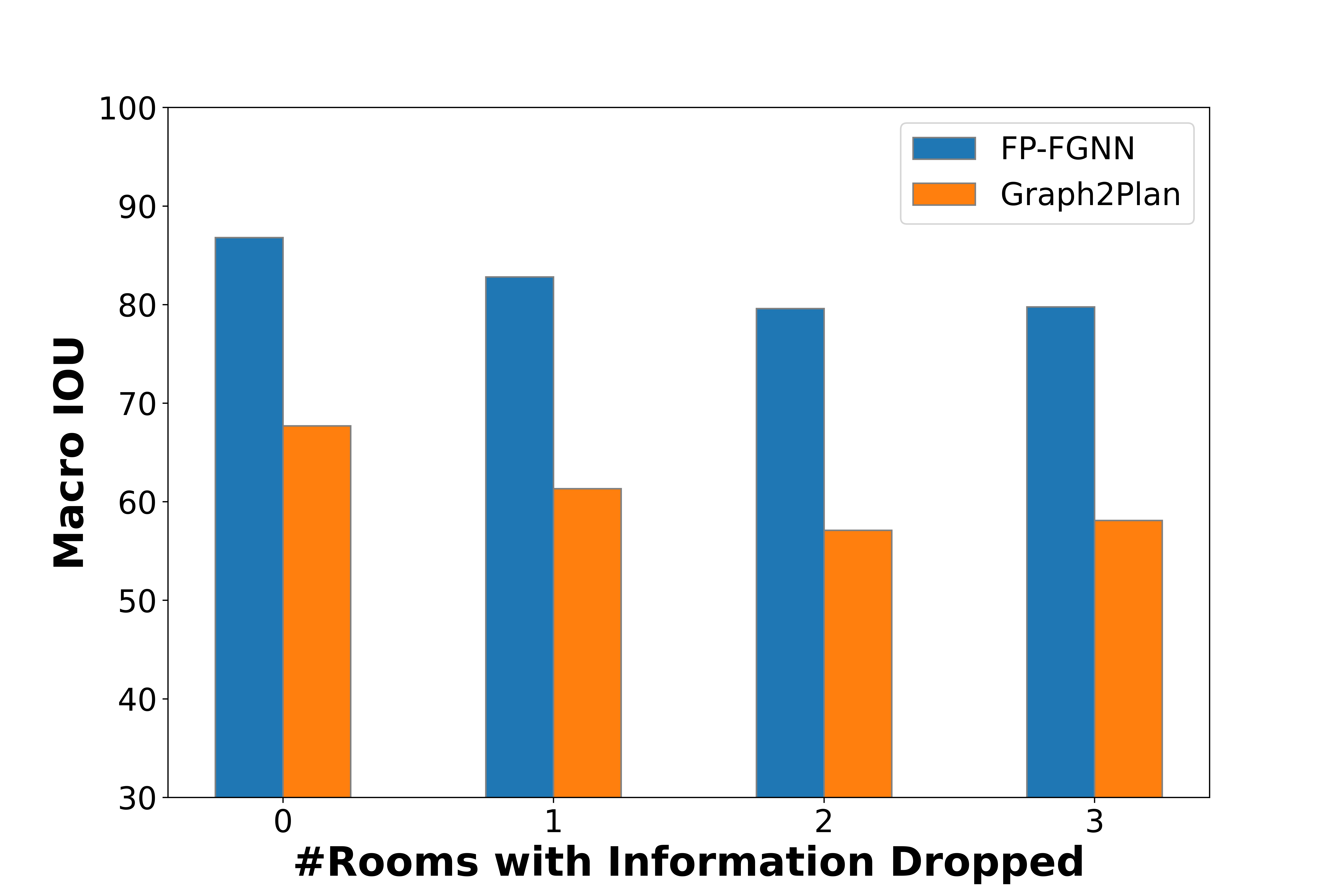}
\end{center}
\vspace{-1em}
   \caption{\small Macro IOU on specified rooms under partial information.}
\label{fig:response}
\vskip -0.1in
\end{wrapfigure}
Tab.~\ref{tab:ierative_design} shows that the preservation of both relation and location constraints for specified rooms is highly effective.  It also reveals that the overall accuracy, accounting for dropped constraints, remains stable, despite an increase in the number of dropped rooms. This contrasts sharply with the baseline Graph2Plan model, which exhibits a significant decline in constraint accuracy under similar conditions. 

This pattern is also reflected in macro IOU scores (Fig.~\ref{fig:response}). Notably, our model sustains robust performance, even when information for half of the rooms in the dataset (with an average of 6.79 rooms) is dropped.

\subsubsection{Generative Model}
\begin{figure}[t]
    \centering
    \includegraphics[width=\columnwidth]{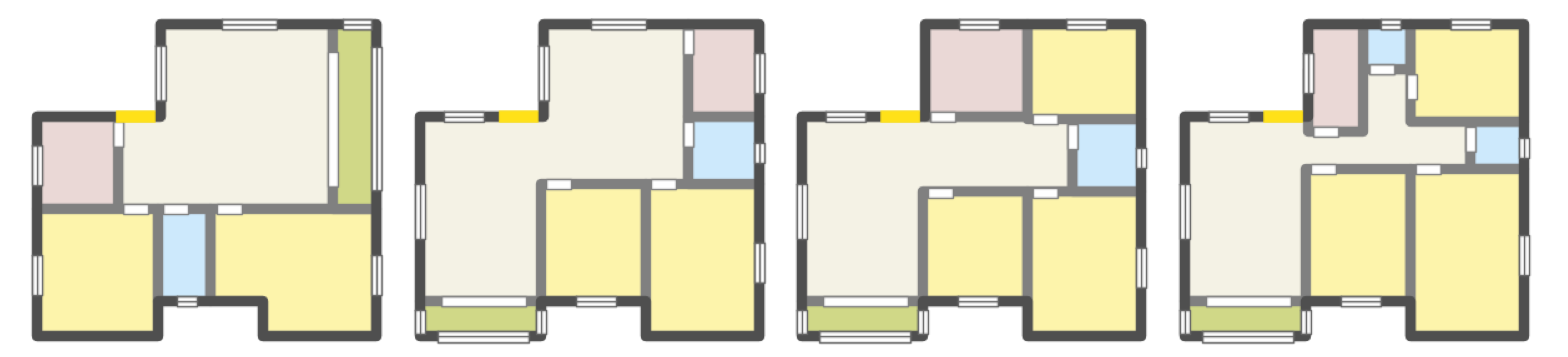}
    \caption{\footnotesize FP-FGNN's diverse generations with same input boundary.}
    \label{fig:qual_generation}
\end{figure}

In another practical scenario, users find it helpful if they are given with a choice of multiple floorplans to choose from. This use case requires a generative model which can generate diverse floorplans given the same input boundary. Although FP-FGNN is a predictive model, it can be easily used as part of a generative pipeline for generating diverse layouts. To demonstrate this, we use Graph2Plan~\citep{hu2020graph2plan} methodology for the generative mechanism. In Graph2Plan approach, given an input boundary, we first retrieve its $k$ nearest neighbour boundaries from the training dataset. Then we extract the layout graphs from those examples and use it along with the given input boundary as input to the predictive FP-FGNN model. Fig.~\ref{fig:qual_generation} shows the diversity in generations of our model given the same input boundary. 
Furthermore, we quantitatively measure the generated images with the FID~\citep{heusel2017gans} score and compare with baselines  Graph2Plan~\citep{hu2020graph2plan}, RPLAN~\citep{wu2019data}, Constr-LGen~\citep{para2021generative}. Table~\ref{tab:fid} shows that FP-FGNN generation significantly outperforms the baselines.

\begin{table}[t]
\vspace{-5pt}
\centering
     \resizebox{0.9\columnwidth}{!}{
        \begin{tabular}{ c c c c c}
            \toprule
             \textbf{Method} & Graph2Plan & RPLAN & Constr-LGen  & FP-FGNN\\
             \midrule
             FID &  29.26 & 21.29 & 21.47  & \textbf{12.87}\\ 
            \bottomrule
        \end{tabular}
         }
    \caption{{\footnotesize FID score comparison of FP-FGNN's generation}}
    \label{tab:fid}
    \vspace{-7pt}
\end{table}

\section{Conclusion}
In this paper, we address the problem of generating layouts under spatial constraints, particularly in the context of floorplan design. We propose a novel approach that employs factor graphs to capture the pairwise and higher-order dependencies between the latent variables representing the rooms' bounding boxes and spatial attributes. Our method employs message-passing on the bipartite factor graph, forming a factor graph neural network that effectively leverages the available domain knowledge to produce floorplans that meet the desired specification. Our results demonstrate a significant improvement in IOU scores over existing methods. 
%
%
In future work, we aim to extend FP-FGNN to learn to generate  floorplan layouts from language descriptions~\citep{leng2023tell2design}.
\paragraph{Acknowledgements:} This research is supported in part by the National Research Foundation (NRF), Singapore and DSO National Laboratories under the AI Singapore Program (No. AISG2-RP-2020-016).

{
    \small
    \bibliographystyle{ieeenat_fullname}
    \bibliography{main}

\begin{thebibliography}{38}
\providecommand{\natexlab}[1]{#1}
\providecommand{\url}[1]{\texttt{#1}}
\expandafter\ifx\csname urlstyle\endcsname\relax
  \providecommand{\doi}[1]{doi: #1}\else
  \providecommand{\doi}{doi: \begingroup \urlstyle{rm}\Url}\fi

\bibitem[Arroyo et~al.(2021)Arroyo, Postels, and
  Tombari]{arroyo2021variational}
Diego~Martin Arroyo, Janis Postels, and Federico Tombari.
\newblock Variational transformer networks for layout generation.
\newblock In \emph{Proceedings of the IEEE/CVF Conference on Computer Vision
  and Pattern Recognition}, pages 13642--13652, 2021.

\bibitem[Cao et~al.(2022)Cao, Ma, Zhou, Liu, Xie, Ge, and
  Jiang]{cao2022geometry}
Yunning Cao, Ye Ma, Min Zhou, Chuanbin Liu, Hongtao Xie, Tiezheng Ge, and
  Yuning Jiang.
\newblock Geometry aligned variational transformer for image-conditioned layout
  generation.
\newblock In \emph{In Proc. of ACM Multimedia}, 2022.

\bibitem[Chang et~al.(2021)Chang, Cheng, Luo, Murata, Nourbakhsh, and
  Tsuji]{chang2021building}
Kai-Hung Chang, Chin-Yi Cheng, Jieliang Luo, Shingo Murata, Mehdi Nourbakhsh,
  and Yoshito Tsuji.
\newblock Building-gan: Graph-conditioned architectural volumetric design
  generation.
\newblock In \emph{Proceedings of the IEEE/CVF International Conference on
  Computer Vision}, pages 11956--11965, 2021.

\bibitem[Chen and Koltun(2017)]{chen2017photographic}
Qifeng Chen and Vladlen Koltun.
\newblock Photographic image synthesis with cascaded refinement networks.
\newblock In \emph{Proceedings of the IEEE international conference on computer
  vision}, pages 1511--1520, 2017.

\bibitem[Chen et~al.(2020)Chen, Wu, Tang, Wang, Wang, and
  Tan]{chen2020intelligent}
Qi Chen, Qi Wu, Rui Tang, Yuhan Wang, Shuai Wang, and Mingkui Tan.
\newblock Intelligent home 3d: Automatic 3d-house design from linguistic
  descriptions only.
\newblock In \emph{Proceedings of the IEEE/CVF Conference on Computer Vision
  and Pattern Recognition}, pages 12625--12634, 2020.

\bibitem[Deka et~al.(2017)Deka, Huang, Franzen, Hibschman, Afergan, Li,
  Nichols, and Kumar]{deka2017rico}
Biplab Deka, Zifeng Huang, Chad Franzen, Joshua Hibschman, Daniel Afergan, Yang
  Li, Jeffrey Nichols, and Ranjitha Kumar.
\newblock Rico: A mobile app dataset for building data-driven design
  applications.
\newblock In \emph{In Proc. of UIST}, 2017.

\bibitem[Du et~al.(2020)Du, Li, and Mordatch]{du2020compositional}
Yilun Du, Shuang Li, and Igor Mordatch.
\newblock Compositional visual generation with energy based models.
\newblock \emph{Advances in Neural Information Processing Systems},
  33:\penalty0 6637--6647, 2020.

\bibitem[Dupty and Lee(2020)]{dupty2020neuralizing}
Mohammed~Haroon Dupty and Wee~Sun Lee.
\newblock Neuralizing efficient higher-order belief propagation.
\newblock \emph{arXiv preprint arXiv:2010.09283}, 2020.

\bibitem[He et~al.(2022)He, Huang, and Wang]{he2022iplan}
Feixiang He, Yanlong Huang, and He Wang.
\newblock iplan: interactive and procedural layout planning.
\newblock In \emph{Proceedings of the IEEE/CVF Conference on Computer Vision
  and Pattern Recognition}, pages 7793--7802, 2022.

\bibitem[Heusel et~al.(2017)Heusel, Ramsauer, Unterthiner, Nessler, and
  Hochreiter]{heusel2017gans}
Martin Heusel, Hubert Ramsauer, Thomas Unterthiner, Bernhard Nessler, and Sepp
  Hochreiter.
\newblock Gans trained by a two time-scale update rule converge to a local nash
  equilibrium.
\newblock \emph{Advances in neural information processing systems}, 30, 2017.

\bibitem[Hu et~al.(2020)Hu, Huang, Tang, Van~Kaick, Zhang, and
  Huang]{hu2020graph2plan}
Ruizhen Hu, Zeyu Huang, Yuhan Tang, Oliver Van~Kaick, Hao Zhang, and Hui Huang.
\newblock Graph2plan: Learning floorplan generation from layout graphs.
\newblock \emph{ACM Transactions on Graphics (TOG)}, 39\penalty0 (4):\penalty0
  118--1, 2020.

\bibitem[Kikuchi et~al.(2021)Kikuchi, Simo-Serra, Otani, and
  Yamaguchi]{kikuchi2021constrained}
Kotaro Kikuchi, Edgar Simo-Serra, Mayu Otani, and Kota Yamaguchi.
\newblock Constrained graphic layout generation via latent optimization.
\newblock In \emph{Proceedings of the 29th ACM International Conference on
  Multimedia}, pages 88--96, 2021.

\bibitem[Kingma and Ba(2014)]{kingma2014adam}
Diederik~P Kingma and Jimmy Ba.
\newblock Adam: A method for stochastic optimization.
\newblock \emph{arXiv preprint arXiv:1412.6980}, 2014.

\bibitem[Laignel et~al.(2021)Laignel, Pozin, Geffrier, Delevaux, Brun, and
  Dolla]{laignel2021floor}
Graziella Laignel, Nicolas Pozin, Xavier Geffrier, Loukas Delevaux, Florian
  Brun, and Bastien Dolla.
\newblock Floor plan generation through a mixed constraint programming-genetic
  optimization approach.
\newblock \emph{Automation in Construction}, 123:\penalty0 103491, 2021.

\bibitem[Lee et~al.(2020)Lee, Jiang, Essa, Le, Gong, Yang, and
  Yang]{lee2020neural}
Hsin-Ying Lee, Lu Jiang, Irfan Essa, Phuong~B Le, Haifeng Gong, Ming-Hsuan
  Yang, and Weilong Yang.
\newblock Neural design network: Graphic layout generation with constraints.
\newblock In \emph{Computer Vision--ECCV 2020: 16th European Conference,
  Glasgow, UK, August 23--28, 2020, Proceedings, Part III 16}, pages 491--506.
  Springer, 2020.

\bibitem[Leng et~al.(2023)Leng, Zhou, Dupty, Lee, Joyce, and
  Lu]{leng2023tell2design}
Sicong Leng, Yang Zhou, Mohammed~Haroon Dupty, Wee~Sun Lee, Sam Joyce, and Wei
  Lu.
\newblock Tell2design: A dataset for language-guided floor plan generation.
\newblock In \emph{Proceedings of the 61st Annual Meeting of the Association
  for Computational Linguistics (Volume 1: Long Papers)}, pages 14680--14697,
  2023.

\bibitem[Li et~al.(2019)Li, Yang, Hertzmann, Zhang, and Xu]{li2019layoutgan}
Jianan Li, Jimei Yang, Aaron Hertzmann, Jianming Zhang, and Tingfa Xu.
\newblock Layoutgan: Generating graphic layouts with wireframe discriminators.
\newblock \emph{arXiv preprint arXiv:1901.06767}, 2019.

\bibitem[Luo and Huang(2022)]{luo2022floorplangan}
Ziniu Luo and Weixin Huang.
\newblock Floorplangan: Vector residential floorplan adversarial generation.
\newblock \emph{Automation in Construction}, 142:\penalty0 104470, 2022.

\bibitem[Merrell et~al.(2010)Merrell, Schkufza, and
  Koltun]{merrell2010computer}
Paul Merrell, Eric Schkufza, and Vladlen Koltun.
\newblock Computer-generated residential building layouts.
\newblock In \emph{ACM SIGGRAPH Asia 2010 papers}, pages 1--12. 2010.

\bibitem[Nauata et~al.(2020)Nauata, Chang, Cheng, Mori, and
  Furukawa]{nauata2020house}
Nelson Nauata, Kai-Hung Chang, Chin-Yi Cheng, Greg Mori, and Yasutaka Furukawa.
\newblock House-gan: Relational generative adversarial networks for
  graph-constrained house layout generation.
\newblock In \emph{Computer Vision--ECCV 2020: 16th European Conference,
  Glasgow, UK, August 23--28, 2020, Proceedings, Part I 16}, pages 162--177.
  Springer, 2020.

\bibitem[Nauata et~al.(2021)Nauata, Hosseini, Chang, Chu, Cheng, and
  Furukawa]{nauata2021house}
Nelson Nauata, Sepidehsadat Hosseini, Kai-Hung Chang, Hang Chu, Chin-Yi Cheng,
  and Yasutaka Furukawa.
\newblock House-gan++: Generative adversarial layout refinement network towards
  intelligent computational agent for professional architects.
\newblock In \emph{Proceedings of the IEEE/CVF Conference on Computer Vision
  and Pattern Recognition}, pages 13632--13641, 2021.

\bibitem[{\"O}zerol and ARSLAN~SEL{\c{C}}UK(2021)]{ozerol2021generation}
Gizem {\"O}zerol and SEMRA ARSLAN~SEL{\c{C}}UK.
\newblock Generation of architectural drawings through generative adversarial
  networks (gans): A case on apartment plan layouts.
\newblock \emph{Architectural Research Current Studies and Future Trends},
  page~35, 2021.

\bibitem[Para et~al.(2021)Para, Guerrero, Kelly, Guibas, and
  Wonka]{para2021generative}
Wamiq Para, Paul Guerrero, Tom Kelly, Leonidas~J Guibas, and Peter Wonka.
\newblock Generative layout modeling using constraint graphs.
\newblock In \emph{Proceedings of the IEEE/CVF International Conference on
  Computer Vision}, pages 6690--6700, 2021.

\bibitem[Paszke et~al.(2017)Paszke, Gross, Chintala, Chanan, Yang, DeVito, Lin,
  Desmaison, Antiga, and Lerer]{paszke2017automatic}
Adam Paszke, Sam Gross, Soumith Chintala, Gregory Chanan, Edward Yang, Zachary
  DeVito, Zeming Lin, Alban Desmaison, Luca Antiga, and Adam Lerer.
\newblock Automatic differentiation in pytorch.
\newblock 2017.

\bibitem[Rodrigues et~al.(2013{\natexlab{a}})Rodrigues, Gaspar, and
  Gomes]{rodrigues2013evolutionarya}
Eug{\'e}nio Rodrigues, Ad{\'e}lio~Rodrigues Gaspar, and {\'A}lvaro Gomes.
\newblock An evolutionary strategy enhanced with a local search technique for
  the space allocation problem in architecture, part 1: Methodology.
\newblock \emph{Computer-Aided Design}, 45\penalty0 (5):\penalty0 887--897,
  2013{\natexlab{a}}.

\bibitem[Rodrigues et~al.(2013{\natexlab{b}})Rodrigues, Gaspar, and
  Gomes]{rodrigues2013evolutionaryb}
Eug{\'e}nio Rodrigues, Ad{\'e}lio~Rodrigues Gaspar, and {\'A}lvaro Gomes.
\newblock An evolutionary strategy enhanced with a local search technique for
  the space allocation problem in architecture, part 2: Validation and
  performance tests.
\newblock \emph{Computer-Aided Design}, 45\penalty0 (5):\penalty0 898--910,
  2013{\natexlab{b}}.

\bibitem[Rosser et~al.(2017)Rosser, Smith, and Morley]{rosser2017data}
Julian~F Rosser, Gavin Smith, and Jeremy~G Morley.
\newblock Data-driven estimation of building interior plans.
\newblock \emph{International Journal of Geographical Information Science},
  31\penalty0 (8):\penalty0 1652--1674, 2017.

\bibitem[Satorras and Welling(2021)]{satorras2021neural}
Victor~Garcia Satorras and Max Welling.
\newblock Neural enhanced belief propagation on factor graphs.
\newblock In \emph{International Conference on Artificial Intelligence and
  Statistics}, pages 685--693. PMLR, 2021.

\bibitem[Shabani et~al.(2022)Shabani, Hosseini, and
  Furukawa]{shabani2022housediffusion}
Mohammad~Amin Shabani, Sepidehsadat Hosseini, and Yasutaka Furukawa.
\newblock Housediffusion: Vector floorplan generation via a diffusion model
  with discrete and continuous denoising.
\newblock \emph{arXiv preprint arXiv:2211.13287}, 2022.

\bibitem[Sodhi et~al.(2022)Sodhi, Dexheimer, Mukadam, Anderson, and
  Kaess]{sodhi2022leo}
Paloma Sodhi, Eric Dexheimer, Mustafa Mukadam, Stuart Anderson, and Michael
  Kaess.
\newblock Leo: Learning energy-based models in factor graph optimization.
\newblock In \emph{Conference on Robot Learning}, pages 234--244. PMLR, 2022.

\bibitem[Sun et~al.(2022)Sun, Wu, Liu, Min, Zhang, and Zheng]{sun2022wallplan}
Jiahui Sun, Wenming Wu, Ligang Liu, Wenjie Min, Gaofeng Zhang, and Liping
  Zheng.
\newblock Wallplan: synthesizing floorplans by learning to generate wall
  graphs.
\newblock \emph{ACM Transactions on Graphics (TOG)}, 41\penalty0 (4):\penalty0
  1--14, 2022.

\bibitem[Upadhyay et~al.(2023)Upadhyay, Dubey, Mani~Kuriakose, and
  Agarawal]{upadhyay2023floorgan}
Abhinav Upadhyay, Alpana Dubey, Suma Mani~Kuriakose, and Shaurya Agarawal.
\newblock Floorgan: Generative network for automated floor layout generation.
\newblock In \emph{Proceedings of the 6th Joint International Conference on
  Data Science \& Management of Data (10th ACM IKDD CODS and 28th COMAD)},
  pages 140--148, 2023.

\bibitem[Wu et~al.(2019)Wu, Fu, Tang, Wang, Qi, and Liu]{wu2019data}
Wenming Wu, Xiao-Ming Fu, Rui Tang, Yuhan Wang, Yu-Hao Qi, and Ligang Liu.
\newblock Data-driven interior plan generation for residential buildings.
\newblock \emph{ACM Transactions on Graphics (TOG)}, 38\penalty0 (6):\penalty0
  1--12, 2019.

\bibitem[Xiao et~al.(2013)Xiao, Owens, and Torralba]{xiao2013sun3d}
Jianxiong Xiao, Andrew Owens, and Antonio Torralba.
\newblock Sun3d: A database of big spaces reconstructed using sfm and object
  labels.
\newblock In \emph{In Proc. of ICCV}, 2013.

\bibitem[Yoon et~al.(2019)Yoon, Liao, Xiong, Zhang, Fetaya, Urtasun, Zemel, and
  Pitkow]{yoon2019inference}
KiJung Yoon, Renjie Liao, Yuwen Xiong, Lisa Zhang, Ethan Fetaya, Raquel
  Urtasun, Richard Zemel, and Xaq Pitkow.
\newblock Inference in probabilistic graphical models by graph neural networks.
\newblock In \emph{2019 53rd Asilomar Conference on Signals, Systems, and
  Computers}, pages 868--875. IEEE, 2019.

\bibitem[Zhang et~al.(2020)Zhang, Wu, and Lee]{zhang2020factor}
Zhen Zhang, Fan Wu, and Wee~Sun Lee.
\newblock Factor graph neural networks.
\newblock \emph{Advances in Neural Information Processing Systems},
  33:\penalty0 8577--8587, 2020.

\bibitem[Zhang et~al.(2023)Zhang, Dupty, Wu, Shi, and Lee]{JMLR:v24:21-0434}
Zhen Zhang, Mohammed~Haroon Dupty, Fan Wu, Javen~Qinfeng Shi, and Wee~Sun Lee.
\newblock Factor graph neural networks.
\newblock \emph{Journal of Machine Learning Research}, 24\penalty0
  (181):\penalty0 1--54, 2023.

\bibitem[Zhong et~al.(2019)Zhong, Tang, and Yepes]{zhong2019publaynet}
Xu Zhong, Jianbin Tang, and Antonio~Jimeno Yepes.
\newblock Publaynet: largest dataset ever for document layout analysis.
\newblock In \emph{In Proc. of ICDAR}, 2019.

\end{thebibliography}
}

\clearpage
\maketitlesupplementary
\setcounter{page}{1}
\setcounter{section}{0}

\section{Additional Experiments}
\subsection{Analysis of Class-wise IOU scores}
\label{sec:analysis_iou}
In Tab.~\ref{tab:main-results}, the pixel-level IOU scores with macro averaging appear to be relatively low when compared to micro averaged scores. In this section, we investigate the reason for this discrepancy in scores. Macro IOU scores are computed by first calculating the class-wise IOU scores separately before averaging across each class. Therefore, we list the IOU scores for each class separately and compare them for both Graph2Plan and FP-FGNN.
\begin{table*}[bp]
\centering
    \resizebox{\textwidth}{!}{
        \begin{tabular}{ c c c c c c c c c }
            \toprule
             \textbf{Roomtype} & LivingRoom & MasterRoom & Kitchen & Bathroom & DiningRoom & ChildRoom & StudyRoom & SecondRoom \\
            \midrule
            \addlinespace
             Graph2Plan & 0.8593 & 0.8400 & 0.7164 & 0.6870 & 0.2018 & 0.4577 & 0.7559 & 0.8214\\
             FP-FGNN & 0.8470 & 0.9062 & 0.8013 & 0.7721 & 0.5743 & 0.8029 & 0.8487 & 0.8868\\    
             \bottomrule
             \addlinespace
             \toprule
             \textbf{Roomtype} &  GuestRoom & Balcony & Entrance & Storage & Wall-in & External & ExteriorWall \\
            \midrule
            \addlinespace
             Graph2Plan & 0.1433 & 0.7594 & 0.0000 & 0.2862 & 0.1303 & 0.9987 & 0.0000\\
             FP-FGNN & 0.4193 & 0.8686 & 0.0000 & 0.6066 & 0.3822 & 0.9987 & 0.0000\\
             \bottomrule
        \end{tabular}
        }
    \caption{\small IOU scores per each class for Graph2Plan and FGNN.}
    \label{tab:ablation-macro}
\end{table*}
\begin{table*}[b]
\centering
        \begin{tabular}{ c c c c}
            \toprule
             \textbf{Method} & Graph2Plan & FP-FGNN & GT \\
             \midrule
             Average overlap &  112.47 & 94.84 & 95.33  \\
            \bottomrule
        \end{tabular}
    \caption{\small Average area of intersection between all pairs of predicted boxes for Graph2Plan and FGNN.}
    \label{tab:ablation-intersectionboxes}    
\end{table*}

The RPLAN dataset has a total of $15$ classes. Tab.~\ref{tab:ablation-macro} shows the results of class-wise IOU scores for each of the $15$ classes in the dataset. Firstly, for almost all the classes, FP-FGNN scores improve on Graph2Plan scores by a significant margin. The relative improvement is especially large in \emph{DiningRoom, ChildRoom, GuestRoom} and \emph{Storage}. However, for some of the classes like \emph{Wall-in, Entrance} and \emph{Exterior Wall}, the IOU scores are quite low and seems to be the primary reason for overall lower IOU-Macro score relative to IOU-Micro, since all classes are equally weighted in Macro averaging. It is worth noting that these rooms \ie \emph{Wall-in, Entrance} and \emph{Exterior Wall} are mostly irregular \ie not easily modeled with a bounding box and and hence, it is not surprising that the scores for these classes are lower. Additionally, bounding box for classes like \emph{Wall-in} are usually too thin and hence, the scores are likely to be zero if the predicted box is not aligned to the same position as ground truth.

\subsection{Analysis of Room overlap}
\label{app:overlap_analysis}
One notable difference between the predicted floorplans generated by Graph2Plan and FP-FGNN is the appearance of the internal boundaries between the rooms. This is clearly visible in Fig.~\ref{fig:qualitative-predictions-pre} which shows that the internal boundaries between rooms are uneven in the floorplans generated by Graph2Plan. In contrast, the floorplans generated by FP-FGNN contain straight boundaries which closely resemble the ground truth. We seek to investigate the reason for this difference, since both methods use the same CRN-based approach in generating the floorplan image from the set of bounding boxes.

One plausible reason for this difference is the amount of overlap between the predicted bounding boxes of each model. Ideally, the overlap between predicted bounding boxes should be minimal, although there may be some overlap between ground truth bounding boxes due to the non-rectangular shapes of rooms. Measuring the average overlap between predicted bounding boxes provides some information about the internal boundaries of the predicted rooms.

In Fig.~\ref{tab:ablation-intersectionboxes}, we show the area of intersection between each pair of predicted bounding boxes. Firstly, the average overlap area in the Ground Truth bounding boxes is $95.33$. The GT box overlap area is significant since living room boxes would overlap with other rooms. FP-FGNN's average overlap area of $94.84$ is similar to GT overlap and is significantly less than $112.4$ of Graph2Plan, a reduction of around $18\%$. This supports the observation of clear and straight boundaries in floorplans generated by FP-FGNN. This reduction is likely to contribute to the clear and straight boundaries observed in the floorplans generated by FP-FGNN.

\begin{table*}[!tbp]
\centering
     \resizebox{0.65\textwidth}{!}{
        \begin{tabular}{ l c c c c }
            \toprule
            \multirow{2}{*}{\textbf{Method}} & \multicolumn{2}{c}{Box-level} & \multicolumn{2}{c}{Pixel-level} \\
            \cmidrule{2-3} \cmidrule{4-5}
             & IOU - Macro & IOU - Micro & Accuracy & IOU-Micro \\
             \midrule
             FP-FGNN & \textbf{0.8685} &  \textbf{0.9165} & \textbf{0.8900} & \textbf{0.8017} \\
            \midrule
            w/o box factors & 0.7300 & 0.8182 &  0.8180 &  0.6923\\
            w/o relation factors &  0.8180 & 0.8789 &0.8645 & 0.7609\\
            w/o boundary factors & 0.6163 & 0.7149 &0.7903 &0.6488\\
            w/o complete factor & 0.7841 &0.8476 & 0.8473 & 0.7351\\    
             \bottomrule
        \end{tabular}
     }
    \caption{\small Ablation study on the type of factors.}
    \label{tab:ablation_factor}
\end{table*}

\begin{table*}[!tbp]
\centering
     \resizebox{0.65\textwidth}{!}{
        \begin{tabular}{ l c c c c }
            \toprule
            \multirow{2}{*}{\textbf{Method}} & \multicolumn{2}{c}{Box-level} & \multicolumn{2}{c}{Pixel-level} \\
            \cmidrule{2-3} \cmidrule{4-5}
             & IOU - Macro & IOU - Micro & Accuracy & IOU-Micro \\
            \midrule
            FP-FGNN & \textbf{0.8685} &  \textbf{0.9165} & \textbf{0.8900} & \textbf{0.8017} \\
            \midrule
            w/o \emph{inside, surrounding} & 0.8481 & 0.9018 & 0.8790 & 0.7840\\
            w/o \emph{left, right} & 0.8517 &0.9051 & 0.8838 &0.7918\\
            w/o \emph{above, below} & 0.8412 &0.8973 & 0.8804 &0.7862\\
            \multirow{2}{*}{\shortstack[c]{ w/o \emph{leftabove, leftbelow},\\ \;\;\emph{rightabove, rightbelow}}}
            &\multirow{2}{*}{0.8308} & \multirow{2}{*}{0.8898} & \multirow{2}{*}{0.8710} & \multirow{2}{*}{0.7710}\\ 
            \\
            \bottomrule
        \end{tabular}
     }
    \caption{\small Ablation study on different relation factors.}
    \label{tab:ablation_relation}
\end{table*}
\begin{table*}[!tbp]
\centering
     \resizebox{0.65\textwidth}{!}{
        \begin{tabular}{ l c c c c }
            \toprule
            \multirow{2}{*}{\textbf{Method}} & \multicolumn{2}{c}{Box-level} & \multicolumn{2}{c}{Pixel-level} \\
            \cmidrule{2-3} \cmidrule{4-5}
             & IOU - Macro & IOU - Micro & Accuracy & IOU-Micro \\
             \midrule
             FP-FGNN & \textbf{0.8685} &  \textbf{0.9165} & \textbf{0.8900} & \textbf{0.8017} \\
            \midrule
            w/o distance feature & 0.8411 & 0.8976 &  0.8831 &  0.7907\\
            w/o surrounding feature & 0.8333 &  0.8918 & 0.8806 & 0.7867\\
             \bottomrule
        \end{tabular}
     }
    \caption{\small Effectiveness of the boundary factor features.}
    \label{tab:ablation_boundary}
\end{table*}
\begin{table*}[!thbp]
\centering
     \resizebox{0.65\textwidth}{!}{
        \begin{tabular}{ c c c c c }
            \toprule
            \multirow{2}{*}{\textbf{Method}} & \multicolumn{2}{c}{Box-level} & \multicolumn{2}{c}{Pixel-level} \\
            \cmidrule{2-3} \cmidrule{4-5}
             & IOU - Macro & IOU - Micro & Accuracy & IOU-Micro \\
            \midrule
            FP-FGNN(w/ Softmax) & \textbf{0.8685} &  \textbf{0.9165} & \textbf{0.8900} & \textbf{0.8017} \\
            \midrule
            w/ MAX & 0.8436 & 0.8995 &  0.8739 &  0.7760\\
            w/ SUM & 0.7514 &  0.8272 & 0.8479 & 0.7358\\
            w/ MEAN & 0.8238 &  0.8848 & 0.8751 & 0.7778\\
           \bottomrule
        \end{tabular}
     }
    \caption{\small Effectiveness of the message aggregation function.}
    \label{tab:ablation_aggr}
\end{table*}

\subsubsection{Ablation Studies}
\label{subsec:sup_ablation_analysis}
\paragraph{Different types of factors.}
\label{sec:ablation_factor}
In order to gain a better understanding of how our model functions, we 
study the importance of each type of factor. 
As shown in Tab.~\ref{tab:ablation_factor}, we find that (1) every type of factor is critical and that (2) higher-order factors are more important than pairwise relation factors. Furthermore, we discover that (3) boundary factors, which capture structural constraints, are especially important. Without these boundary factors, our performance measurement, which is based on IOU-macro, experienced a significant 25\% drop.

\paragraph{Different types of relations.}
To further understand our model's functioning, we conduct an ablation study on different types of relation factors. To do this, we grouped together some of the factors that capture the same constraints, such as inside and surrounding relations. Tab.~\ref{tab:ablation_relation} shows that capturing relative positional information using these factors is highly effective. Additionally, we find that each group of constraints contributes to the overall performance of our model. These findings demonstrate the importance of carefully considering the various types of relation factors and their ability to capture critical constraints when designing and refining models for specific tasks.

\paragraph{Features of boundary factors.}
In earlier ablation studies, we discussed the critical role played by boundary factors in our model's performance, underscoring the importance of meeting structural constraints. Hence, we further investigate the different types of features we designed to capture the properties of each corner point, i.e., the distance between each corner point and the outer bounding box that encloses the boundary, and the binary boundary mask surrounding the corner point at a small offset of $\epsilon$. As shown in Tab.~\ref{tab:ablation_boundary}, both features contribute to the effectiveness of the model. However, the distance feature is relatively more informative.
\paragraph{Message aggregation function.}
A crucial aspect of our message passing scheme is the message aggregator, which allows for weighted aggregation of messages using a learnable softmax-based aggregator. This function enables variables to learn from the subset of factors that are most relevant to them. To demonstrate the effectiveness of this approach, we compared our solution with three other aggregation functions: MAX, SUM, and MEAN. Our results, presented in Table~\ref{tab:ablation_aggr}, show that the MAX aggregator is the most effective among the three. Intuitively, MAX acts as a selection function compared to MEAN and SUM, which further validates the motivation behind our design for the softmax aggregator.

\subsection{Iterative Design }
\begin{figure*}
    \centering
    \includegraphics[width=0.9\textwidth,height=7in]{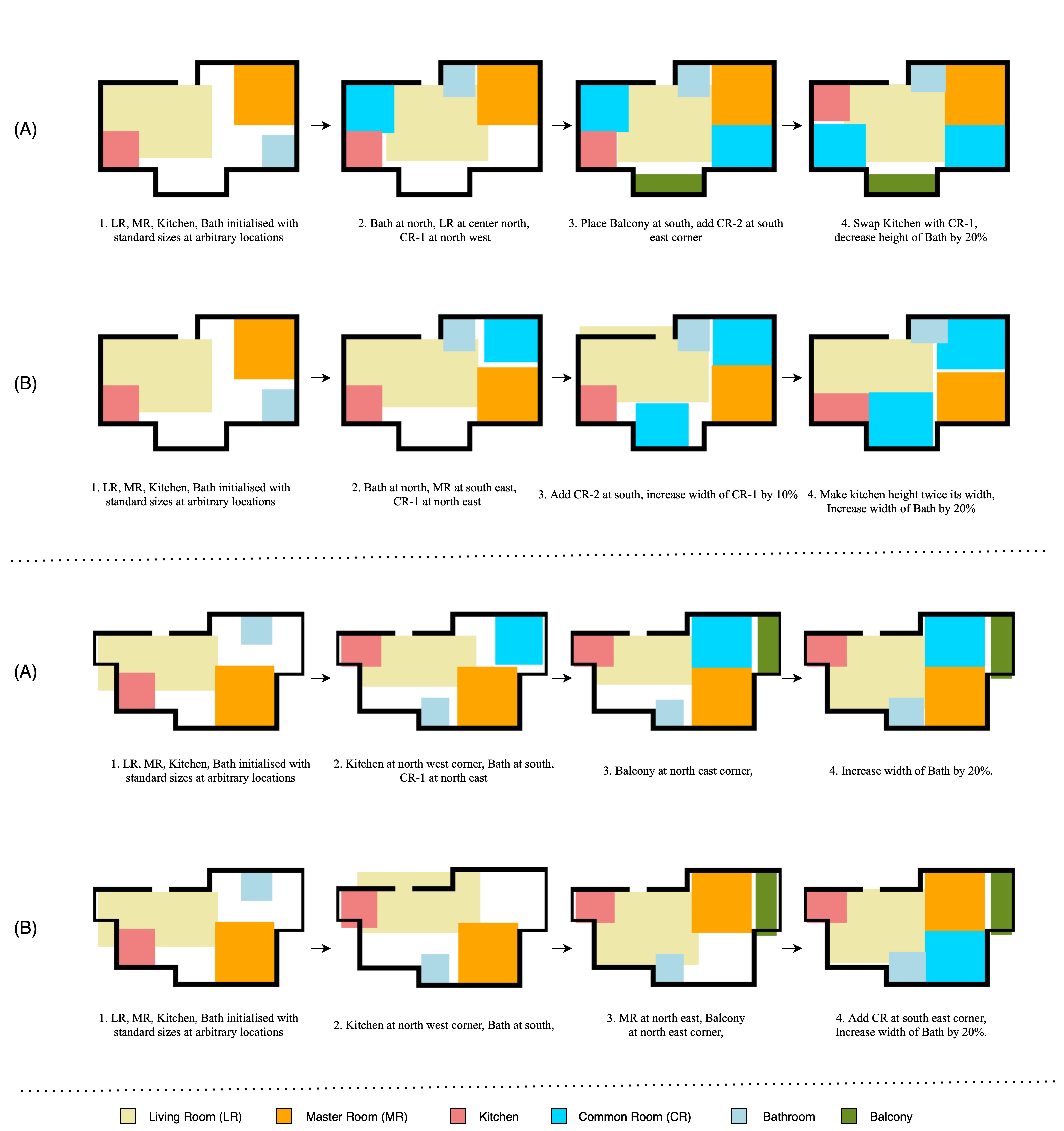}
    \caption{Additional examples showing iterative design process with FP-FGNN}
    \label{fig:supl_iterative}
\end{figure*}

We present additional illustrations of the iterative design process with FP-FGNN. Fig.~\ref{fig:supl_iterative} shows examples where 
a given boundary is initialized with Living room, Master room, Kitchen and Bathroom at arbitrary locations. The floorplans are updated according to the user inputs leading to design of different floorplan layouts from the same boundary. The high fidelity of the predictions along with the good alignment with the boundary given only partial inputs benefits the iterative design process.

\subsection{Additional qualitative Results}
\label{subsec:sup_qualitative}

\begin{figure*}[t]
    \centering
    \resizebox{0.8\textwidth}{!}{
    \begin{tabular}{ c }
    \subfloat{
        \raisebox{1.25in}{\rotatebox[origin=r]{90}{\Large Ground Truth}}
    }\hspace{5pt}
    \begin{subfigure}{0.24\textwidth}
        \includegraphics[width=\textwidth]{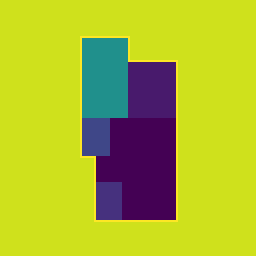}
    \end{subfigure}
    \begin{subfigure}{0.24\textwidth}
        \includegraphics[width=\textwidth]{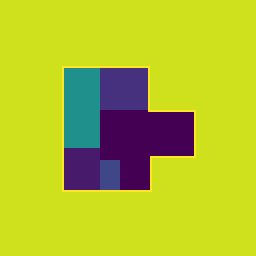}
    \end{subfigure}
    \begin{subfigure}{0.24\textwidth}
        \includegraphics[width=\textwidth]{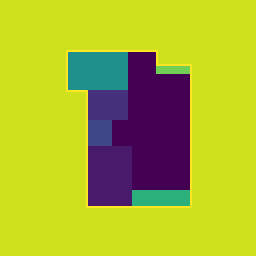}
    \end{subfigure}
    \begin{subfigure}{0.24\textwidth}
        \includegraphics[width=\textwidth]{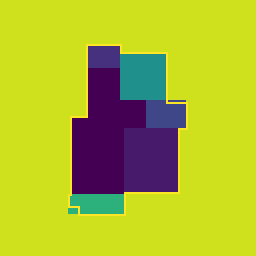}
    \end{subfigure}
    \\
    \subfloat{
        \raisebox{1.15in}{\rotatebox[origin=r]{90}{\Large FP-FGNN}}
    }\hfill
    \begin{subfigure}{0.24\textwidth}
        \includegraphics[width=\textwidth]{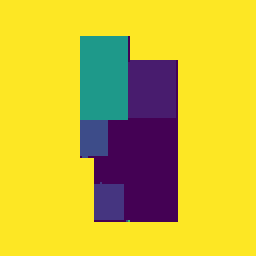}
    \end{subfigure}
    \begin{subfigure}{0.24\textwidth}
        \includegraphics[width=\textwidth]{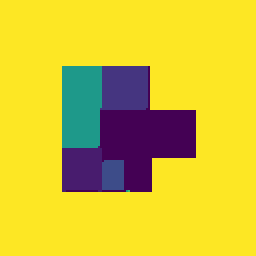}
    \end{subfigure}
    \begin{subfigure}{0.24\textwidth}
        \includegraphics[width=\textwidth]{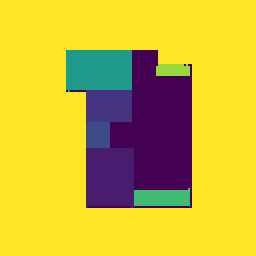}
    \end{subfigure}
    \begin{subfigure}{0.24\textwidth}
        \includegraphics[width=\textwidth]{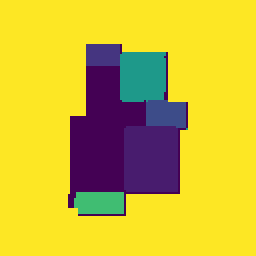}
    \end{subfigure}
    \\
    \subfloat{
        \raisebox{1.25in}{\rotatebox[origin=r]{90}{\Large Ground Truth}}
    }\hfill
    \begin{subfigure}{0.24\textwidth}
        \includegraphics[width=\textwidth]{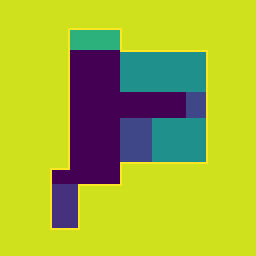}
    \end{subfigure}
    \begin{subfigure}{0.24\textwidth}
        \includegraphics[width=\textwidth]{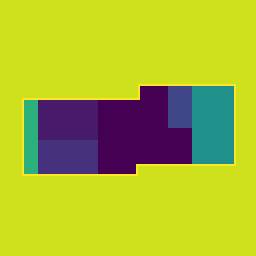}
    \end{subfigure}
    \begin{subfigure}{0.24\textwidth}
        \includegraphics[width=\textwidth]{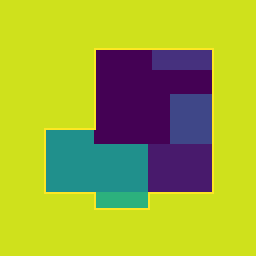}
    \end{subfigure}
    \begin{subfigure}{0.24\textwidth}
        \includegraphics[width=\textwidth]{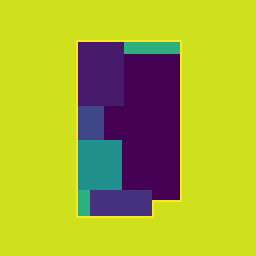}
    \end{subfigure}
    \\
    \subfloat{
        \raisebox{1.15in}{\rotatebox[origin=r]{90}{\Large FP-FGNN}}
    }\hfill
    \begin{subfigure}{0.24\textwidth}
        \includegraphics[width=\textwidth]{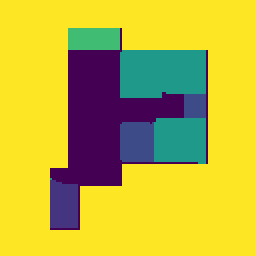}
    \end{subfigure}
    \begin{subfigure}{0.24\textwidth}
        \includegraphics[width=\textwidth]{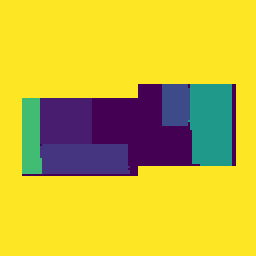}
    \end{subfigure}
    \begin{subfigure}{0.24\textwidth}
        \includegraphics[width=\textwidth]{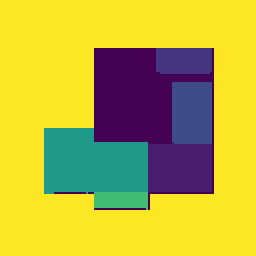}
    \end{subfigure}
    \begin{subfigure}{0.24\textwidth}
        \includegraphics[width=\textwidth]{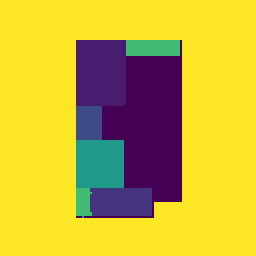}
    \end{subfigure}
    \\
    \subfloat{
        \raisebox{1.25in}{\rotatebox[origin=r]{90}{\Large Ground Truth}}
    }\hfill
    \begin{subfigure}{0.24\textwidth}
        \includegraphics[width=\textwidth]{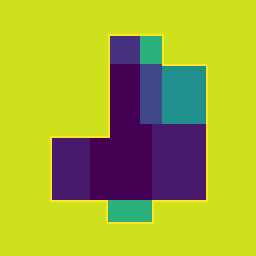}
    \end{subfigure}
    \begin{subfigure}{0.24\textwidth}
        \includegraphics[width=\textwidth]{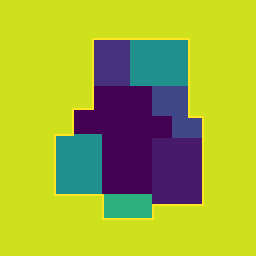}
    \end{subfigure}
    \begin{subfigure}{0.24\textwidth}
        \includegraphics[width=\textwidth]{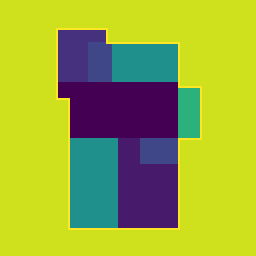}
    \end{subfigure}
    \begin{subfigure}{0.24\textwidth}
        \includegraphics[width=\textwidth]{Figures/qualitative-2/FGNN_12_GT.png}
    \end{subfigure}
    \\
    \subfloat{
        \raisebox{1.15in}{\rotatebox[origin=r]{90}{\Large FP-FGNN}}
    }\hfill
    \begin{subfigure}{0.24\textwidth}
        \includegraphics[width=\textwidth]{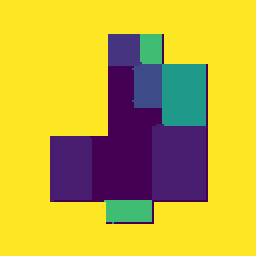}
    \end{subfigure}
    \begin{subfigure}{0.24\textwidth}
        \includegraphics[width=\textwidth]{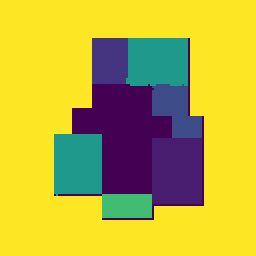}
    \end{subfigure}
    \begin{subfigure}{0.24\textwidth}
        \includegraphics[width=\textwidth]{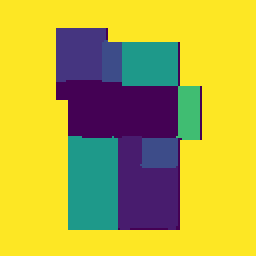}
    \end{subfigure}
    \begin{subfigure}{0.24\textwidth}
        \includegraphics[width=\textwidth]{Figures/qualitative-2/FGNN_12.png}
    \end{subfigure}
    \end{tabular}
    }
    \caption{\small Comparison of floorplans generated by FP-FGNN with the ground truth.}
    \label{fig:qualitative-predictions-im-1}
    \vskip -0.1in
\end{figure*}

\begin{figure*}[t]
    \centering
    \resizebox{0.8\textwidth}{!}{
    \begin{tabular}{ c }
    \subfloat{
        \raisebox{1.25in}{\rotatebox[origin=r]{90}{\Large Ground Truth}}
    }\hspace{5pt}
    \begin{subfigure}{0.24\textwidth}
    
        \includegraphics[width=\textwidth]{Figures/qualitative-2/FGNN_13_GT.png}
    \end{subfigure}
    \begin{subfigure}{0.24\textwidth}
        \includegraphics[width=\textwidth]{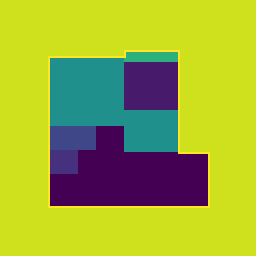}
    \end{subfigure}
    \begin{subfigure}{0.24\textwidth}
        \includegraphics[width=\textwidth]{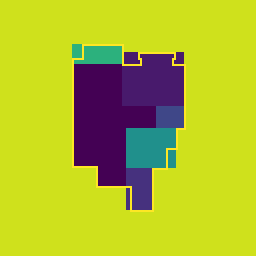}
    \end{subfigure}
    \begin{subfigure}{0.24\textwidth}
        \includegraphics[width=\textwidth]{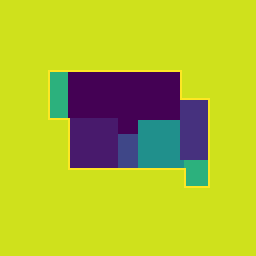}
    \end{subfigure}\\
    \subfloat{
        \raisebox{1.15in}{\rotatebox[origin=r]{90}{\Large FP-FGNN}}
    }\hfill
    \begin{subfigure}{0.24\textwidth}
        \includegraphics[width=\textwidth]{Figures/qualitative-2/FGNN_13.png}
    \end{subfigure}
    \begin{subfigure}{0.24\textwidth}
        \includegraphics[width=\textwidth]{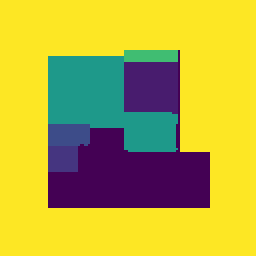}
    \end{subfigure}
    \begin{subfigure}{0.24\textwidth}
        \includegraphics[width=\textwidth]{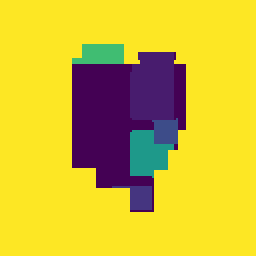}
    \end{subfigure}
    \begin{subfigure}{0.24\textwidth}
        \includegraphics[width=\textwidth]{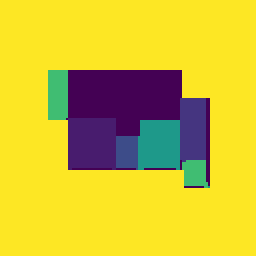}
    \end{subfigure}\\
    %
    %
    \subfloat{
        \raisebox{1.25in}{\rotatebox[origin=r]{90}{\Large Ground Truth}}
   }\hfill
    \begin{subfigure}{0.24\textwidth}
        \includegraphics[width=\textwidth]{Figures/qualitative-2/FGNN_17_GT.png}
    \end{subfigure}
    \begin{subfigure}{0.24\textwidth}
        \includegraphics[width=\textwidth]{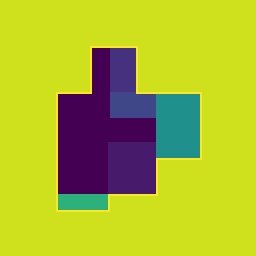}
    \end{subfigure}
    \begin{subfigure}{0.24\textwidth}
        \includegraphics[width=\textwidth]{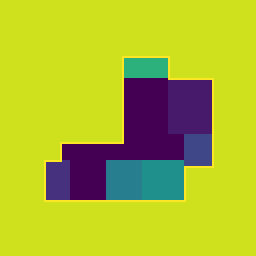}
    \end{subfigure}
    \begin{subfigure}{0.24\textwidth}
        \includegraphics[width=\textwidth]{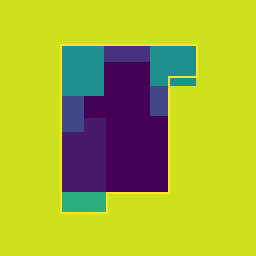}
    \end{subfigure}\\
    \subfloat{
        \raisebox{1.15in}{\rotatebox[origin=r]{90}{\Large FP-FGNN}}
    }\hfill
    \begin{subfigure}{0.24\textwidth}
        \includegraphics[width=\textwidth]{Figures/qualitative-2/FGNN_17.png}
    \end{subfigure}
    \begin{subfigure}{0.24\textwidth}
        \includegraphics[width=\textwidth]{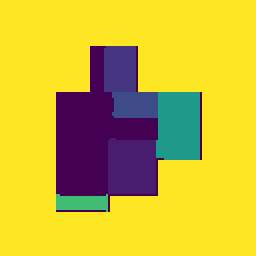}
    \end{subfigure}
    \begin{subfigure}{0.24\textwidth}
        \includegraphics[width=\textwidth]{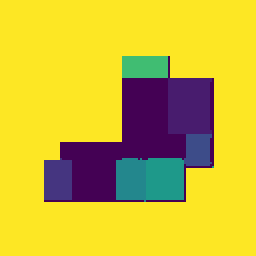}
    \end{subfigure}
    \begin{subfigure}{0.24\textwidth}
        \includegraphics[width=\textwidth]{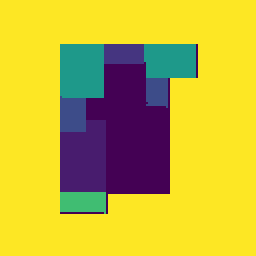}
    \end{subfigure}\\
        %
    %
    \subfloat{
        \raisebox{1.25in}{\rotatebox[origin=r]{90}{\Large Ground Truth}}
    }\hfill
    \begin{subfigure}{0.24\textwidth}
        \includegraphics[width=\textwidth]{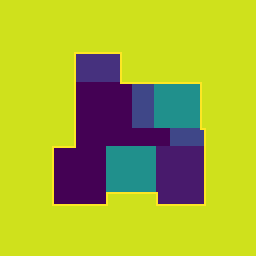}
   \end{subfigure}
    \begin{subfigure}{0.24\textwidth}
        \includegraphics[width=\textwidth]{Figures/qualitative-2/FGNN_22_GT.png}
    \end{subfigure}
    \begin{subfigure}{0.24\textwidth}
        \includegraphics[width=\textwidth]{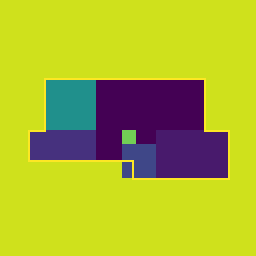}
    \end{subfigure}
    \begin{subfigure}{0.24\textwidth}
        \includegraphics[width=\textwidth]{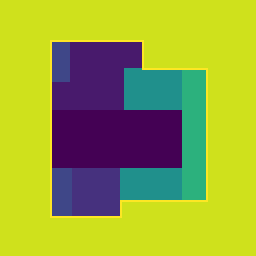}
     \end{subfigure}\\
    \subfloat{
        \raisebox{1.15in}{\rotatebox[origin=r]{90}{\Large FP-FGNN}}
    }\hfill
    \begin{subfigure}{0.24\textwidth}
        \includegraphics[width=\textwidth]{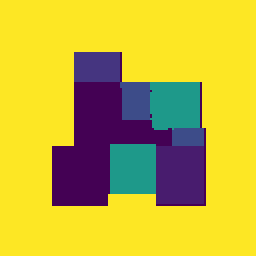}
    \end{subfigure}
    \begin{subfigure}{0.24\textwidth}
        \includegraphics[width=\textwidth]{Figures/qualitative-2/FGNN_22.png}
    \end{subfigure}
        \begin{subfigure}{0.24\textwidth}
        \includegraphics[width=\textwidth]{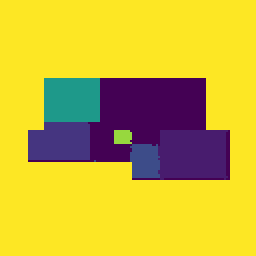}
    \end{subfigure}
    \begin{subfigure}{0.24\textwidth}
        \includegraphics[width=\textwidth]{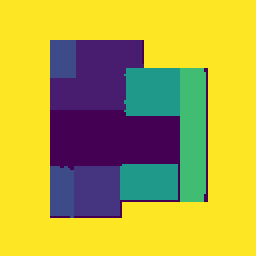}
    \end{subfigure}
    \end{tabular}
    }
    \caption{\small Comparison of floorplans generated by FP-FGNN with the ground truth. }
    \label{fig:qualitative-predictions-im-2}
    \vskip -0.1in
\end{figure*}

\begin{figure*}[t]
    \centering
    \resizebox{0.8\textwidth}{!}{
    \begin{tabular}{ c }
    \subfloat{
        \raisebox{1.25in}{\rotatebox[origin=r]{90}{\Large Ground Truth}}
    }\hspace{5pt}
    \begin{subfigure}{0.24\textwidth}
        \includegraphics[width=\textwidth]{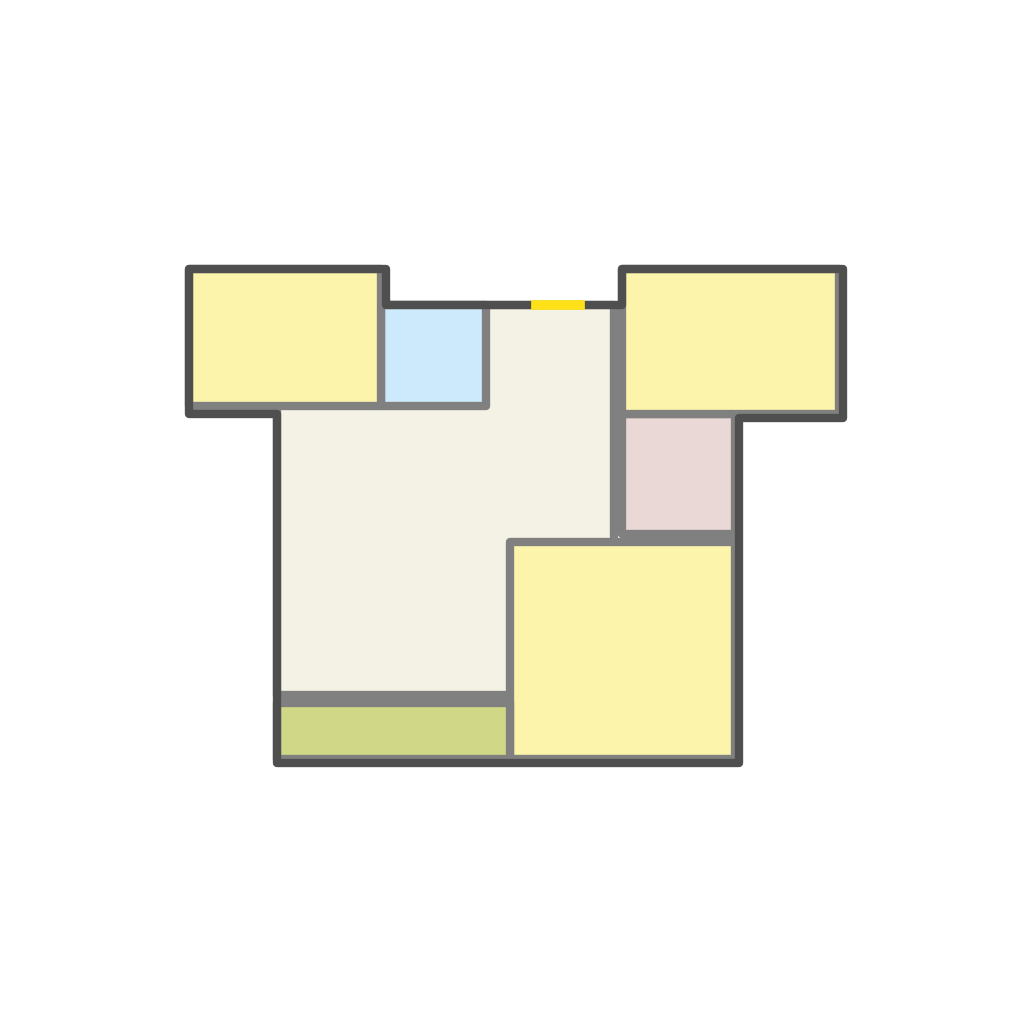}
     \end{subfigure}
    \begin{subfigure}{0.24\textwidth}
        \includegraphics[width=\textwidth]{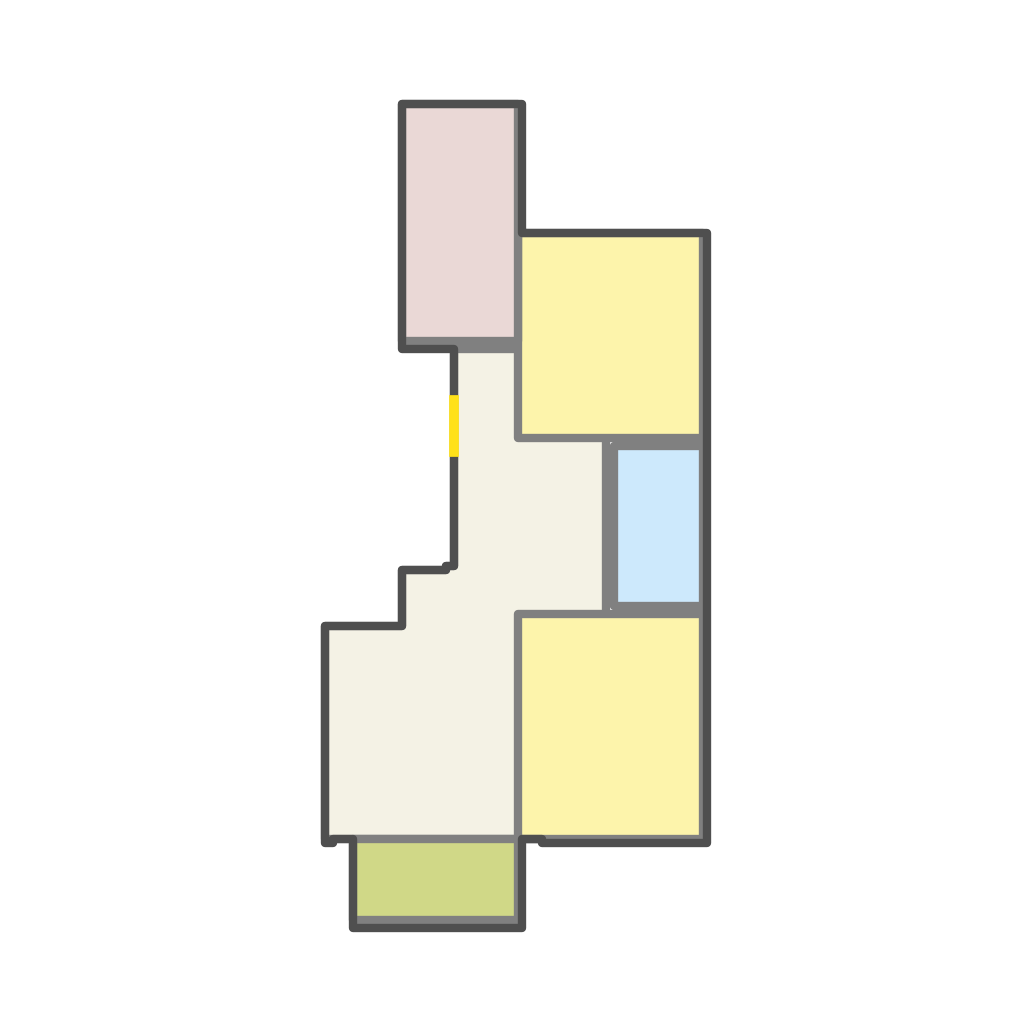}
     \end{subfigure}
    \begin{subfigure}{0.24\textwidth}
        \includegraphics[width=\textwidth]{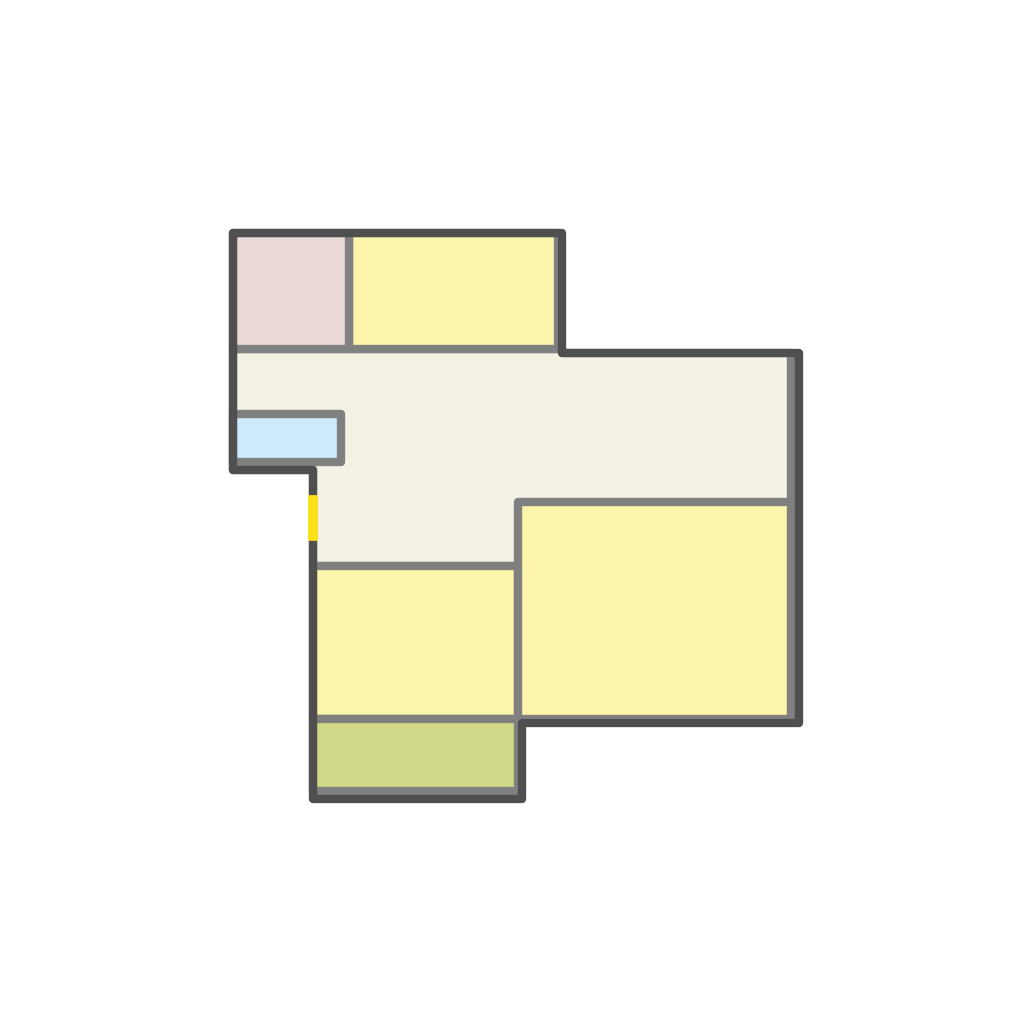}
     \end{subfigure}
    \begin{subfigure}{0.24\textwidth}
        \includegraphics[width=\textwidth]{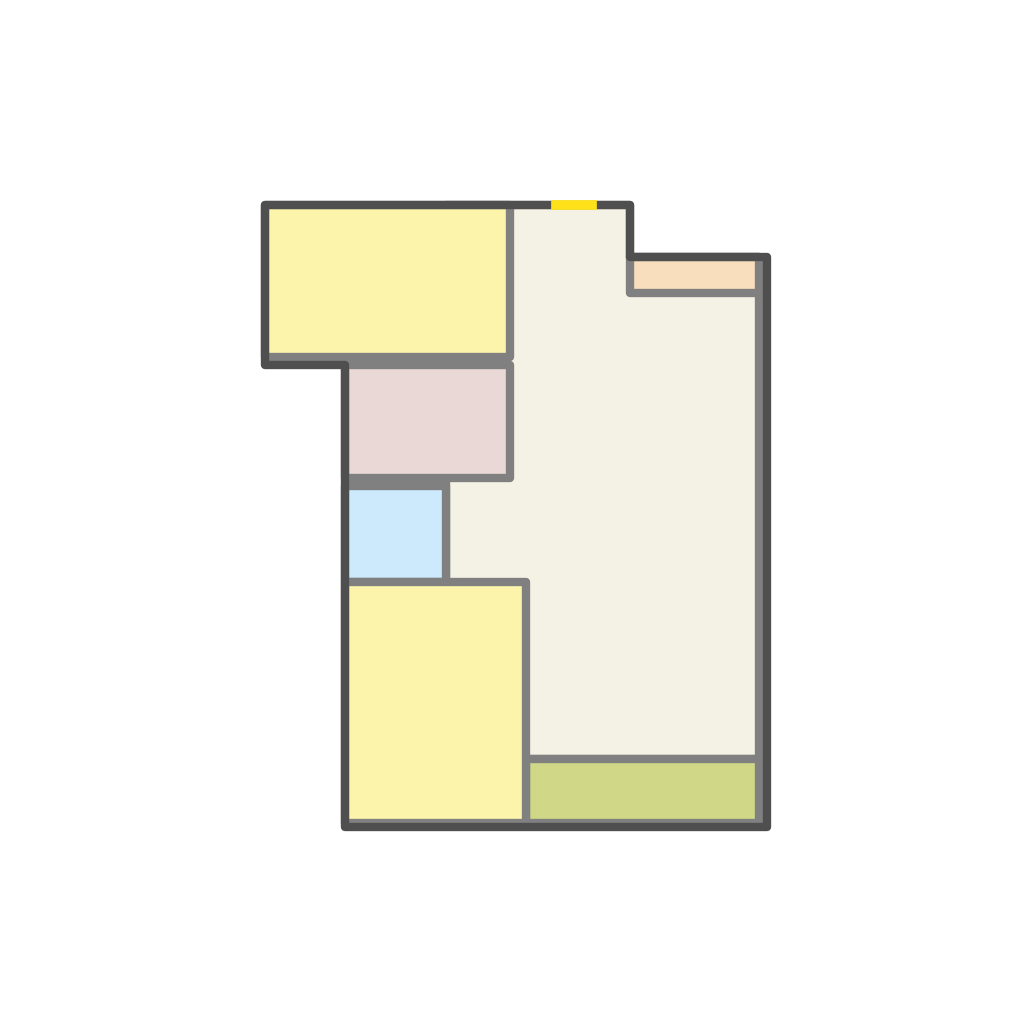}
     \end{subfigure}
     \\
    \subfloat{
        \raisebox{1.15in}{\rotatebox[origin=r]{90}{\Large FP-FGNN}}
    }\hfill
    \begin{subfigure}{0.24\textwidth}
        \includegraphics[width=\textwidth]{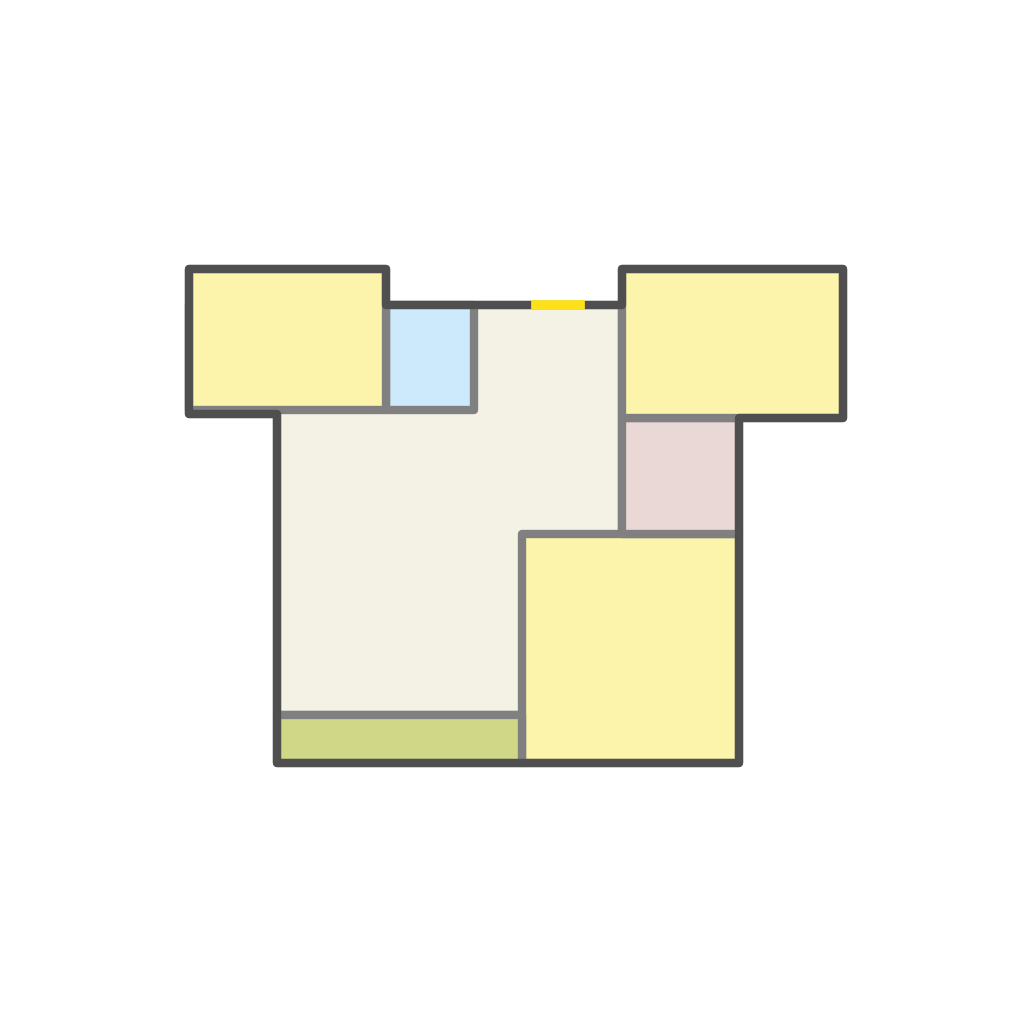}
     \end{subfigure}
    \begin{subfigure}{0.24\textwidth}
        \includegraphics[width=\textwidth]{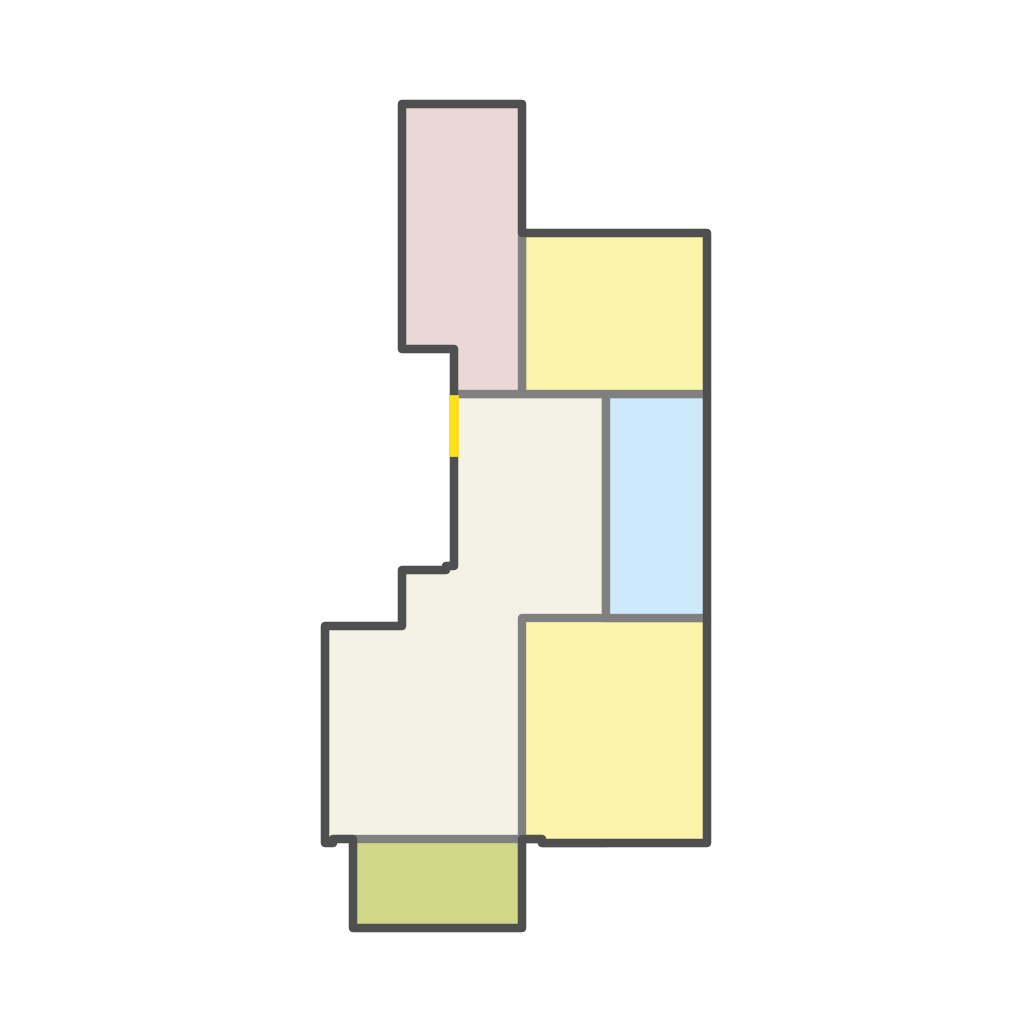}
     \end{subfigure}
    \begin{subfigure}{0.24\textwidth}
        \includegraphics[width=\textwidth]{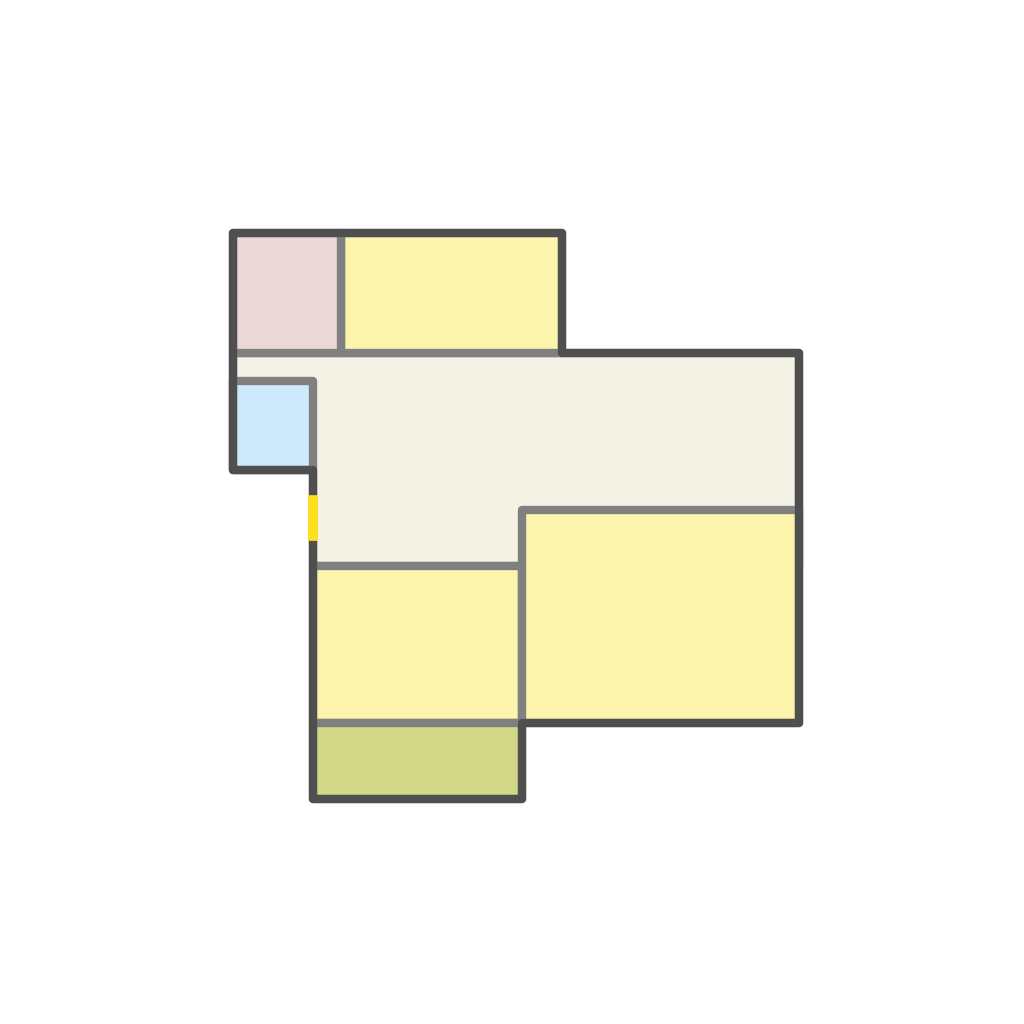}
     \end{subfigure}
    \begin{subfigure}{0.24\textwidth}
        \includegraphics[width=\textwidth]{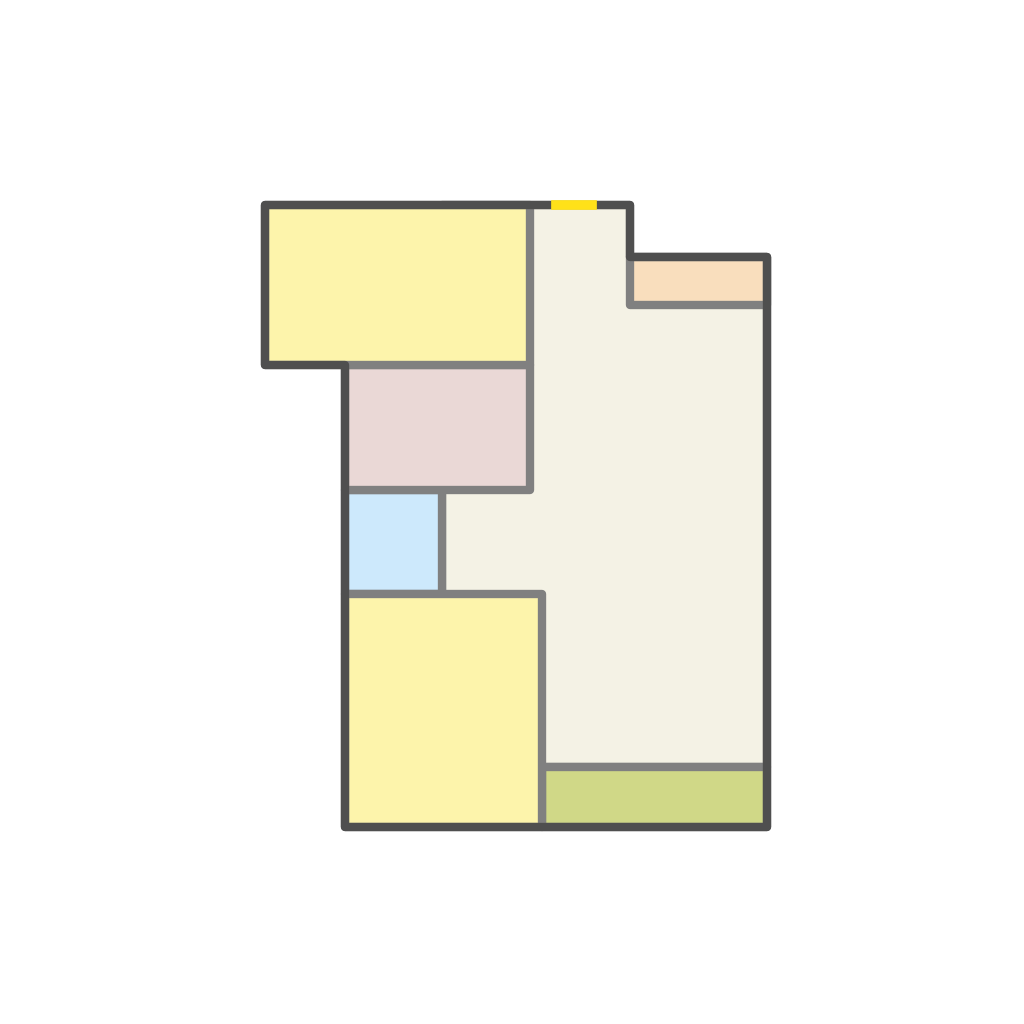}
    \end{subfigure}
    \\
    \subfloat{
        \raisebox{1.25in}{\rotatebox[origin=r]{90}{\Large Ground Truth}}
    }\hfill
    \begin{subfigure}{0.24\textwidth}
        \includegraphics[width=\textwidth]{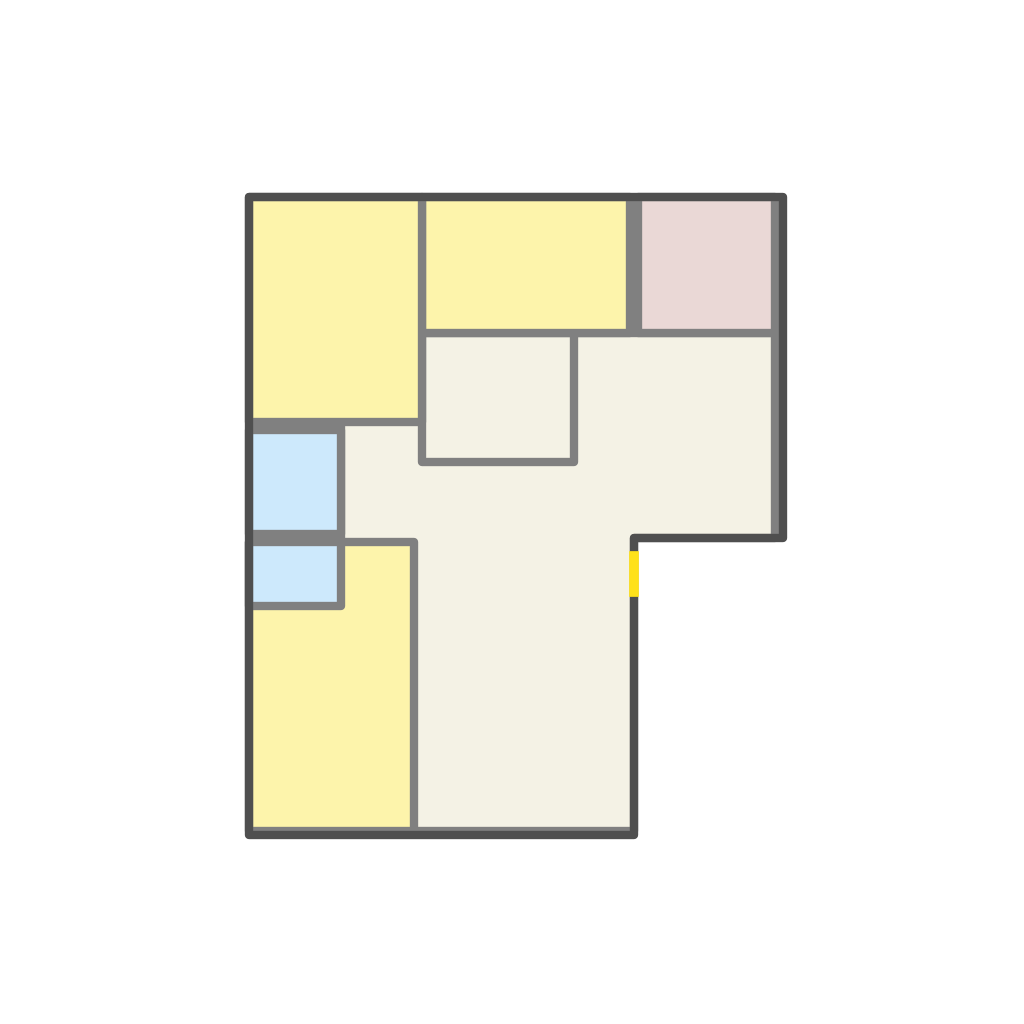}
     \end{subfigure}
    \begin{subfigure}{0.24\textwidth}
        \includegraphics[width=\textwidth]{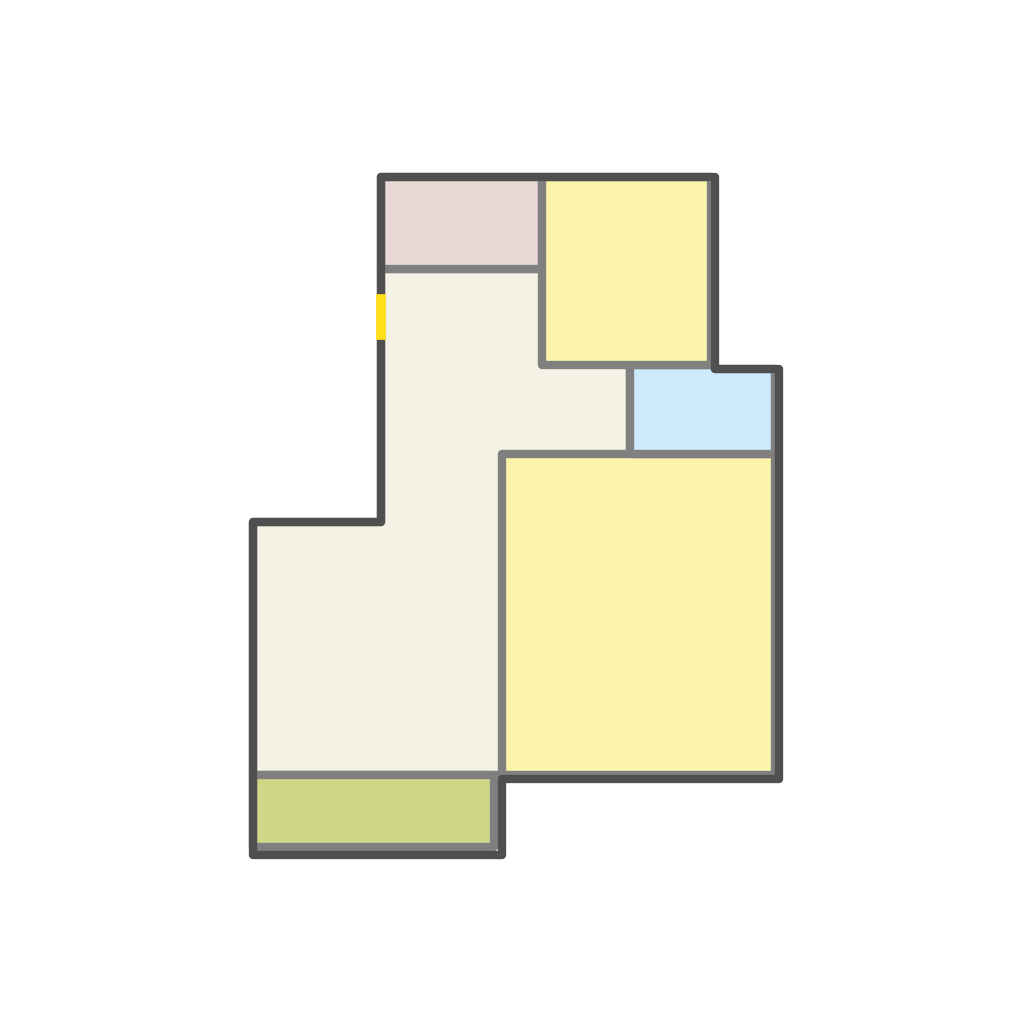}
     \end{subfigure}
    \begin{subfigure}{0.24\textwidth}
        \includegraphics[width=\textwidth]{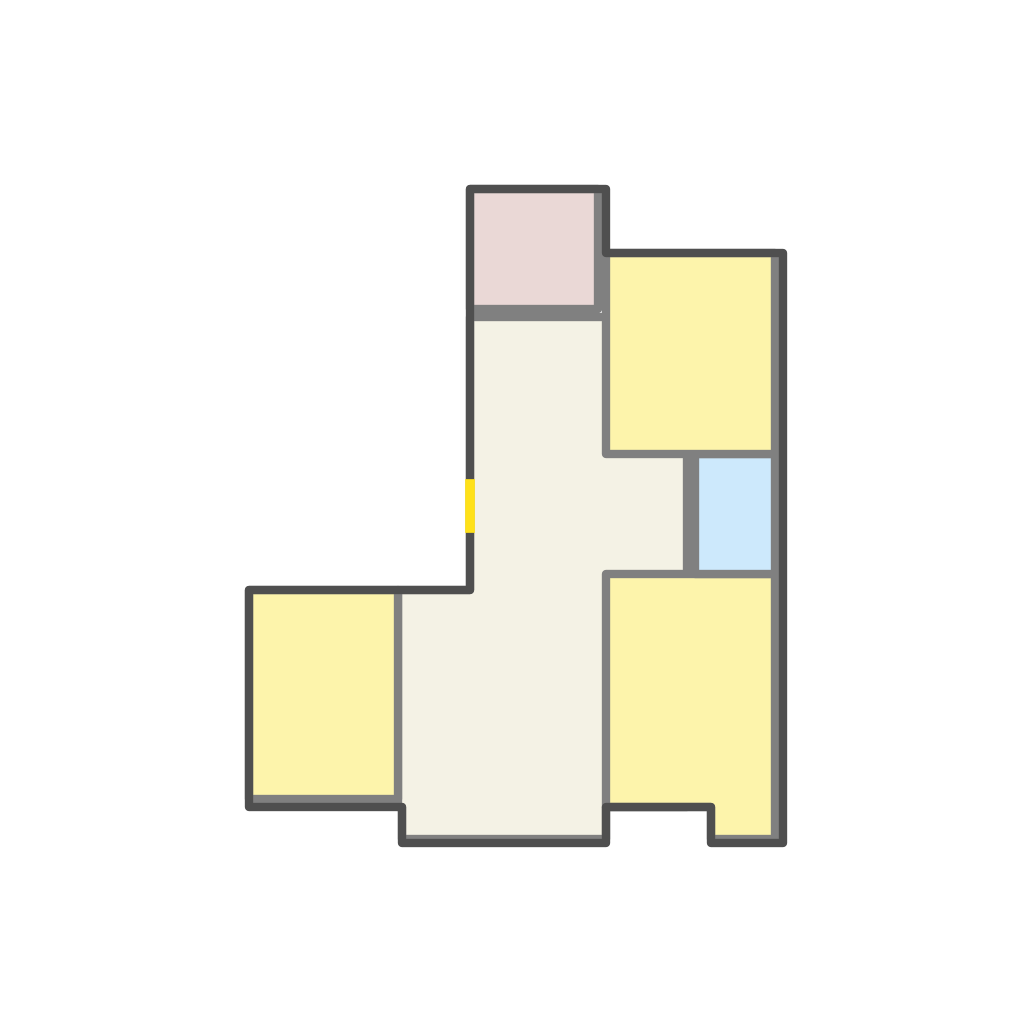}
     \end{subfigure}
    \begin{subfigure}{0.24\textwidth}
        \includegraphics[width=\textwidth]{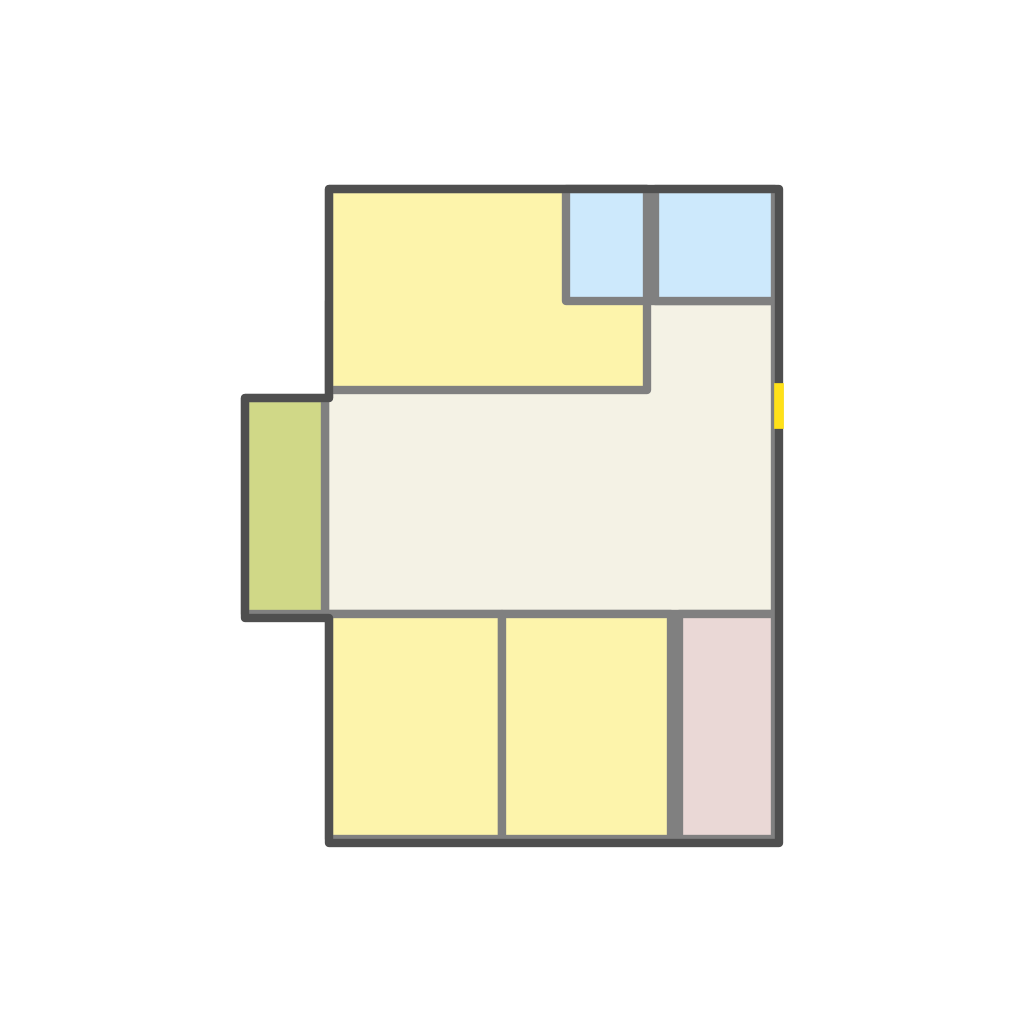}
     \end{subfigure}
     \\
    \subfloat{
        \raisebox{1.15in}{\rotatebox[origin=r]{90}{\Large FP-FGNN}}
    }\hfill\hfill
    \begin{subfigure}{0.24\textwidth}
        \includegraphics[width=\textwidth]{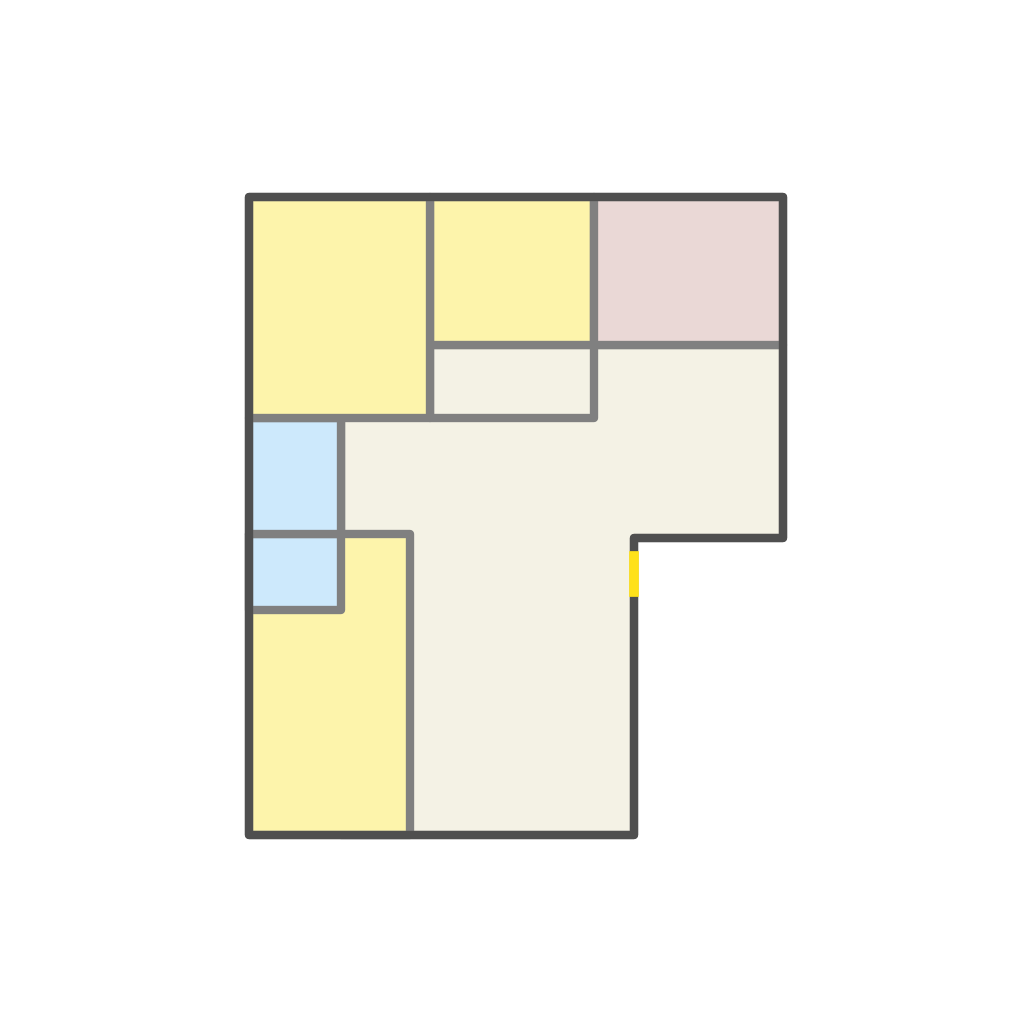}
     \end{subfigure}
    \begin{subfigure}{0.24\textwidth}
        \includegraphics[width=\textwidth]{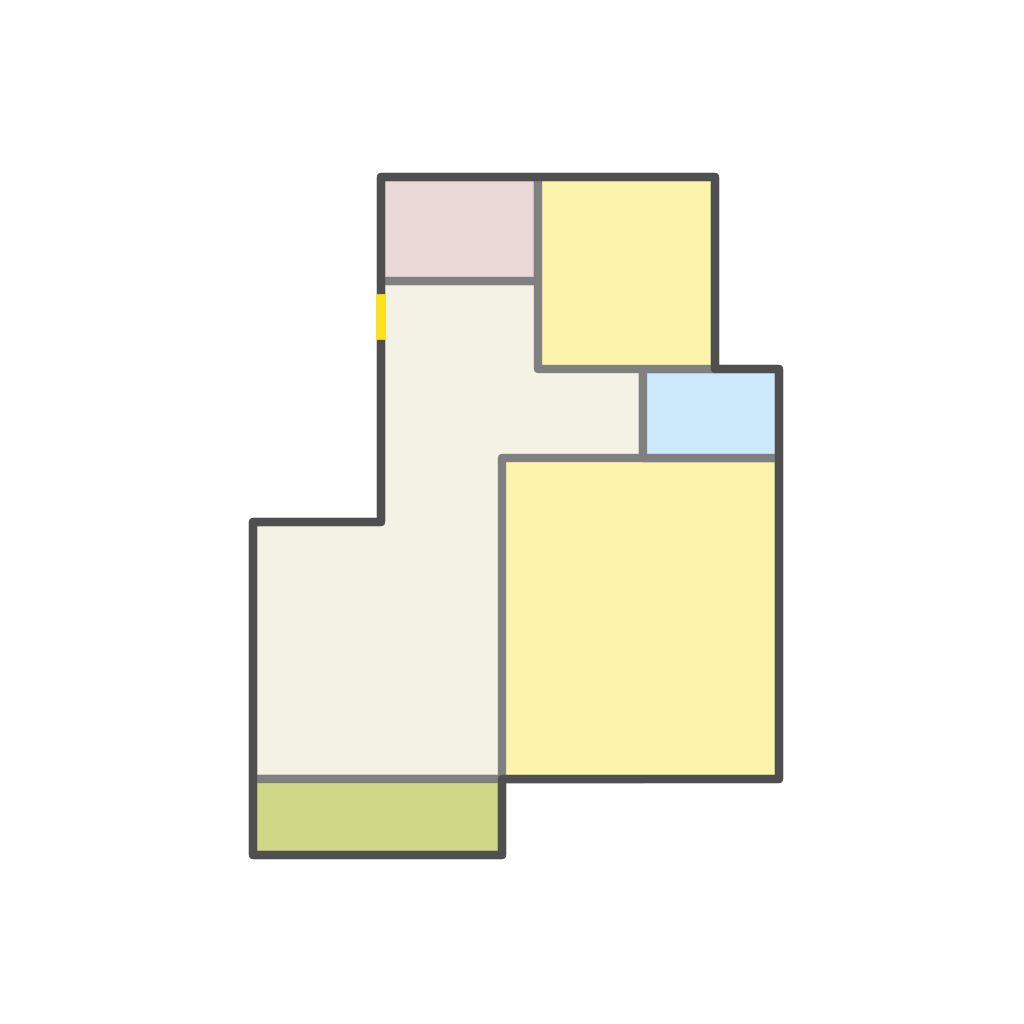}
     \end{subfigure}
    \begin{subfigure}{0.24\textwidth}
        \includegraphics[width=\textwidth]{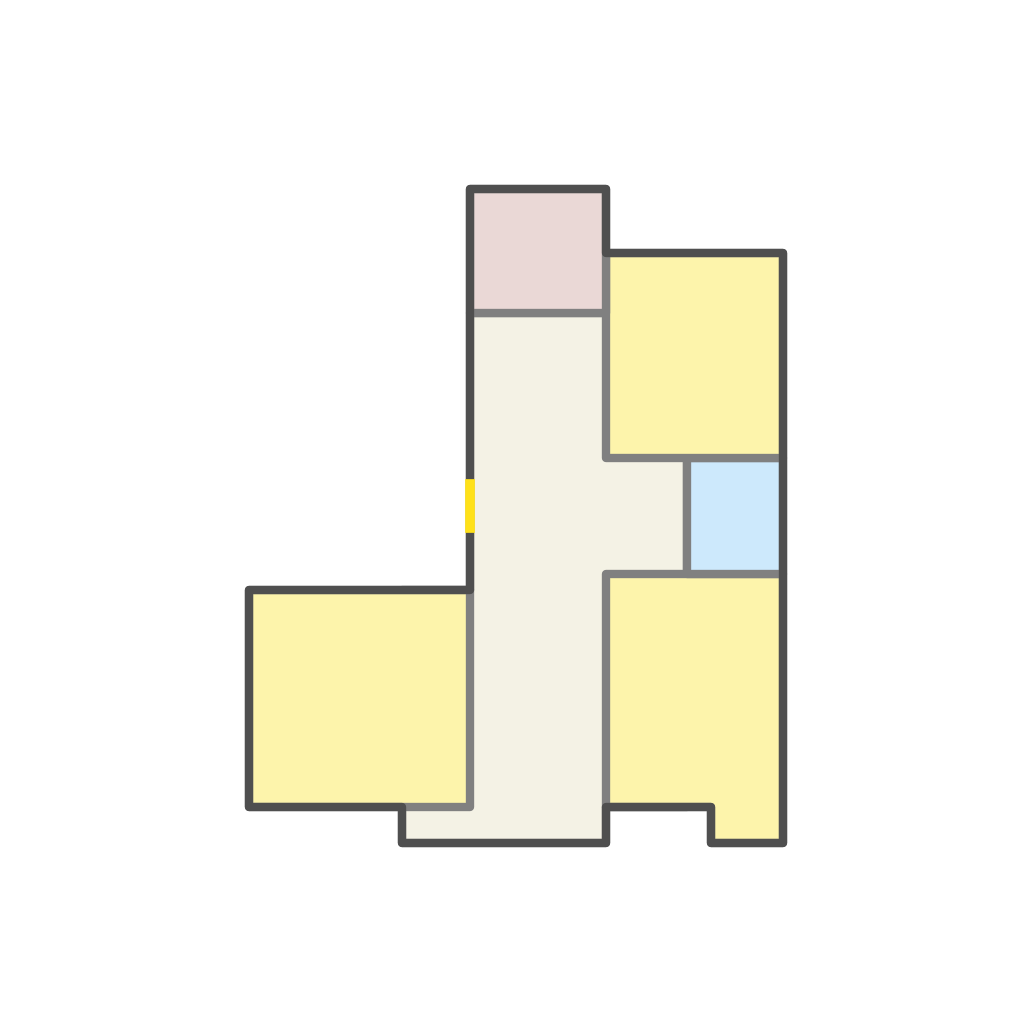}
     \end{subfigure}
    \begin{subfigure}{0.24\textwidth}
        \includegraphics[width=\textwidth]{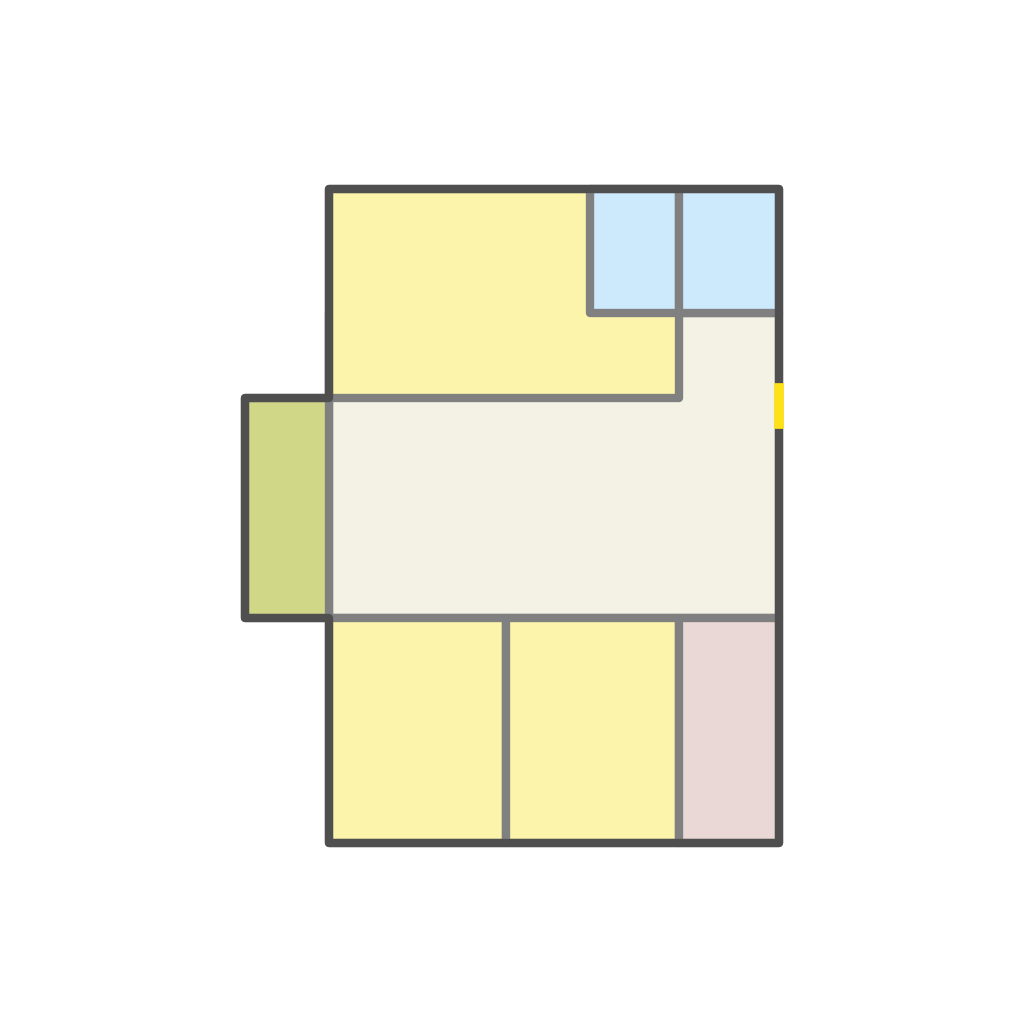}
     \end{subfigure}
    \\
    \subfloat{
        \raisebox{1.25in}{\rotatebox[origin=r]{90}{\Large Ground Truth}}
    }\hfill
    \begin{subfigure}{0.24\textwidth}
        \includegraphics[width=\textwidth]{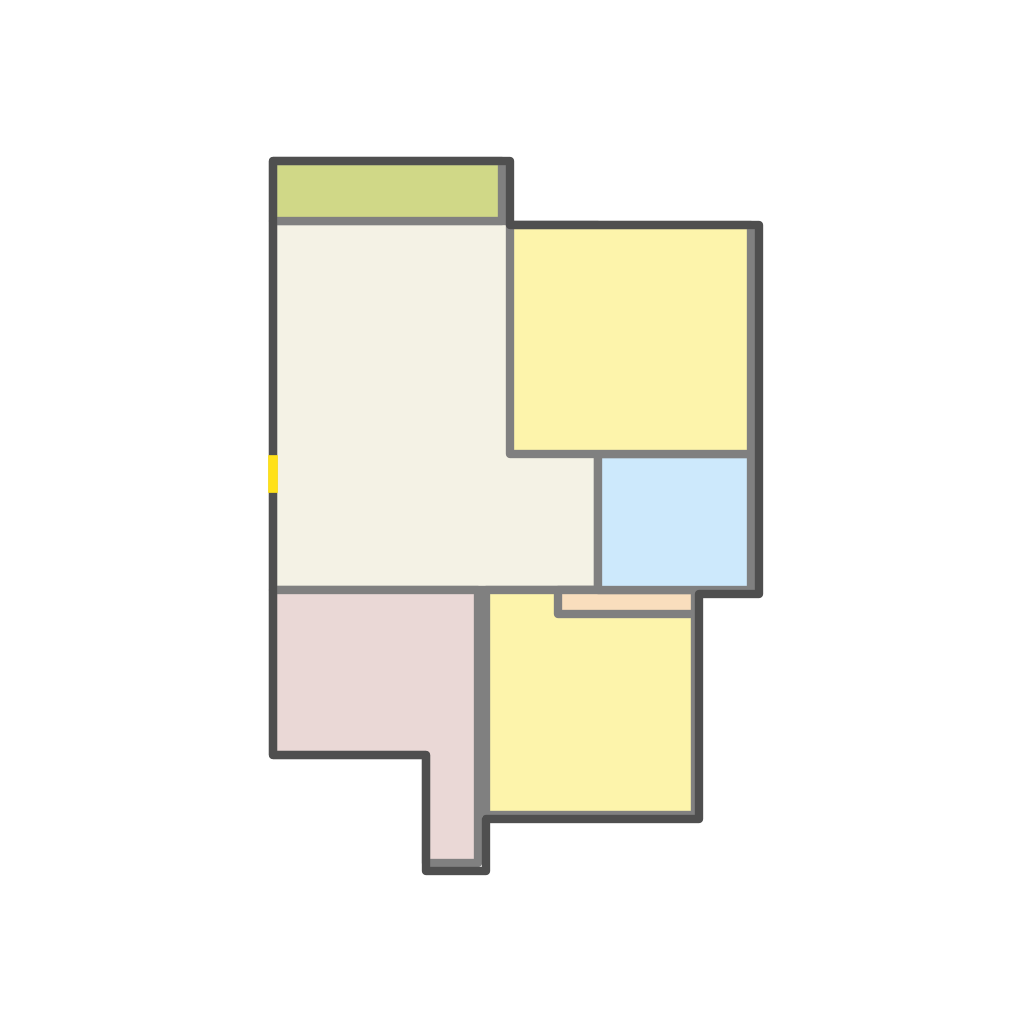}
     \end{subfigure}
    \begin{subfigure}{0.24\textwidth}
        \includegraphics[width=\textwidth]{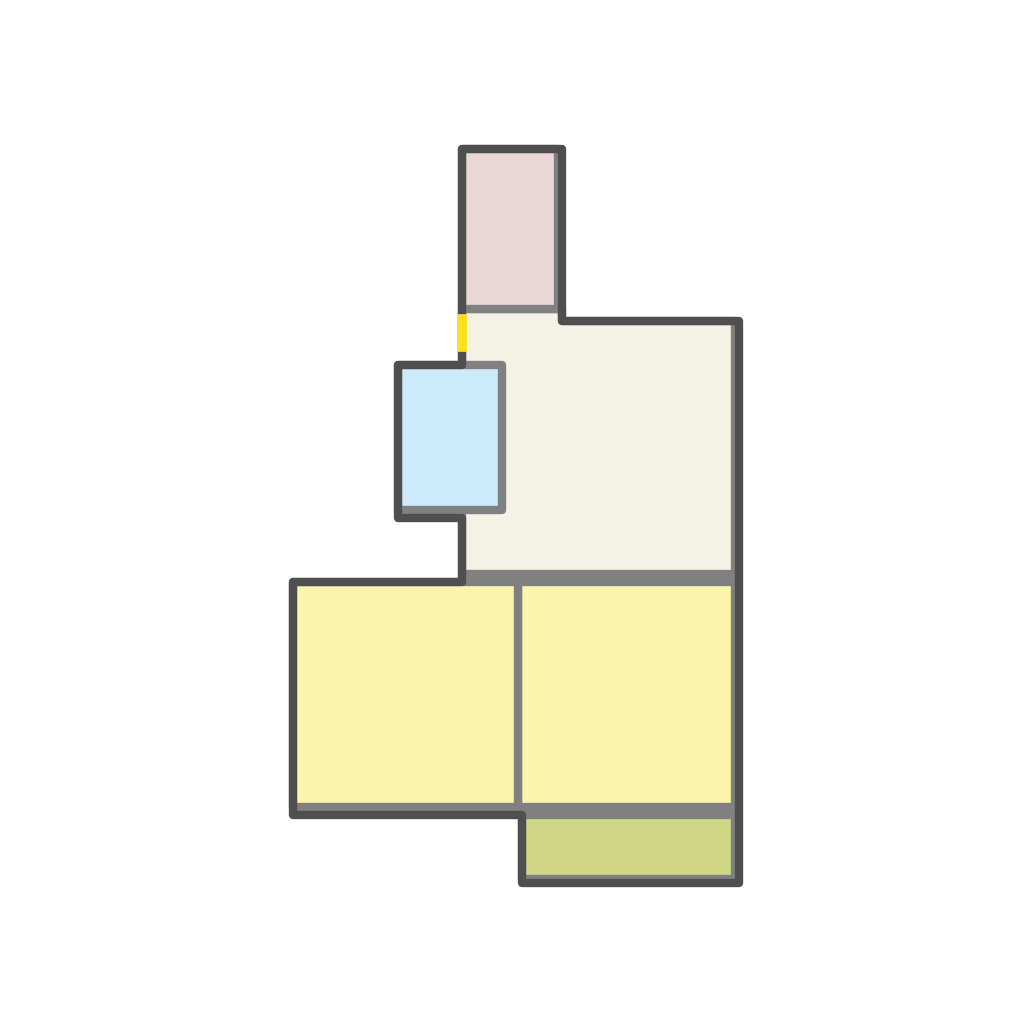}
     \end{subfigure}
    \begin{subfigure}{0.24\textwidth}
        \includegraphics[width=\textwidth]{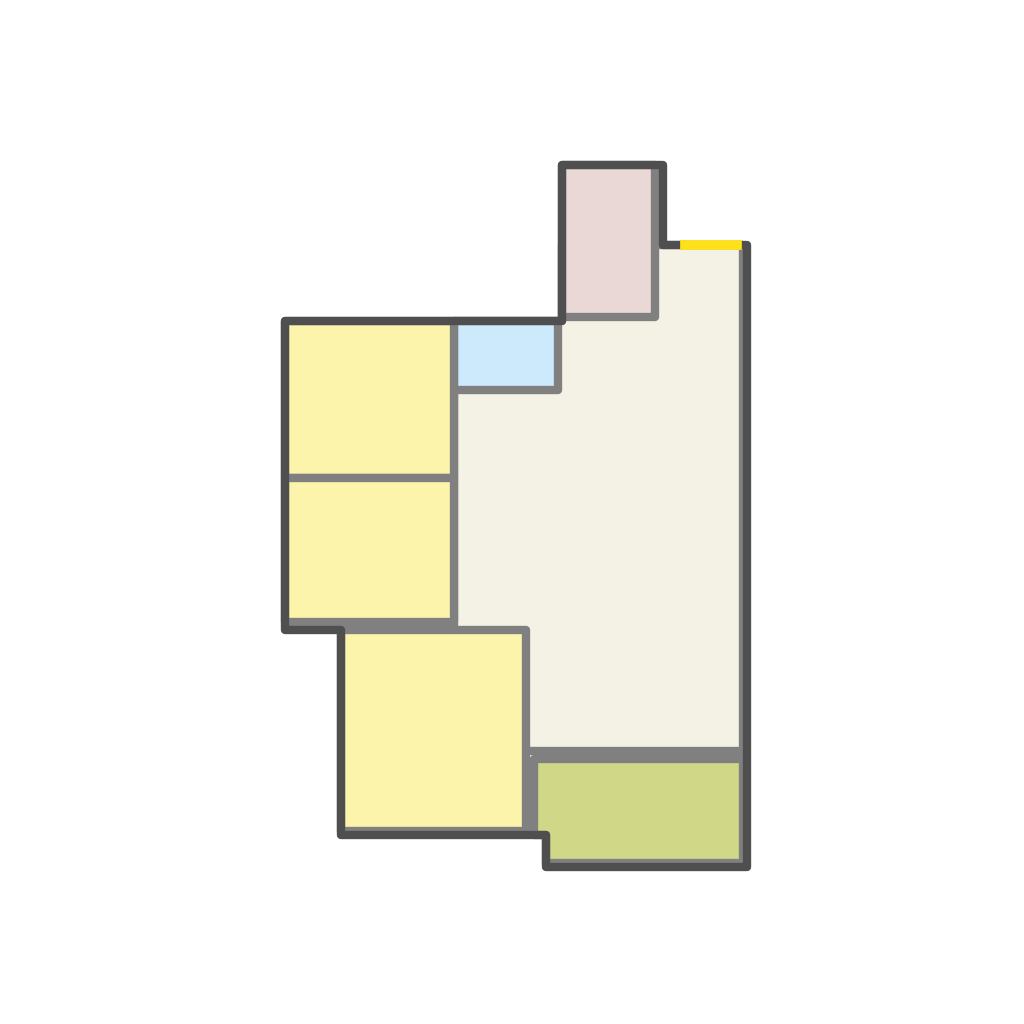}
     \end{subfigure}
    \begin{subfigure}{0.24\textwidth}
        \includegraphics[width=\textwidth]{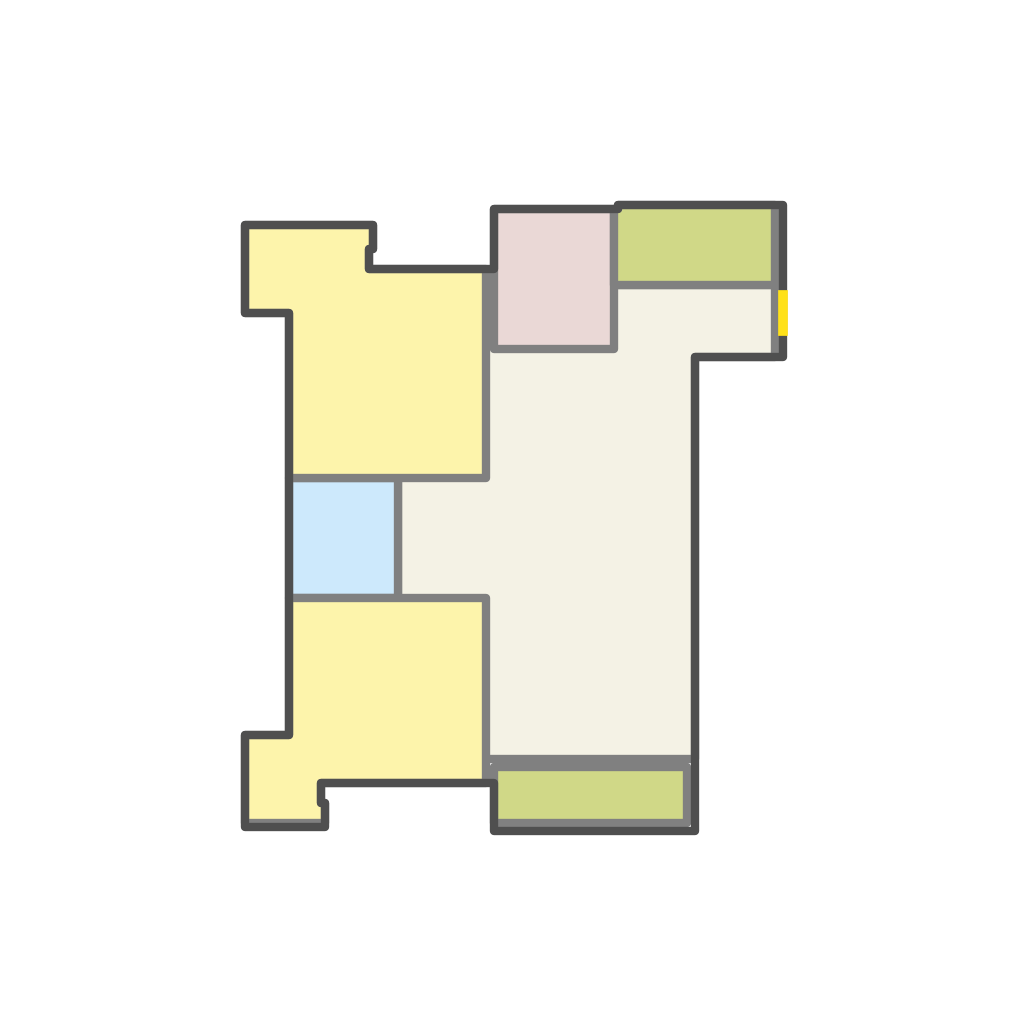}
     \end{subfigure}
     \\
    \subfloat{
        \raisebox{1.15in}{\rotatebox[origin=r]{90}{\Large FP-FGNN}}
    }\hfill
    \begin{subfigure}{0.24\textwidth}
        \includegraphics[width=\textwidth]{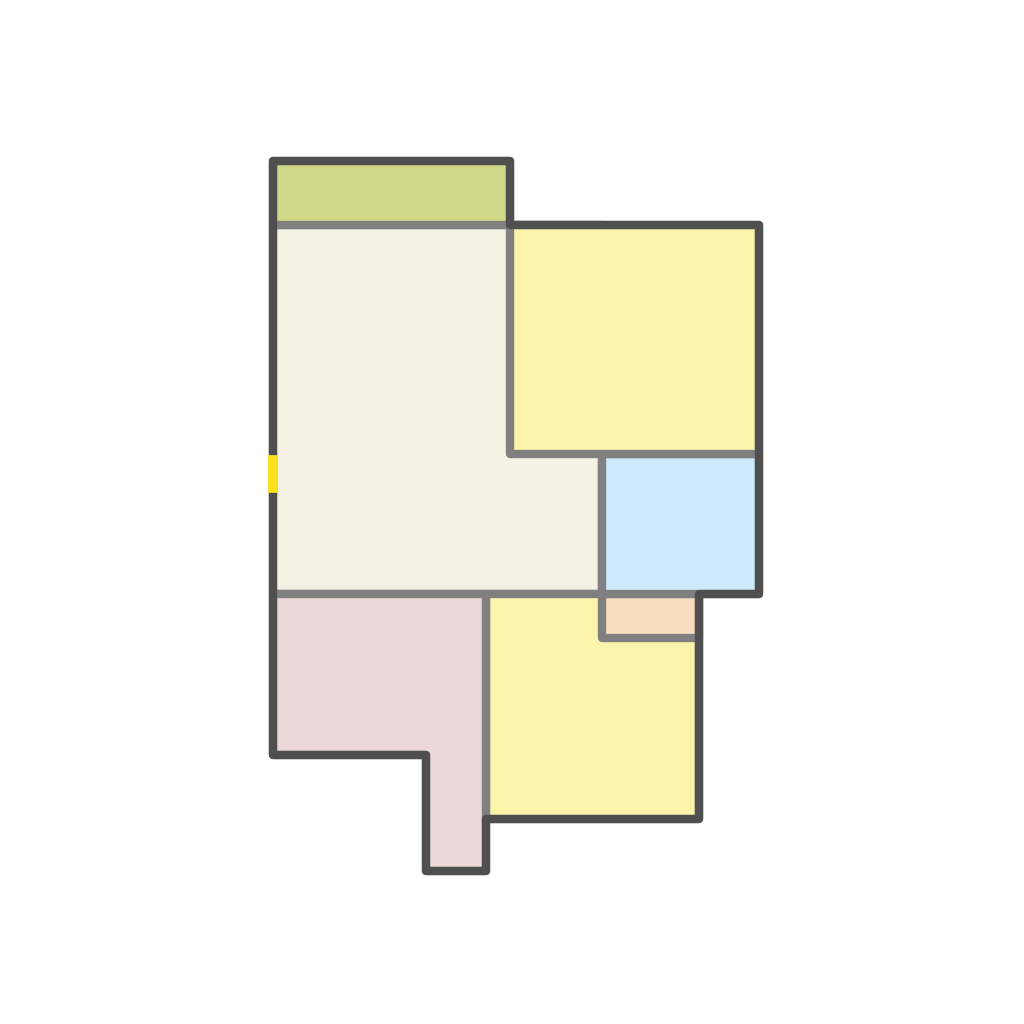}
     \end{subfigure}
    \begin{subfigure}{0.24\textwidth}
        \includegraphics[width=\textwidth]{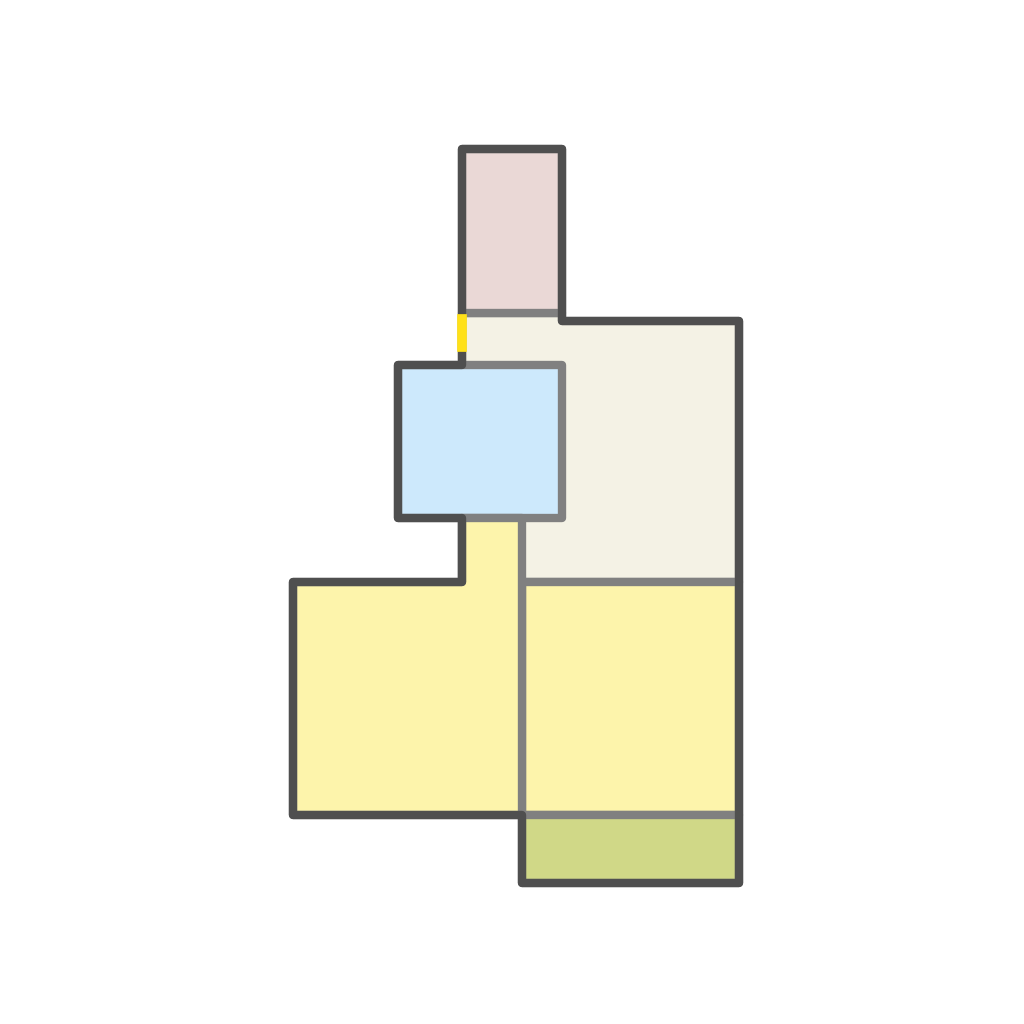}
     \end{subfigure}
    \begin{subfigure}{0.24\textwidth}
        \includegraphics[width=\textwidth]{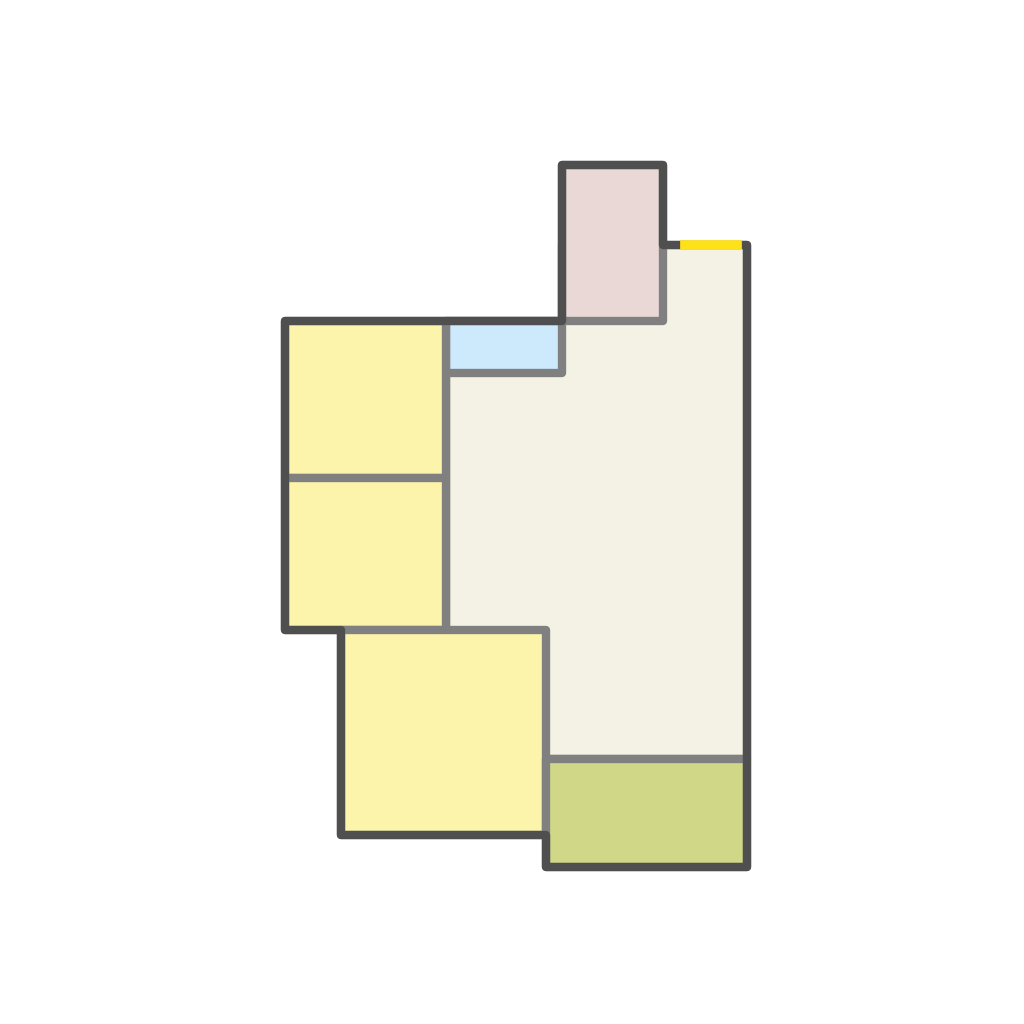}
     \end{subfigure}
    \begin{subfigure}{0.24\textwidth}
        \includegraphics[width=\textwidth]{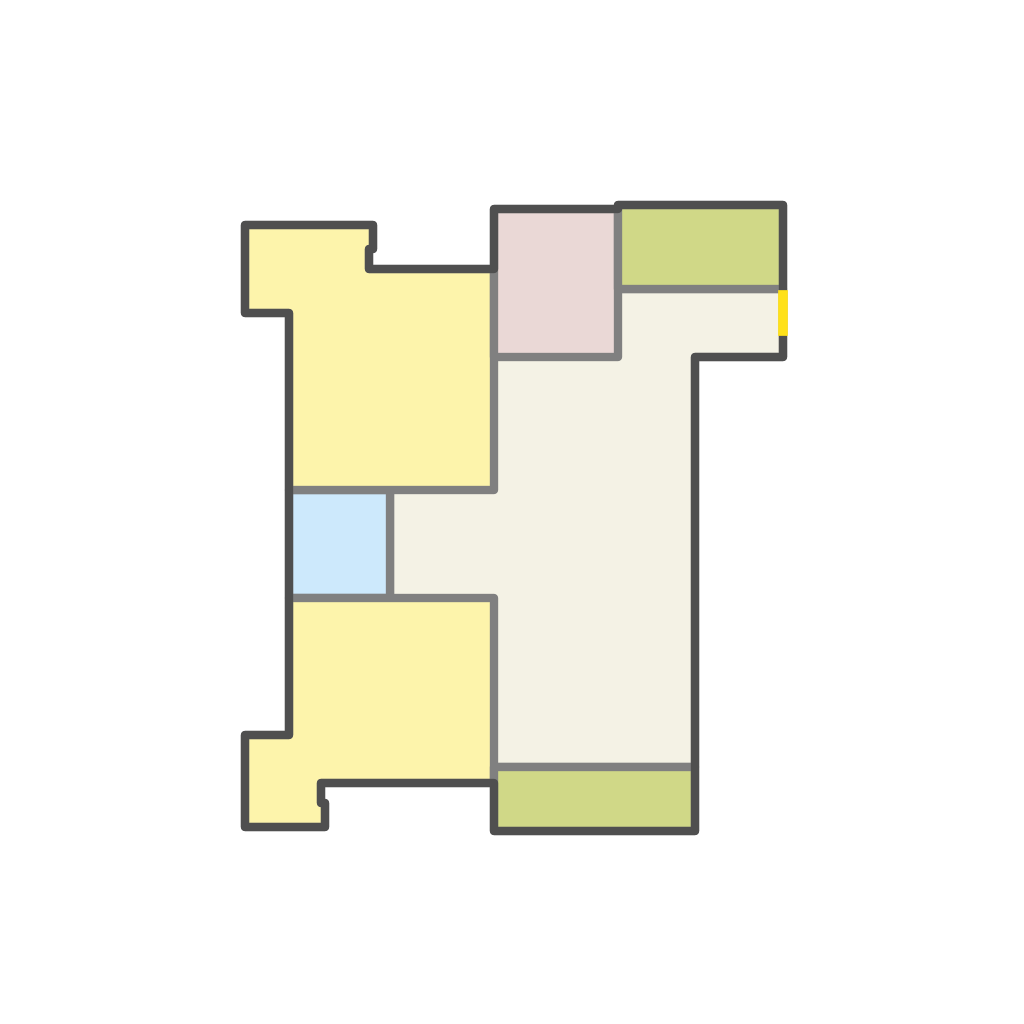}
    \end{subfigure}
    \end{tabular}
    }
    \caption{\small Comparison of floorplans generated by FP-FGNN with the ground truth after post-processing.}
    \label{fig:qualitative-predictions-1}
    \vskip -0.1in
\end{figure*}

\begin{figure*}[t]
    \centering
    \resizebox{0.8\textwidth}{!}{
    \begin{tabular}{ c }
    \subfloat{
        \raisebox{1.25in}{\rotatebox[origin=r]{90}{\Large Ground Truth}}
    }\hspace{5pt}
    \begin{subfigure}{0.24\textwidth}
        \includegraphics[width=\textwidth]{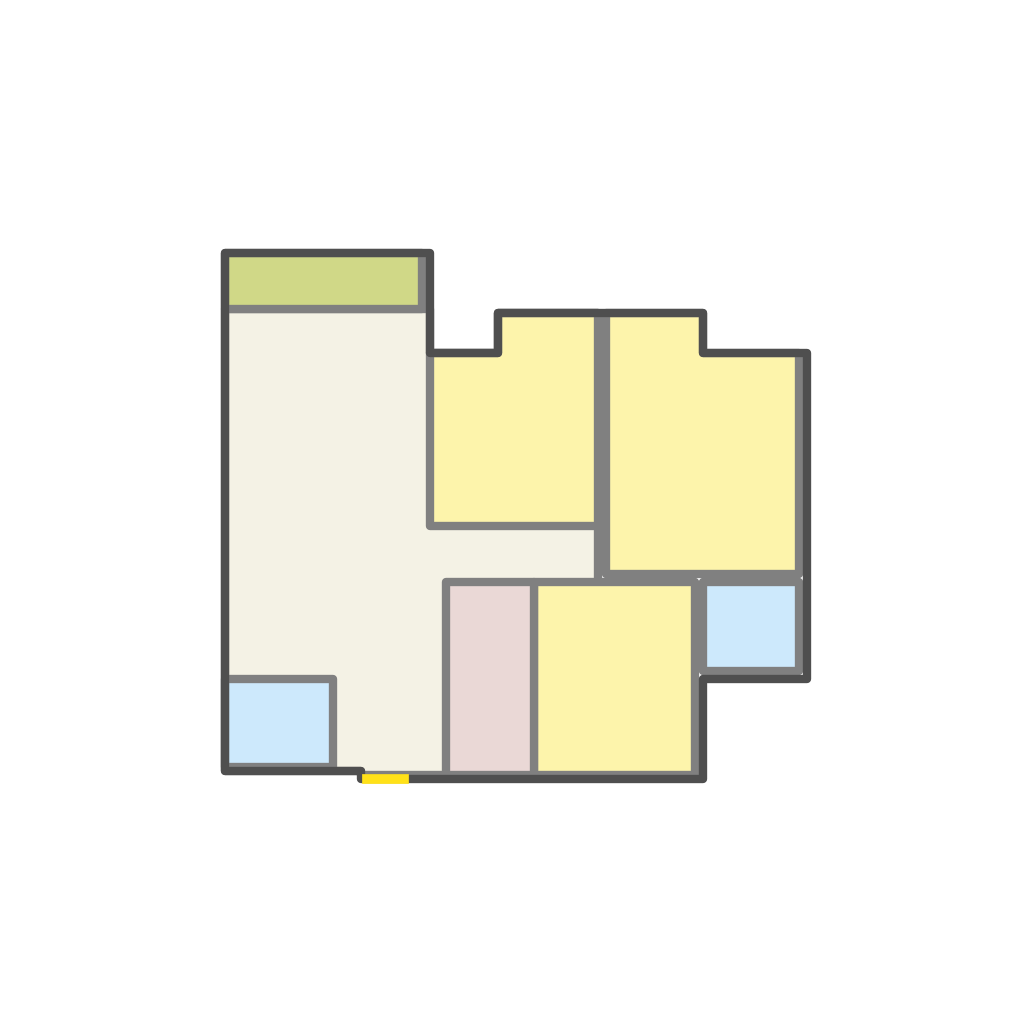}
    \end{subfigure}
    \begin{subfigure}{0.24\textwidth}
        \includegraphics[width=\textwidth]{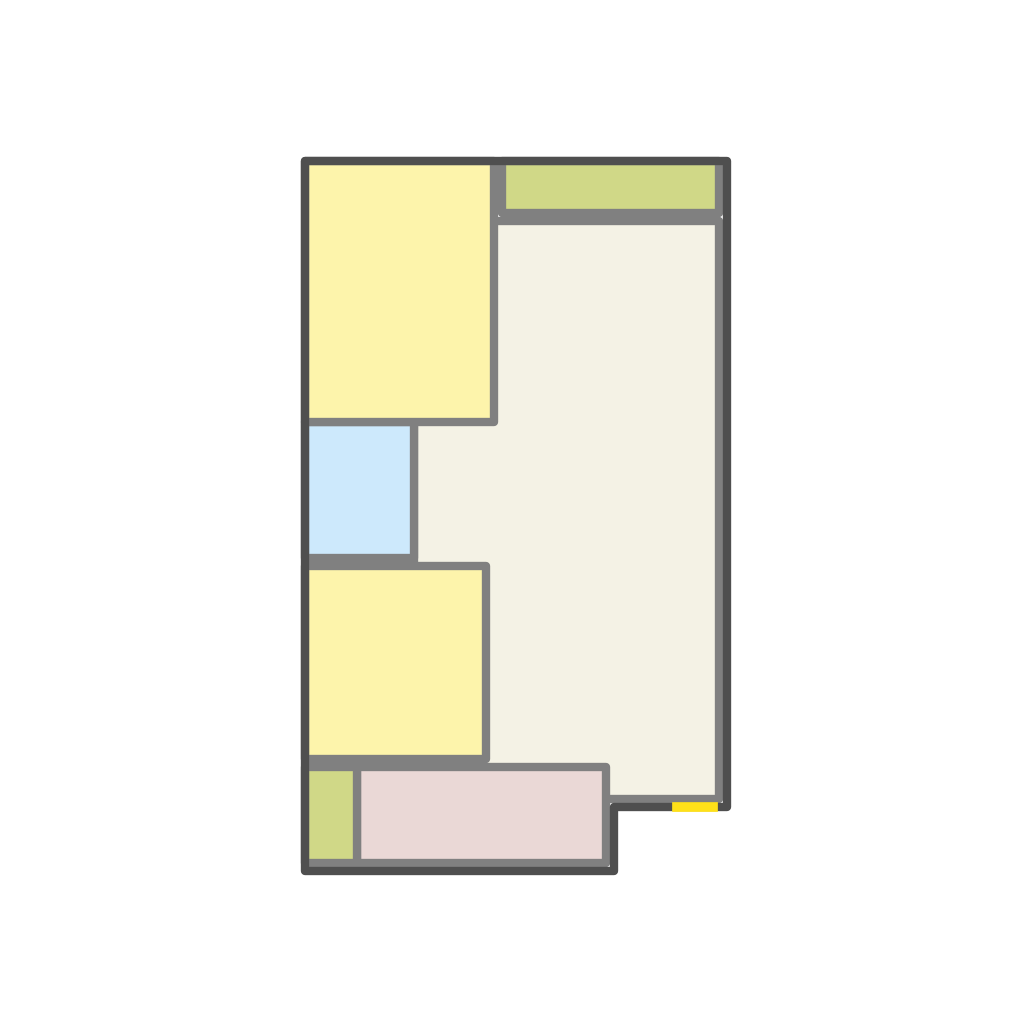}
    \end{subfigure}
    \begin{subfigure}{0.24\textwidth}
        \includegraphics[width=\textwidth]{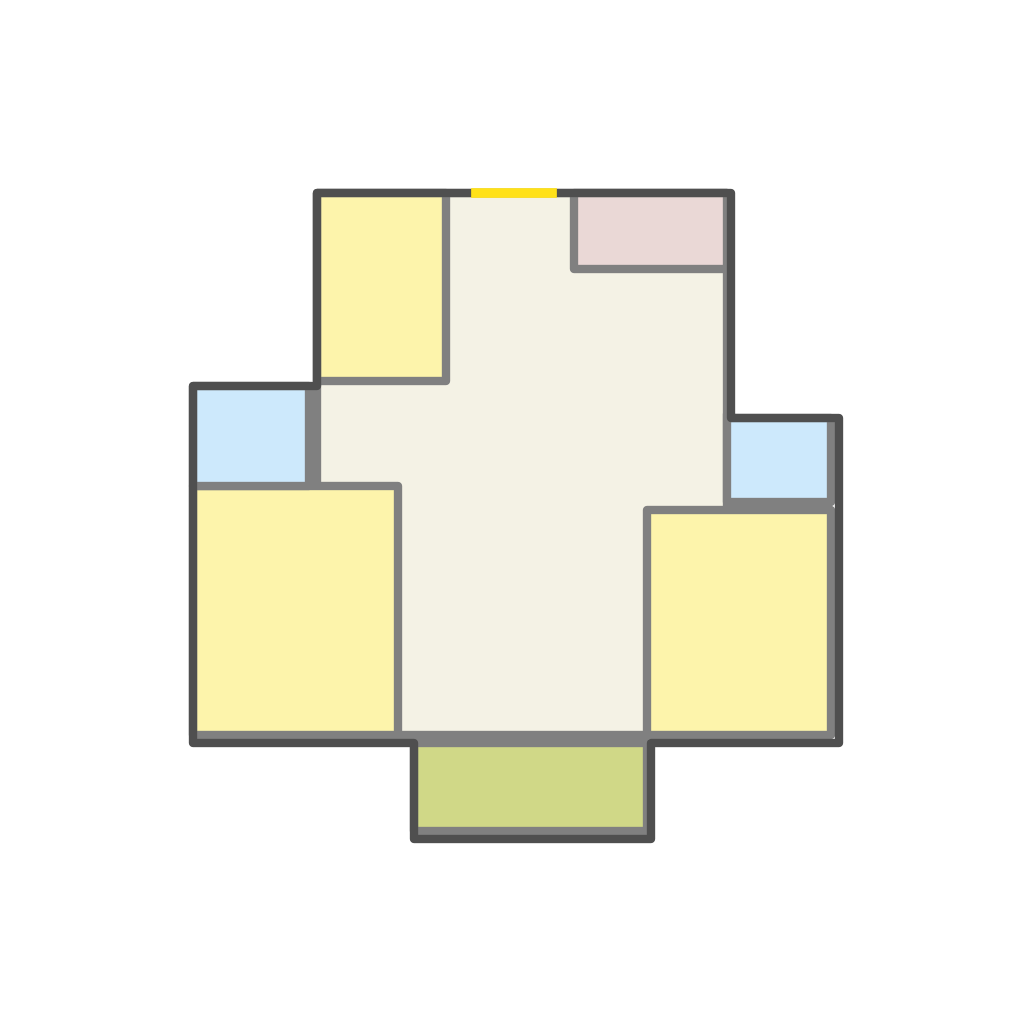}
    \end{subfigure}
    \begin{subfigure}{0.24\textwidth}
        \includegraphics[width=\textwidth]{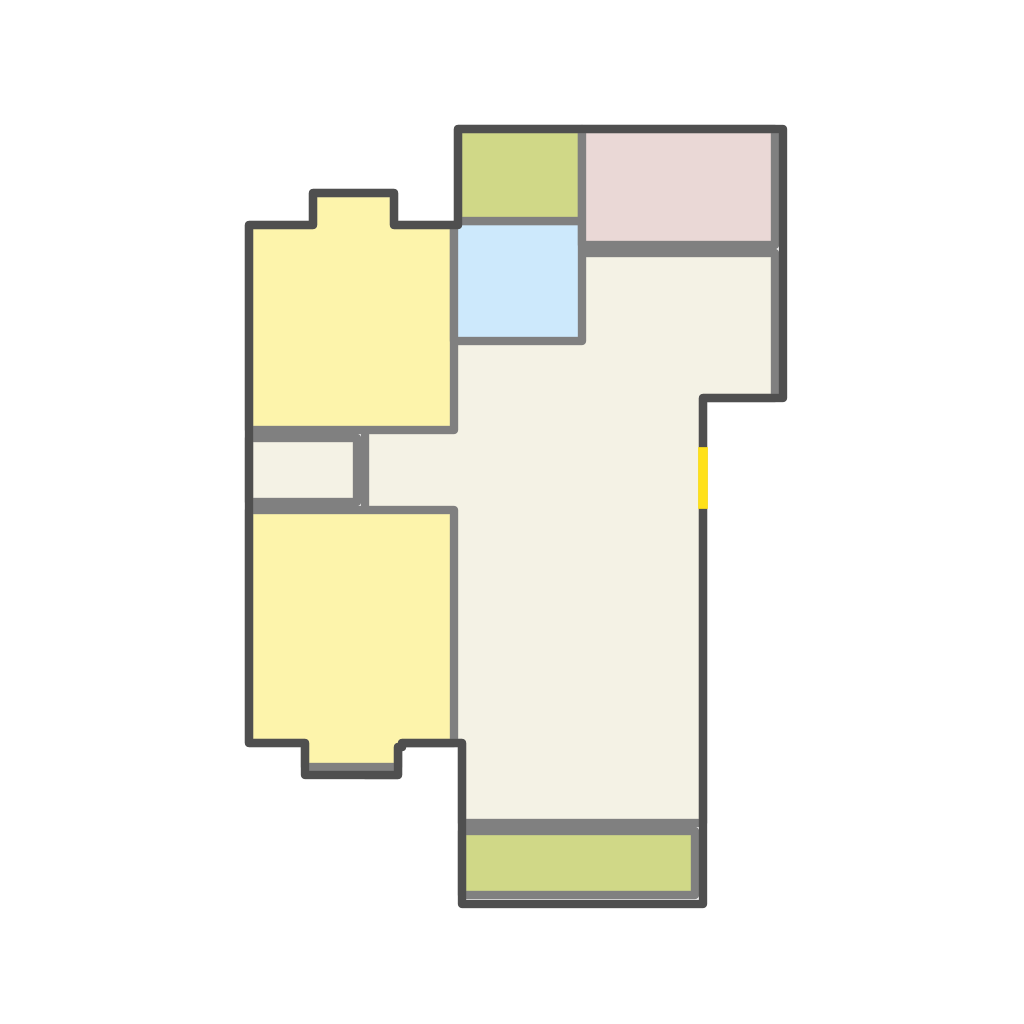}
    \end{subfigure}
    \\ 
    \subfloat{
        \raisebox{1.15in}{\rotatebox[origin=r]{90}{\Large FP-FGNN}}
    }\hfill
    \begin{subfigure}{0.24\textwidth}
        \includegraphics[width=\textwidth]{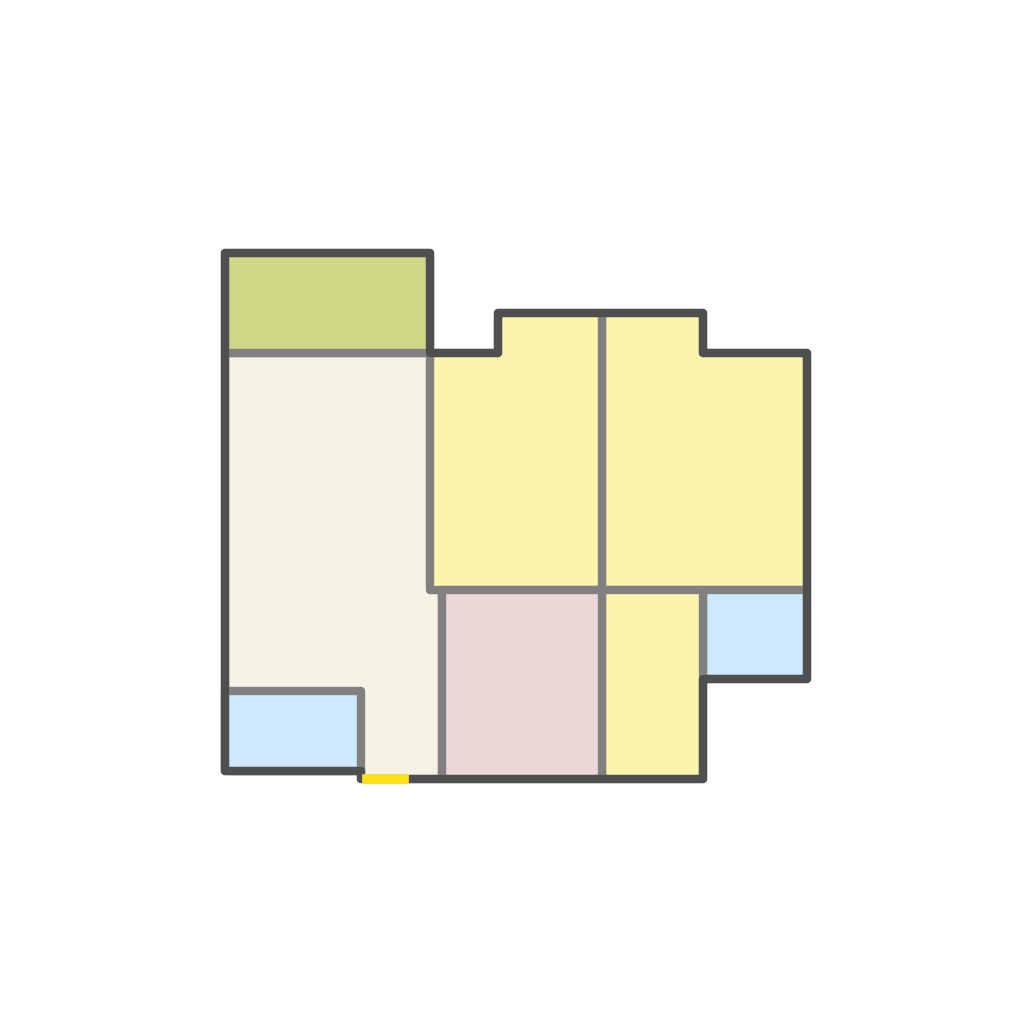}
    \end{subfigure}
    \begin{subfigure}{0.24\textwidth}
        \includegraphics[width=\textwidth]{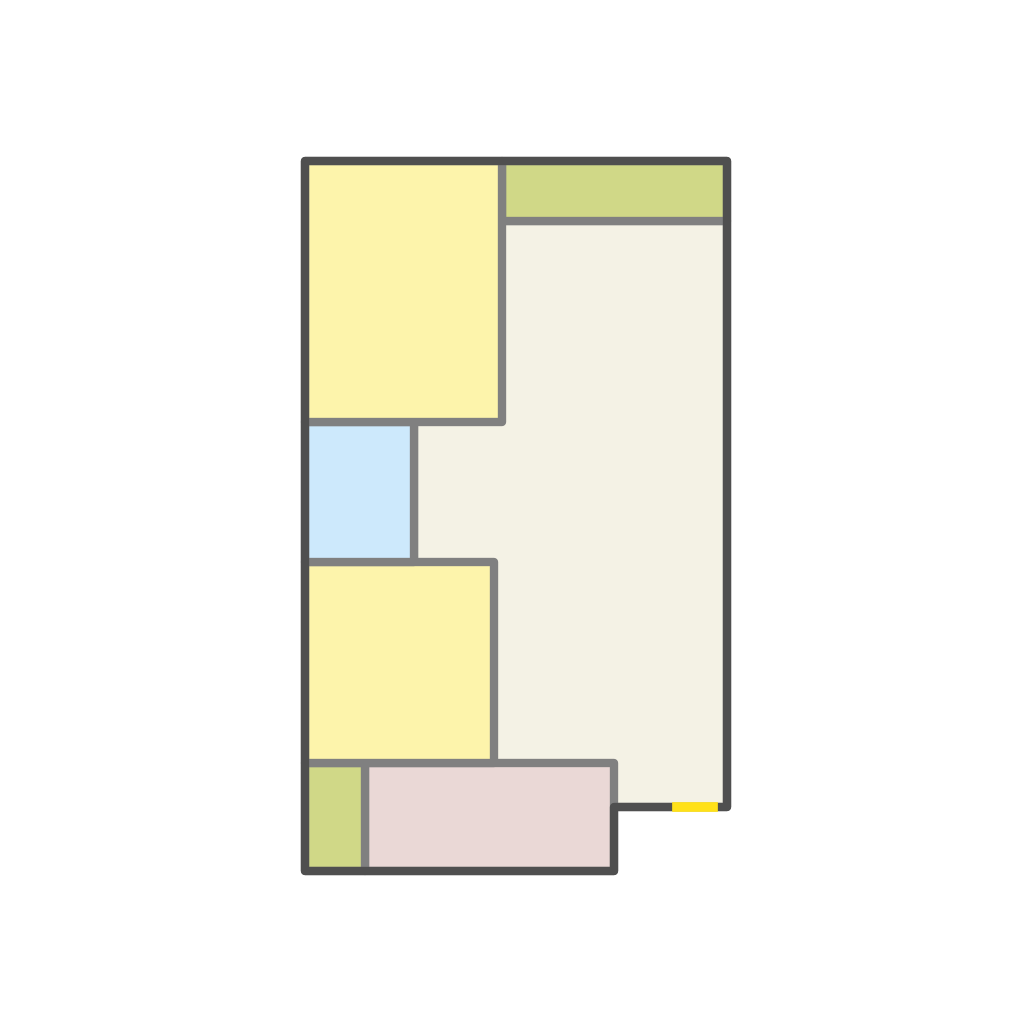}
    \end{subfigure}
    \begin{subfigure}{0.24\textwidth}
        \includegraphics[width=\textwidth]{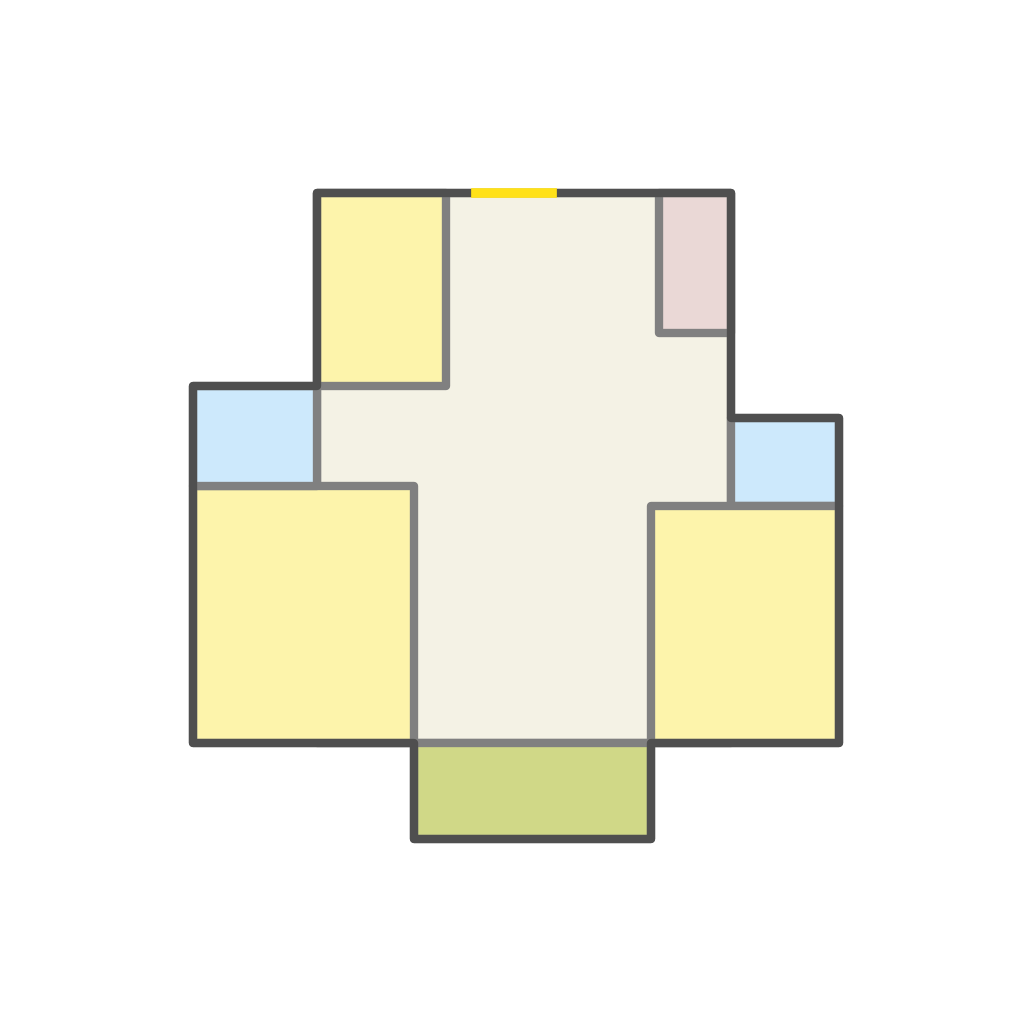}
    \end{subfigure}
    \begin{subfigure}{0.24\textwidth}
        \includegraphics[width=\textwidth]{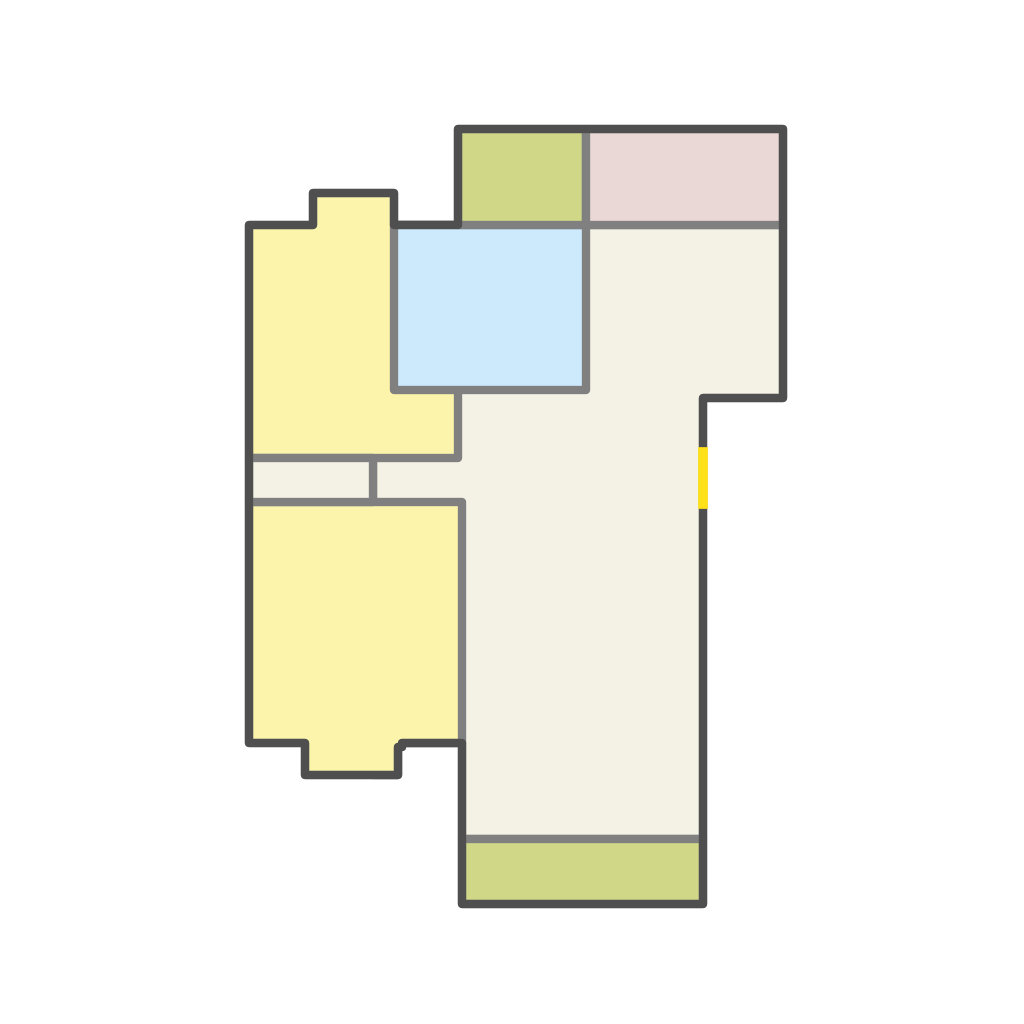}
    \end{subfigure}
    \\
    \subfloat{
        \raisebox{1.25in}{\rotatebox[origin=r]{90}{\Large Ground Truth}}
    }\hfill
    \begin{subfigure}{0.24\textwidth}
        \includegraphics[width=\textwidth]{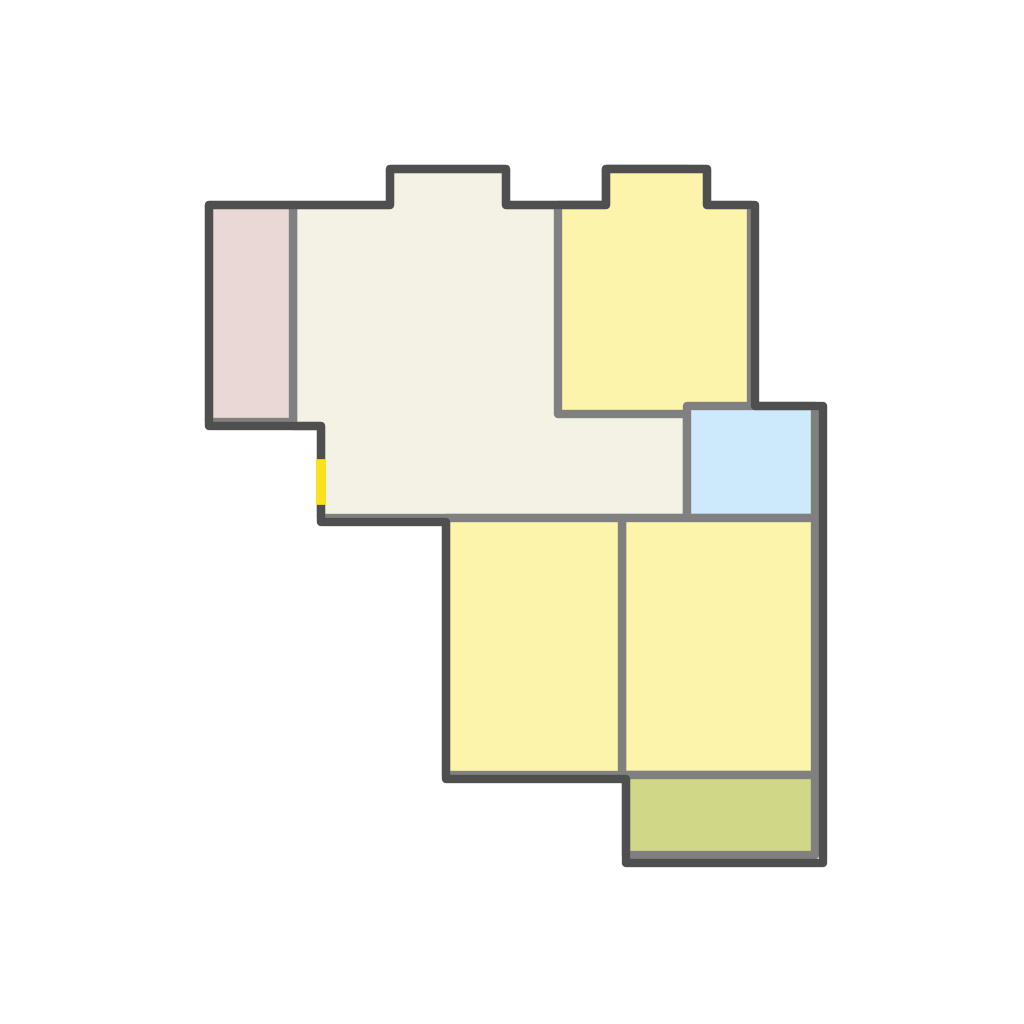}
    \end{subfigure}
    \begin{subfigure}{0.24\textwidth}
        \includegraphics[width=\textwidth]{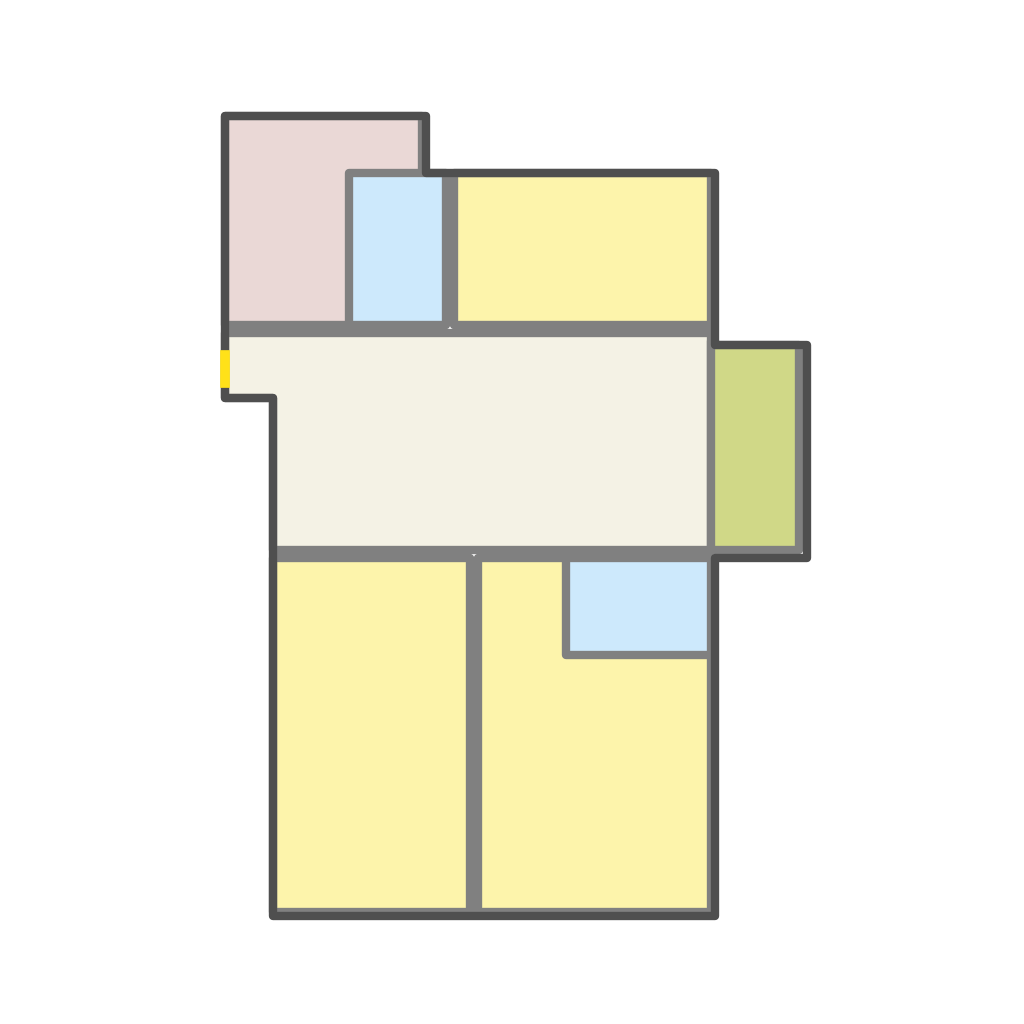}
    \end{subfigure}
    \begin{subfigure}{0.24\textwidth}
        \includegraphics[width=\textwidth]{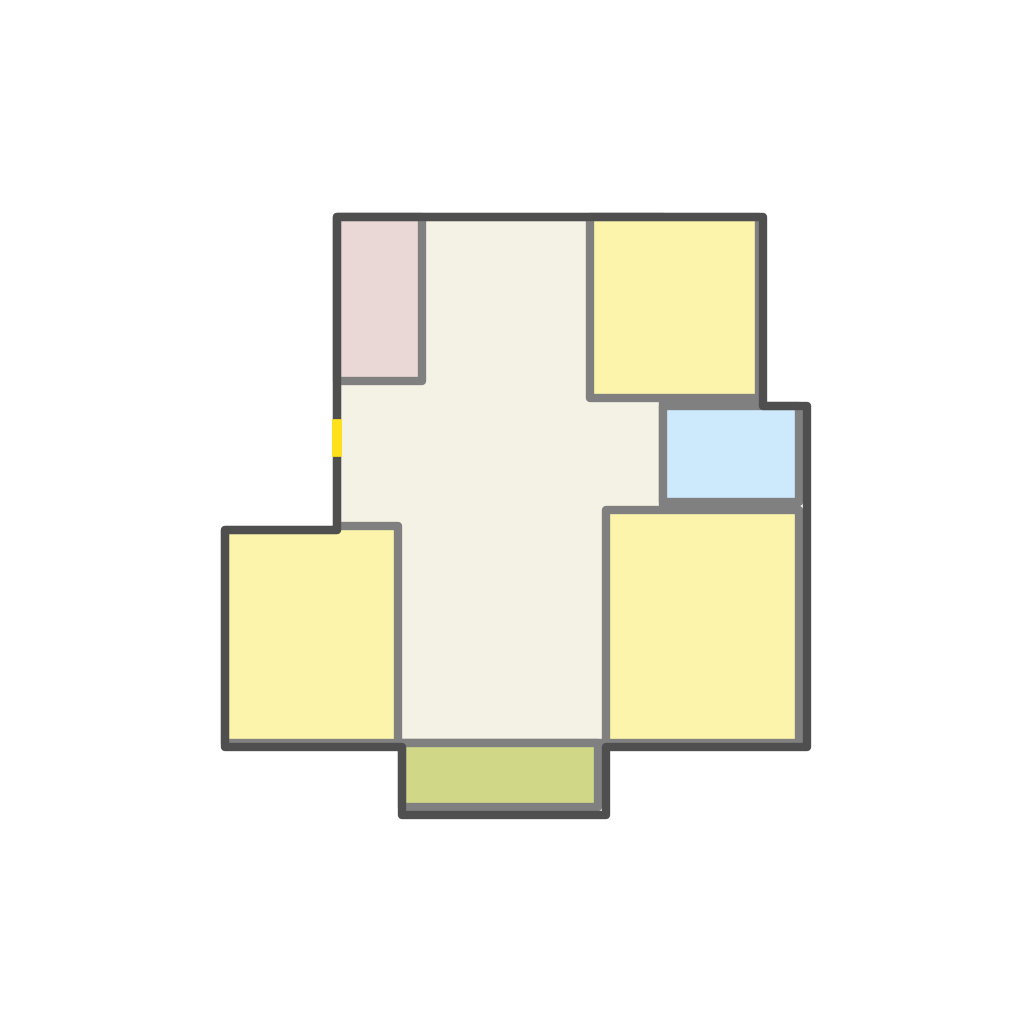}
    \end{subfigure}
    \begin{subfigure}{0.24\textwidth}
        \includegraphics[width=\textwidth]{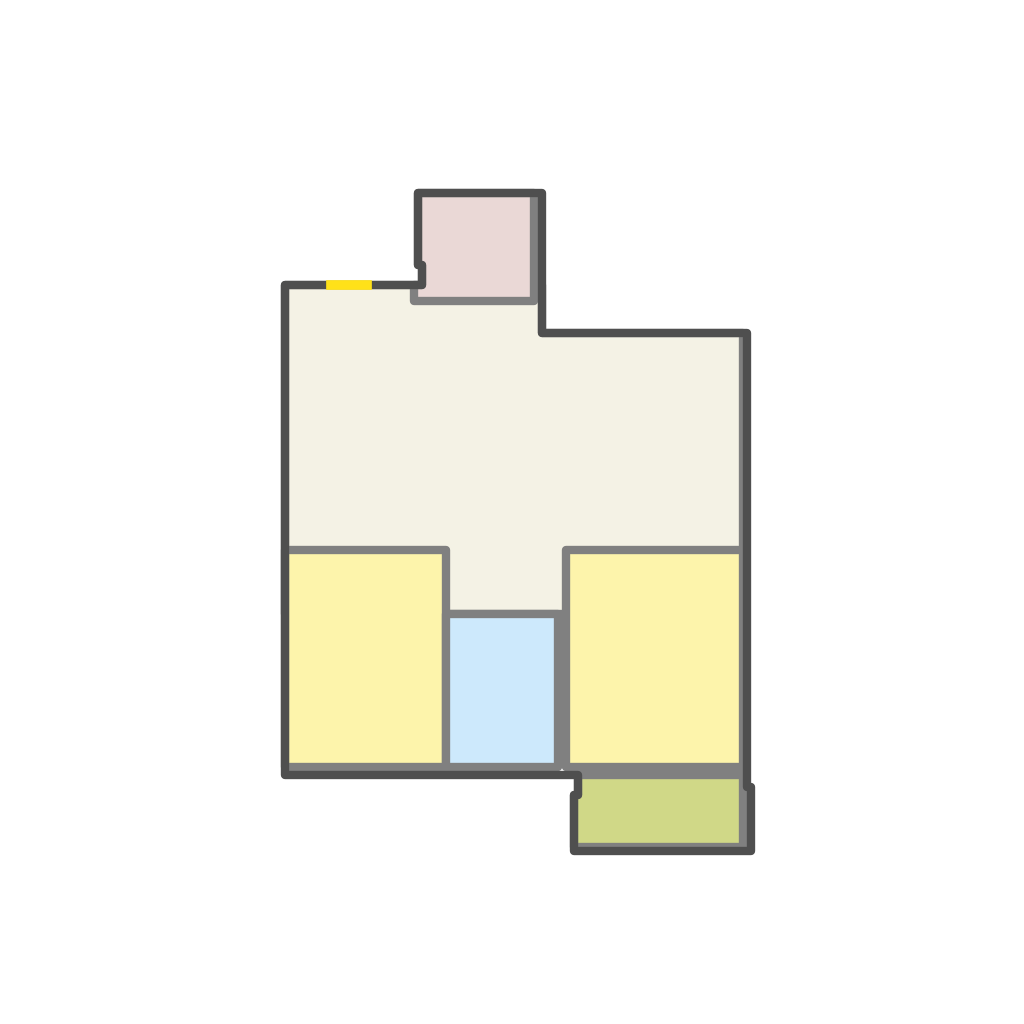}
    \end{subfigure}
    \\\hfill\hfill\hfill
    \subfloat{
        \raisebox{1.15in}{\rotatebox[origin=r]{90}{\Large FP-FGNN}}
    }\hfill\hfill\hfill
    \begin{subfigure}{0.24\textwidth}
        \includegraphics[width=\textwidth]{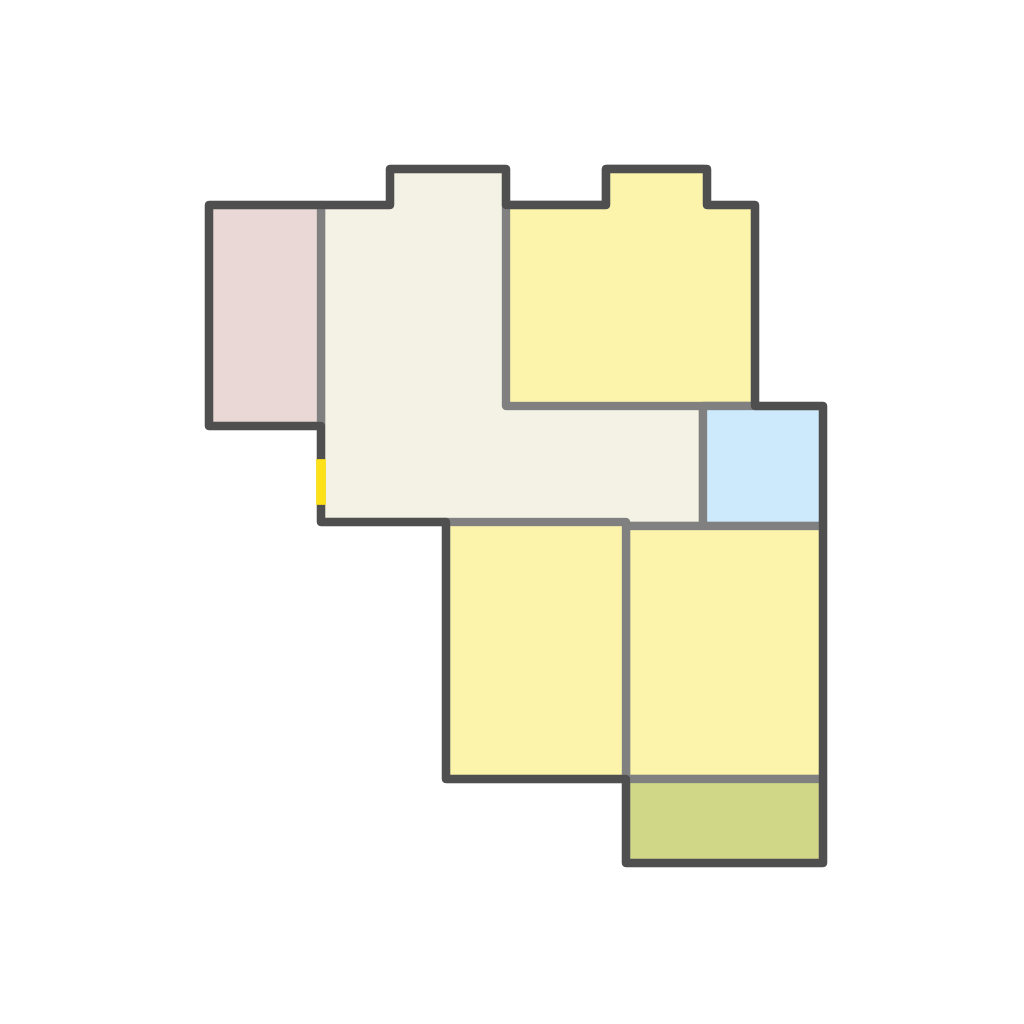}
    \end{subfigure}
    \begin{subfigure}{0.24\textwidth}
        \includegraphics[width=\textwidth]{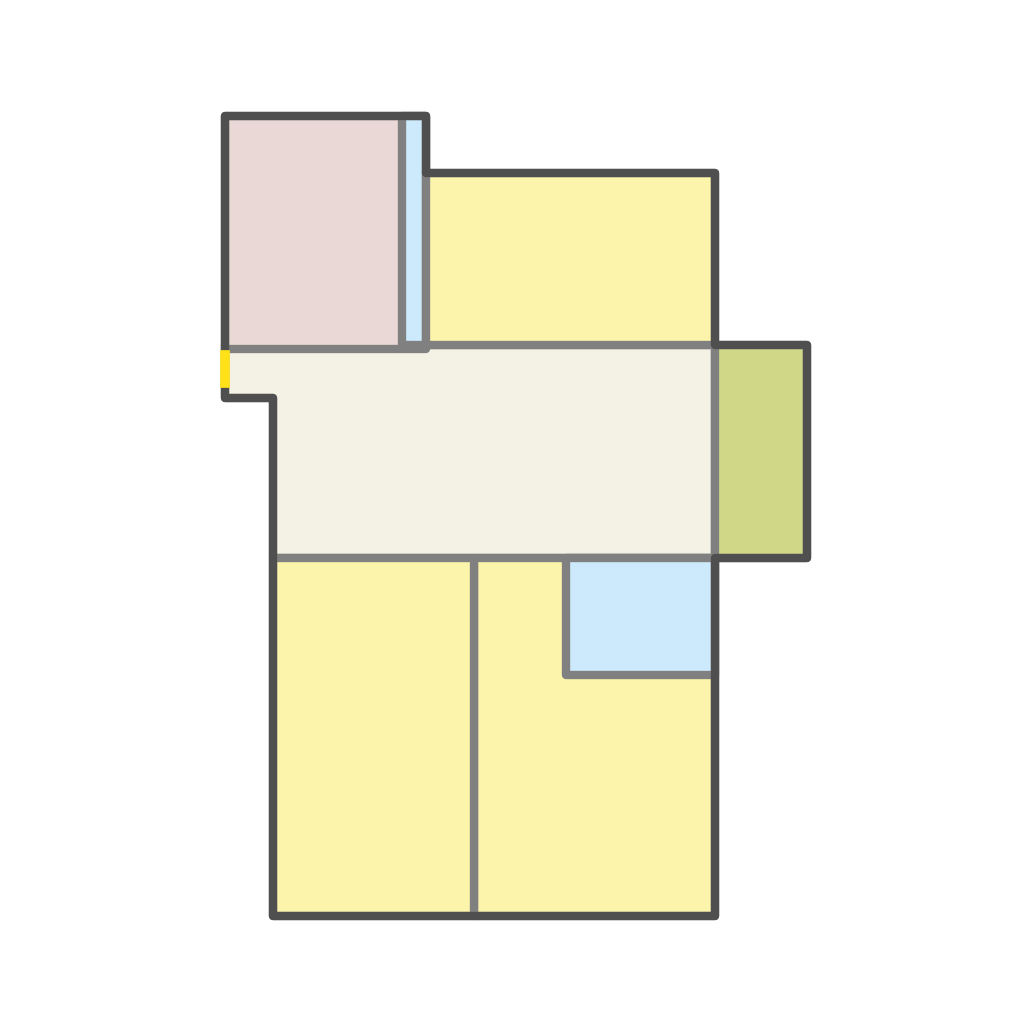}
    \end{subfigure}
    \begin{subfigure}{0.24\textwidth}
        \includegraphics[width=\textwidth]{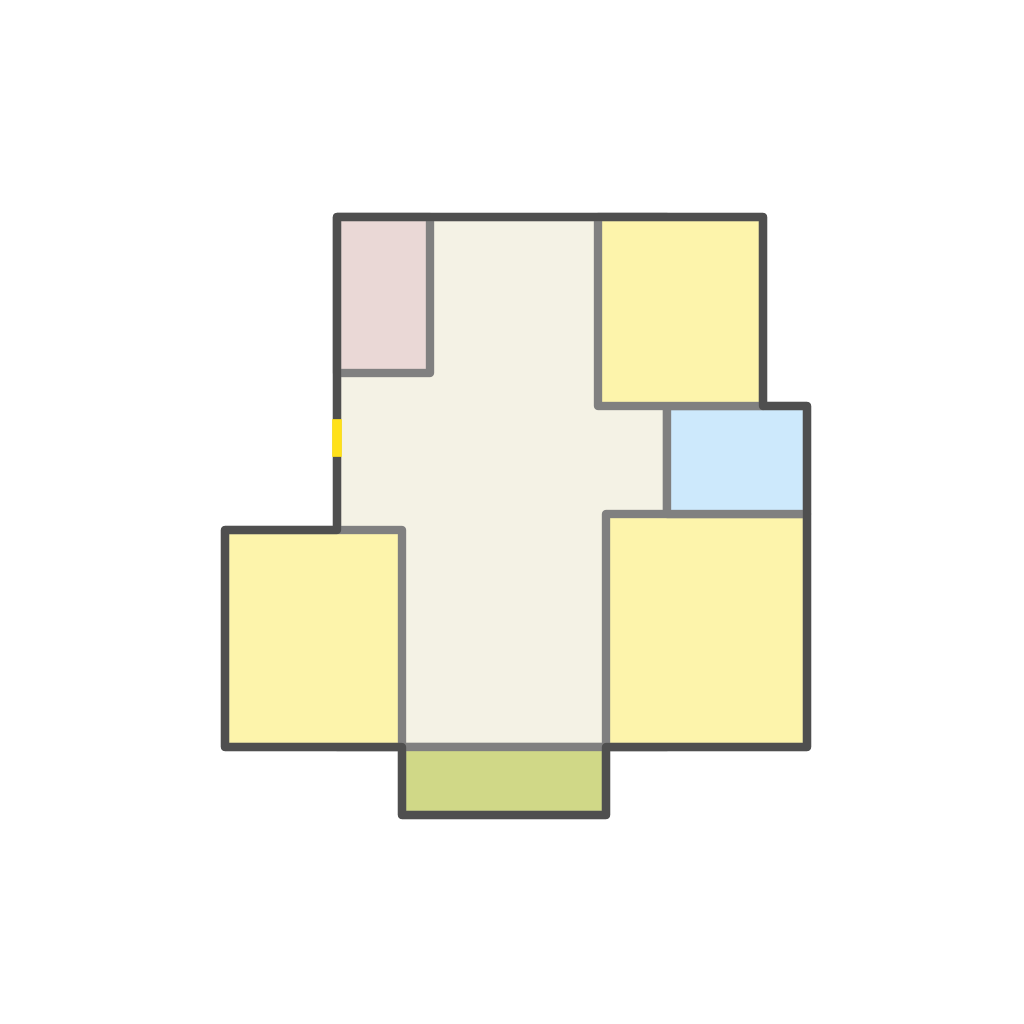}
    \end{subfigure}
    \begin{subfigure}{0.24\textwidth}
        \includegraphics[width=\textwidth]{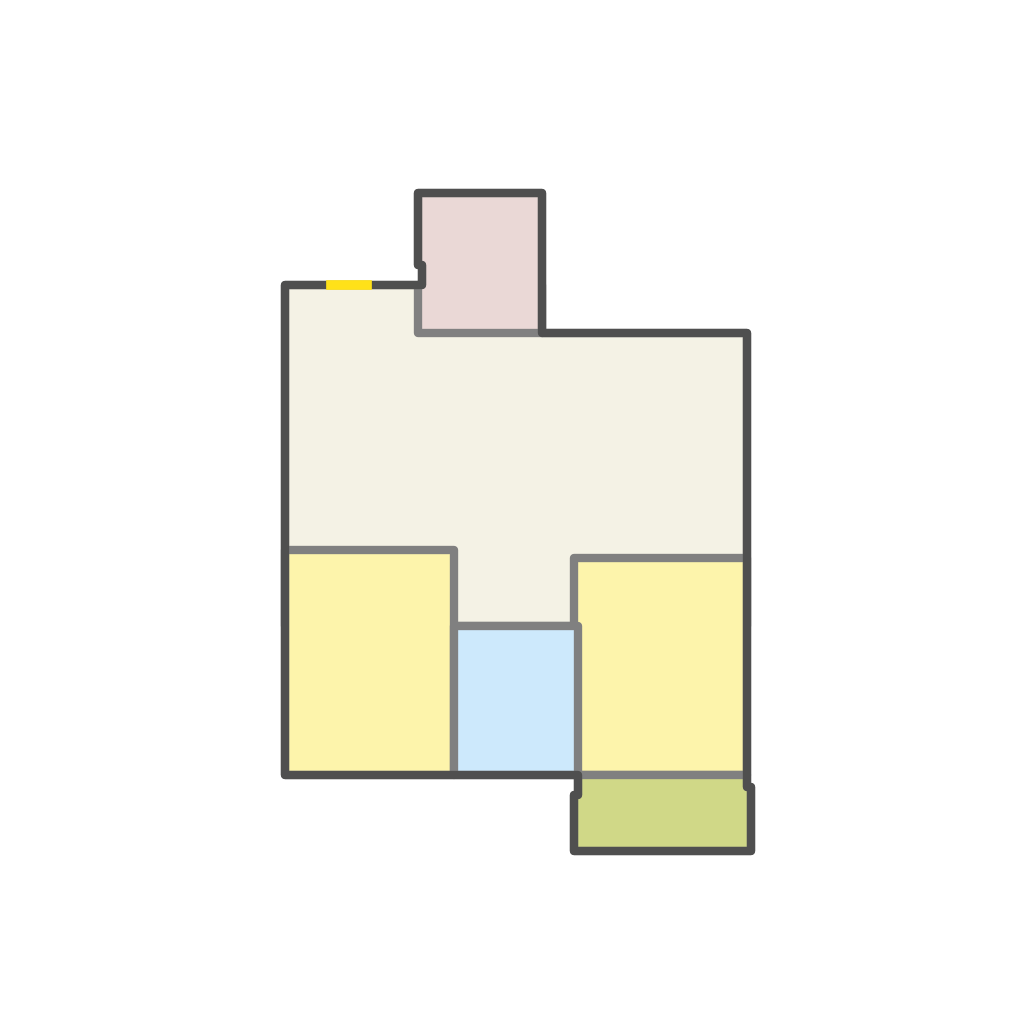}
    \end{subfigure}
    \\
    \subfloat{
        \raisebox{1.25in}{\rotatebox[origin=r]{90}{\Large Ground Truth}}
    }\hfill\hfill
    \begin{subfigure}{0.24\textwidth}
        \includegraphics[width=\textwidth]{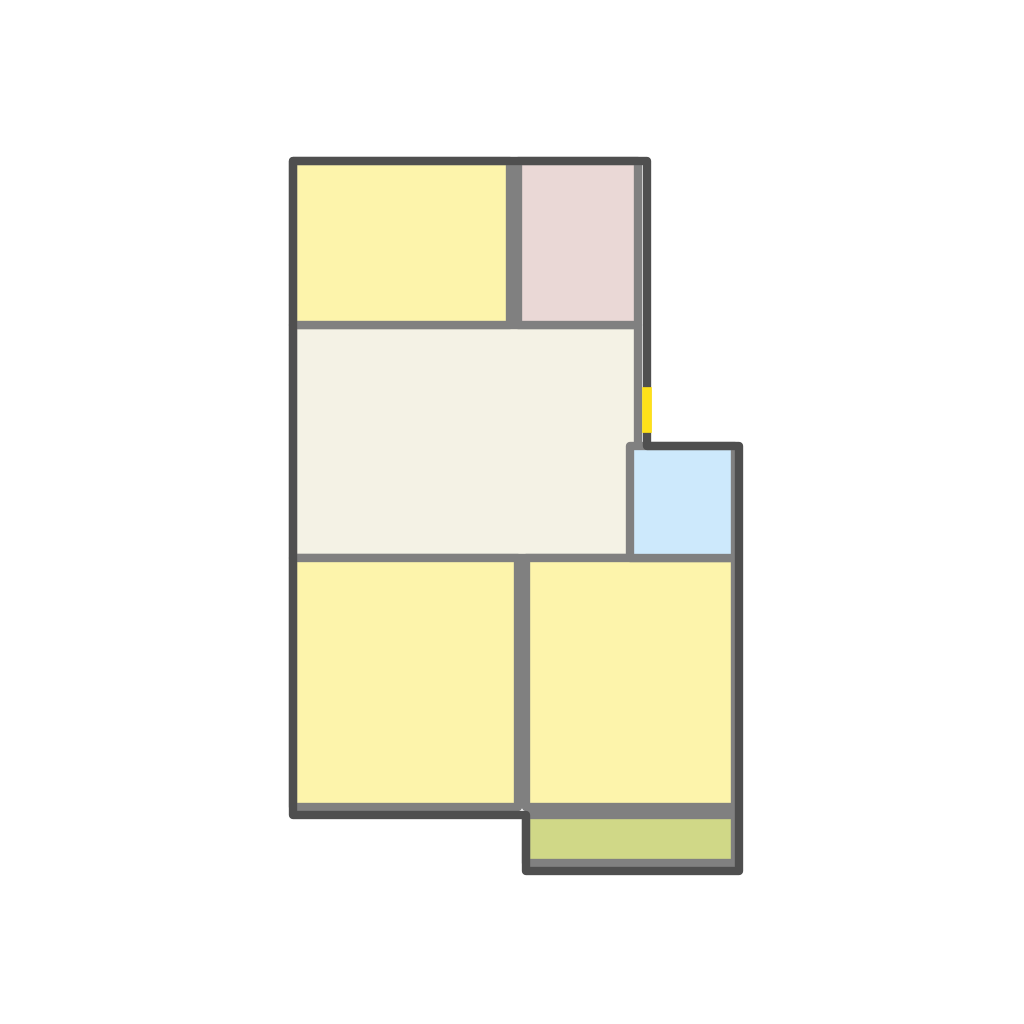}
    \end{subfigure}
    \begin{subfigure}{0.24\textwidth}
        \includegraphics[width=\textwidth]{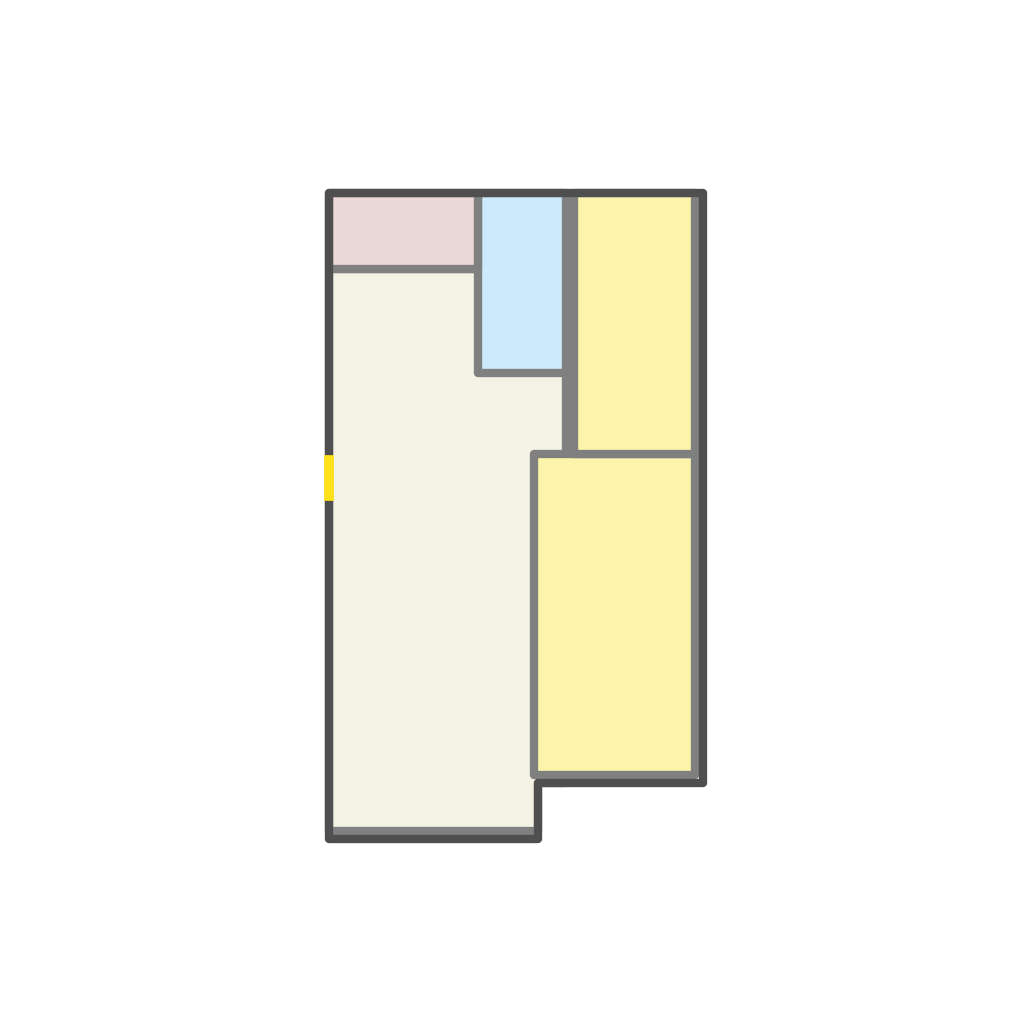}
    \end{subfigure}
    \begin{subfigure}{0.24\textwidth}
        \includegraphics[width=\textwidth]{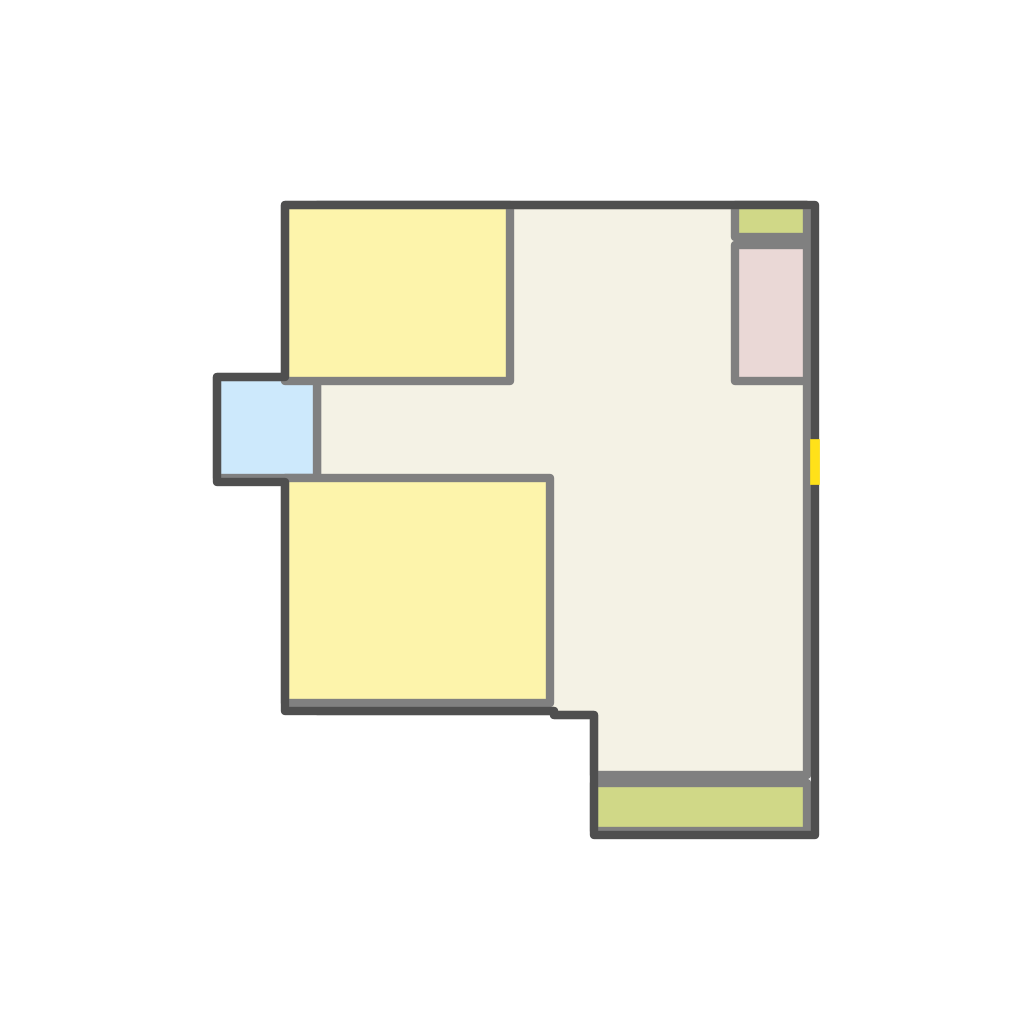}
    \end{subfigure}
    \begin{subfigure}{0.24\textwidth}
        \includegraphics[width=\textwidth]{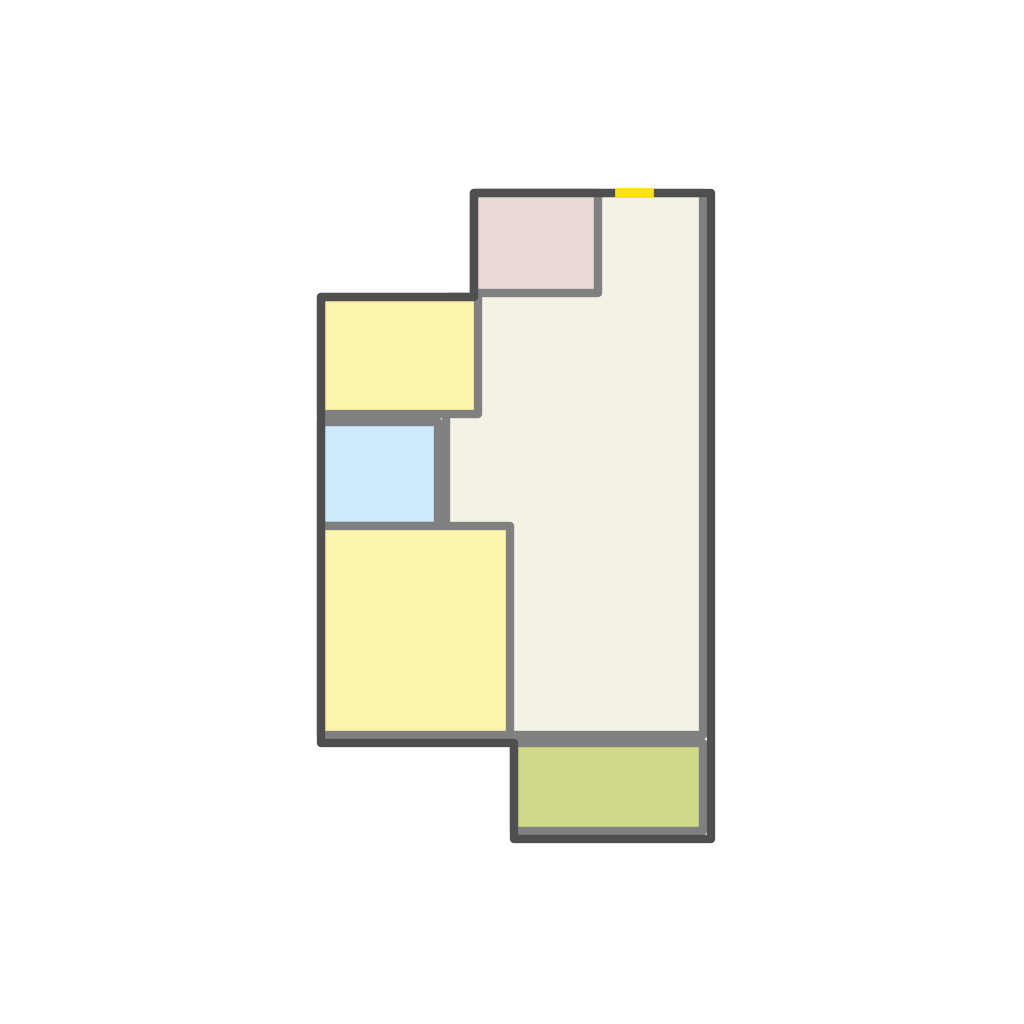}
    \end{subfigure}
    \\
    \subfloat{
        \raisebox{1.15in}{\rotatebox[origin=r]{90}{\Large FP-FGNN}}
    }\hfill
    \begin{subfigure}{0.24\textwidth}
        \includegraphics[width=\textwidth]{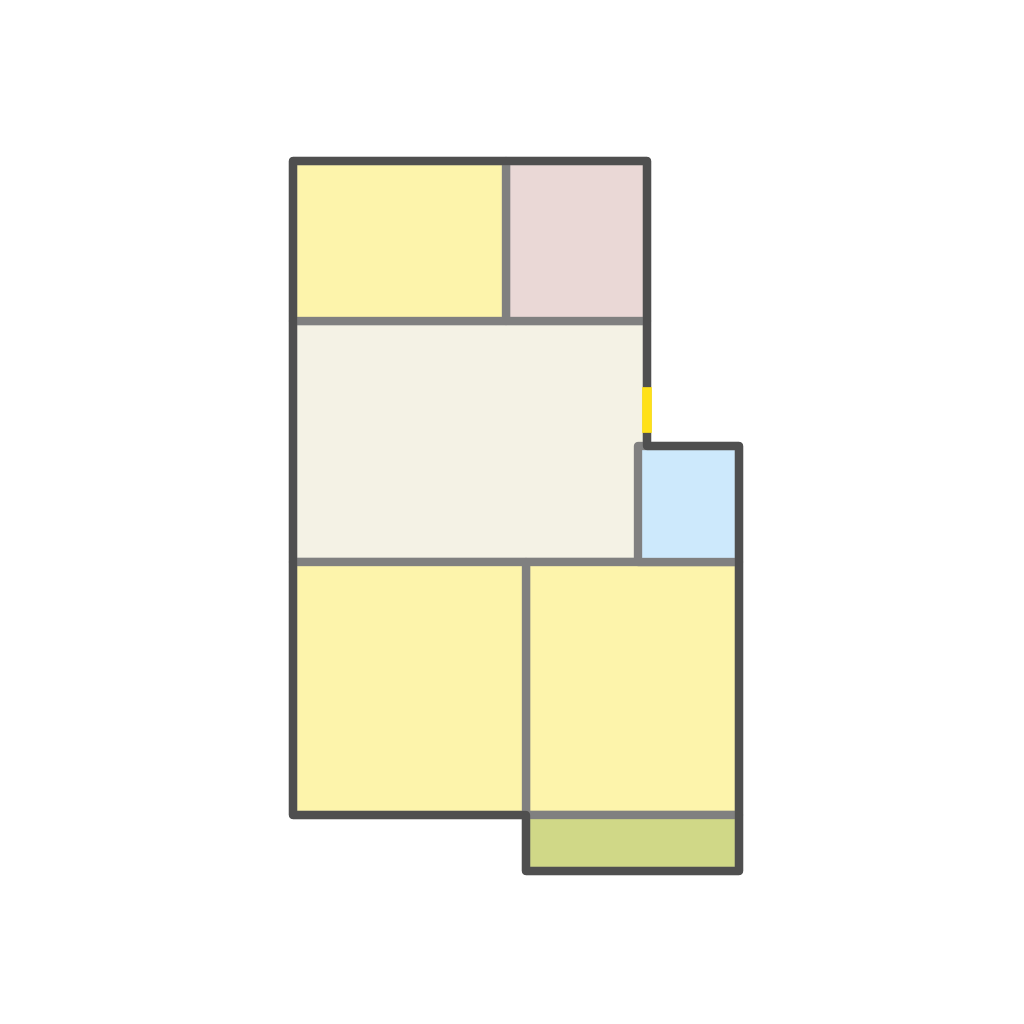}
    \end{subfigure}
    \begin{subfigure}{0.24\textwidth}
        \includegraphics[width=\textwidth]{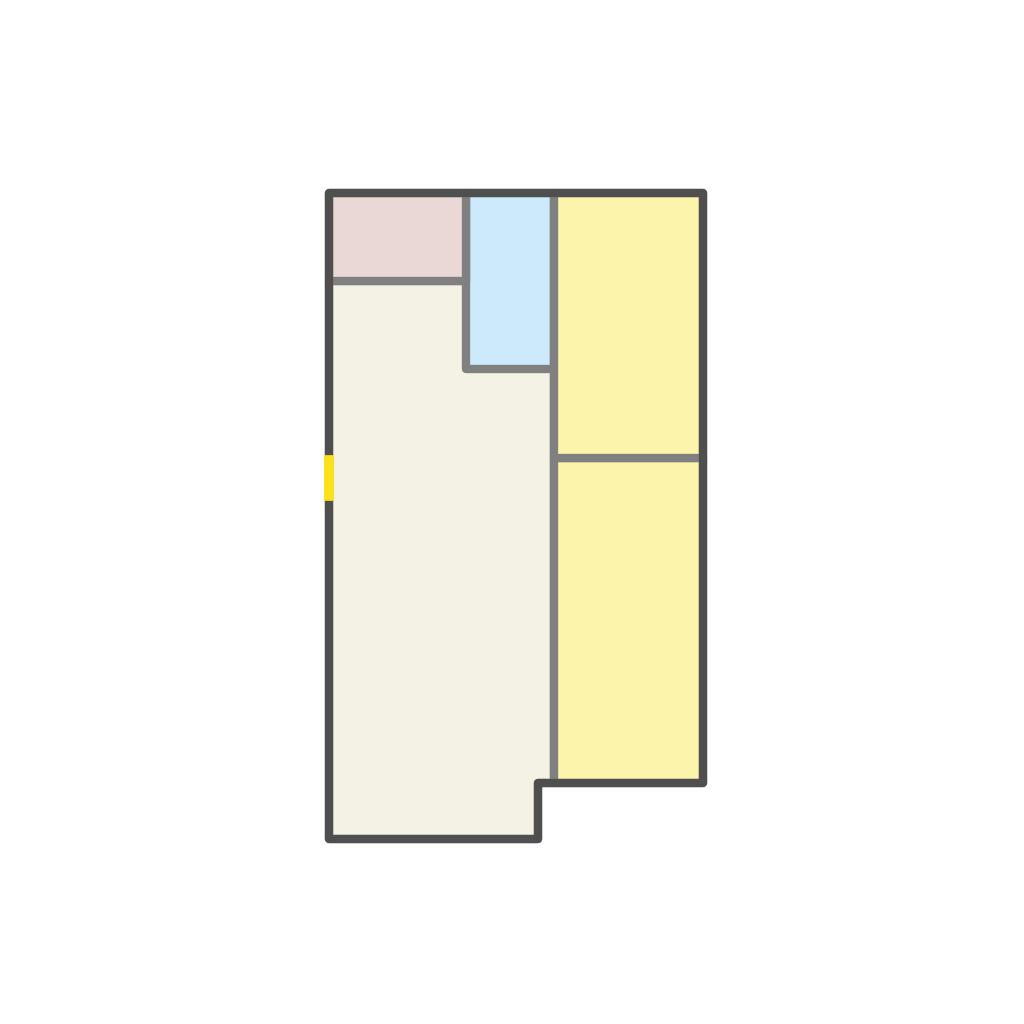}
    \end{subfigure}
    \begin{subfigure}{0.24\textwidth}
        \includegraphics[width=\textwidth]{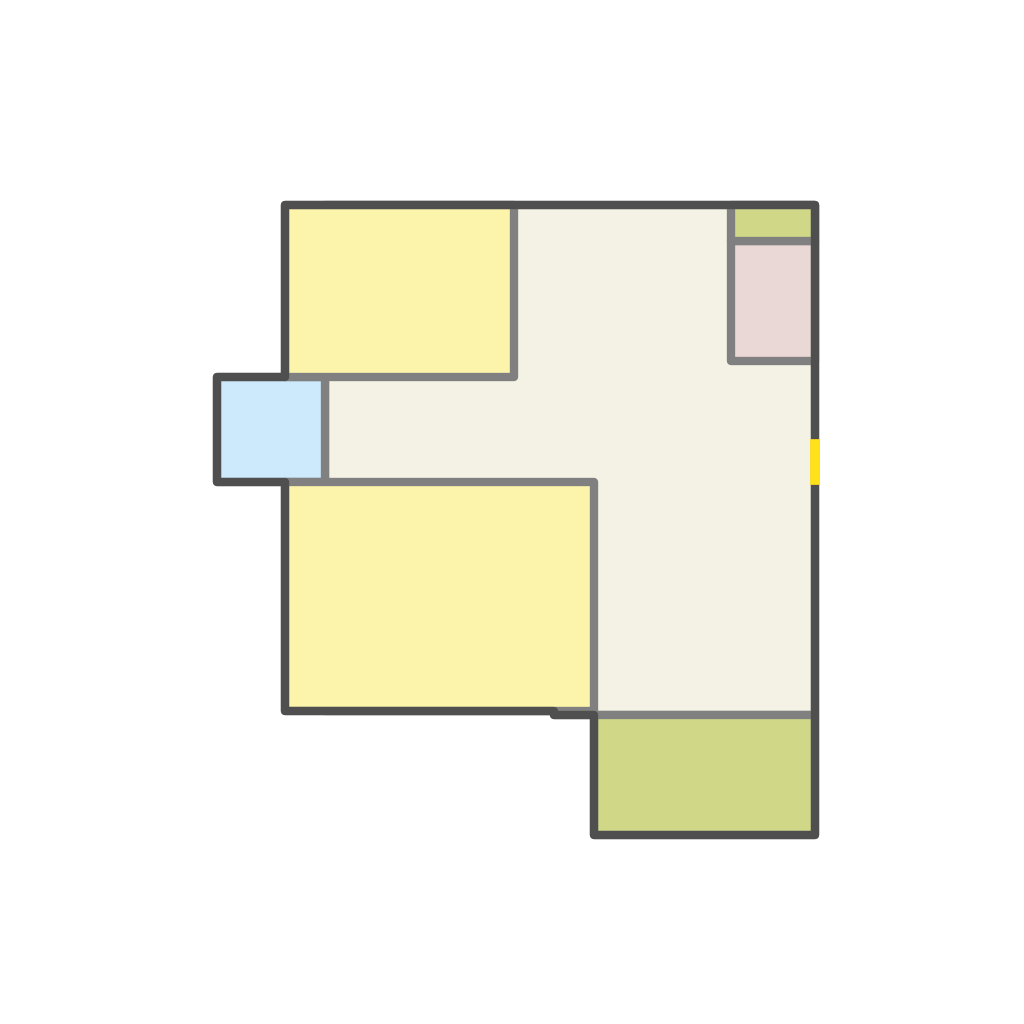}
    \end{subfigure}
    \begin{subfigure}{0.24\textwidth}
        \includegraphics[width=\textwidth]{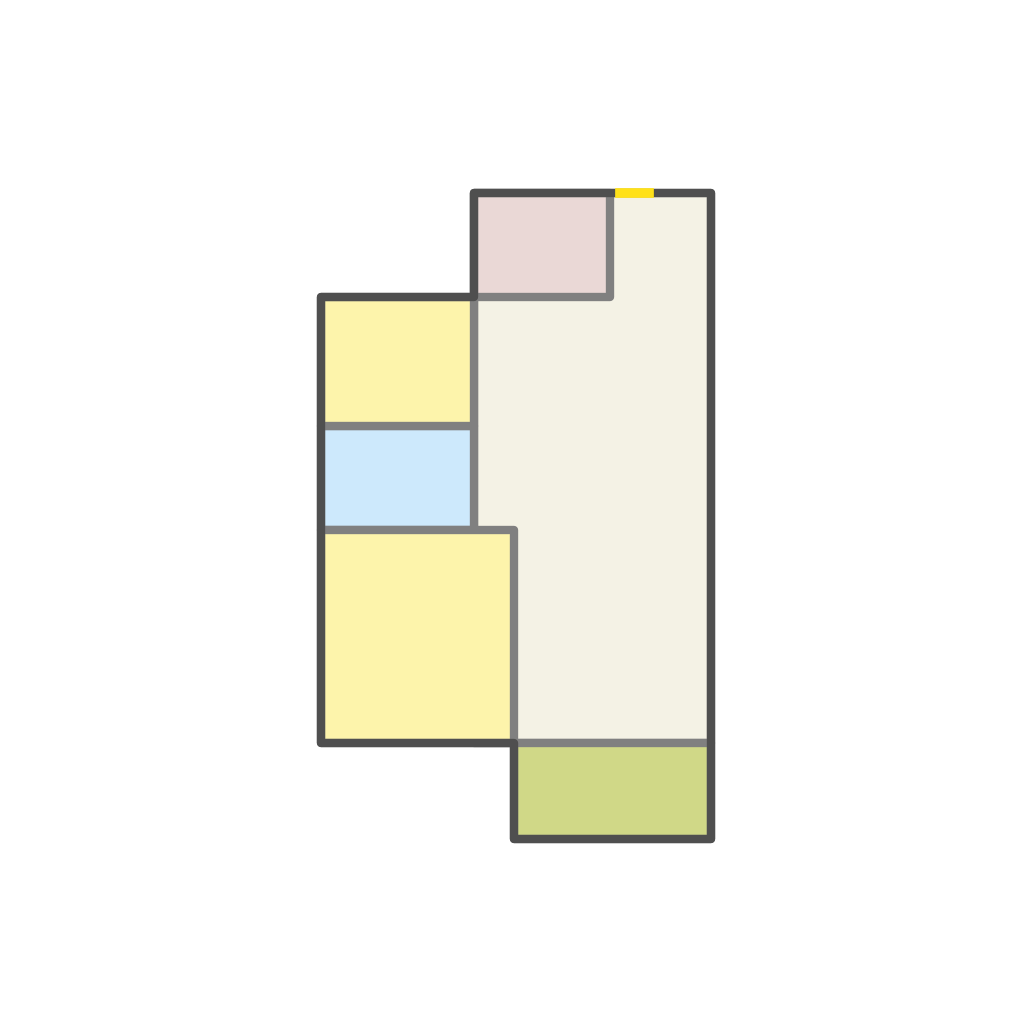}
    \end{subfigure}
    \end{tabular}
    }
    \caption{\small Comparison of floorplans generated by FP-FGNN with the ground truth after post-processing.}
    \label{fig:qualitative-predictions-2}
    \vskip -0.1in
\end{figure*}

In this section, we present a series of additional randomly selected floorplans generated by our model, FP-FGNN. Fig.~\ref{fig:qualitative-predictions-im-1} and Fig.~\ref{fig:qualitative-predictions-im-2} show direct output images of FP-FGNN and Fig.~\ref{fig:qualitative-predictions-1} and Fig.~\ref{fig:qualitative-predictions-2} show the floorplans after post-processing. Alongside each generated floorplan, we provide its corresponding ground truth floorplan to facilitate a comparative analysis of their similarities. The visualized results of this analysis demonstrate that FP-FGNN generates floorplans that bear the closest resemblance to the ground truth ones. Therefore, our model appears to have successfully learned the key features and spatial relationships that underlie the creation of accurate floorplans.

\subsection{Implementation Details}
\label{subsec:sup_implementation_details}
Our code is written with Pytorch~\cite{paszke2017automatic}. We train our model with a learning rate of 0.001, decay rate of 0.0001 along with a step size of $7$, hidden dimension of $128$ and with batch size of $60$. 4 iterations of message passing are performed. We use Adam~\cite{kingma2014adam} as the optimizer. 

\clearpage

\clearpage

\clearpage

\end{document}